%% file: main.tex
\documentclass{book}
\usepackage{roboto}

\usepackage[english]{babel}
\usepackage[letterpaper,top=2cm,bottom=2cm,left=3cm,right=3cm,marginparwidth=1.75cm]{geometry}

\usepackage{amsfonts}
\UseRawInputEncoding

\usepackage{pdflscape}
\usepackage{rotating}
\usepackage{longtable}
\usepackage{array}
\usepackage{float}
\usepackage{amsmath}
\usepackage{graphicx}
\usepackage{inconsolata}
\usepackage[colorlinks=true, allcolors=blue]{hyperref}
\usepackage{enumitem}
\usepackage{tikz} 
\usepackage{listings}
\usepackage{xcolor}
\usepackage[T1]{fontenc}
\usepackage{IEEEtrantools}  
\usepackage{epigraph}  

\definecolor{bgcolor}{rgb}{0.97,0.97,0.97}
\definecolor{codeblue}{rgb}{0.1,0.1,0.8}
\definecolor{codegreen}{rgb}{0,0.4,0}
\definecolor{codegray}{rgb}{0.4,0.4,0.4}
\definecolor{codepurple}{rgb}{0.5,0,0.5}
\definecolor{codered}{rgb}{0.6,0.2,0.2}
\definecolor{lightgray}{rgb}{0.9,0.9,0.9}
\definecolor{darkgray}{rgb}{0.6,0.6,0.6} 

\makeatletter
\renewcommand{\paragraph}{%
  \@startsection{paragraph}{4}{\z@}{1ex}{-1em}{\normalfont\normalsize\bfseries\color{gray}}}
\makeatother

\lstdefinestyle{python}{
    language=Python,
    basicstyle=\ttfamily\small\color{black}\usefont{T1}{zi4}{m}{n},  
    keywordstyle=\bfseries\color{codeblue},  
    stringstyle=\color{codegreen},  
    commentstyle=\slshape\color{codegray},  
    showstringspaces=false,
    numbers=left,
    numberstyle=\tiny\color{codegray},  
    stepnumber=1,
    numbersep=8pt,
    frame=single,
    rulecolor=\color{darkgray},  
    breaklines=true,
    backgroundcolor=\color{bgcolor},
    tabsize=4,
    captionpos=b,
    morekeywords={self}, 
}

\lstdefinestyle{cmd}{
    language=bash,
    basicstyle=\ttfamily\small\color{black}\usefont{T1}{zi4}{m}{n},  
    keywordstyle=\bfseries\color{blue},
    stringstyle=\color{codegreen},
    commentstyle=\itshape\color{gray},
    showstringspaces=false,
    numbers=none,
    frame=single,
    rulecolor=\color{darkgray},  
    breaklines=true,
    backgroundcolor=\color{bgcolor},
    tabsize=4,
    captionpos=b,
}

\title{Deep Learning and Machine Learning - Object Detection and Semantic Segmentation: From Theory to Applications}
\author{
    Jintao Ren\textsuperscript{*} \\
    \textit{Aarhus University } \\
    jintaoren@clin.au.dk
    \and
    Ziqian Bi\textsuperscript{*$\dagger$} \\
    \textit{Indiana University} \\
    bizi@iu.edu
    \and
    Qian Niu \\ 
    \textit{Kyoto University} \\
    niu.qian.f44@kyoto-u.jp
    \and
    Xinyuan Song \\
    \textit{Emory University} \\
    xsong30@emory.edu
    \and
    Zekun Jiang \\
    \textit{Sichuan University} \\
    zekun\_jiang@163.com
    \and
    Junyu Liu \\ 
    \textit{Kyoto University} \\
    liu.junyu.82w@st.kyoto-u.ac.jp
    \and
    Benji Peng \\ 
    \textit{AppCubic} \\
    benji@appcubic.com
    \and
    Sen Zhang \\ 
    \textit{Rutgers University} \\
    sen.z@rutgers.edu
    \and
    Xuanhe Pan \\ 
    \textit{University of Wisconsin-Madison} \\
    xpan73@wisc.edu
    \and
    Jinlang Wang \\ 
    \textit{University of Wisconsin-Madison} \\
    jinlang.wang@wisc.edu
    \and
    Keyu Chen\\ 
    \textit{Georgia Institute of Technology} \\
    kchen637@gatech.edu
    \and
    Caitlyn Heqi Yin \\
    \textit{University of Wisconsin-Madison} \\
    hyin66@wisc.edu
    \and
    Pohsun Feng \\
    \textit{National Taiwan Normal University} \\
    41075018h@ntnu.edu.tw
    \and
    Yizhu Wen \\
    \textit{University of Hawaii} \\
    yizhuw@hawaii.edu
    \and
    Tianyang Wang \\ 
    \textit{Xi'an Jiaotong-Liverpool University} \\
    Tianyang.Wang21@student.xjtlu.edu.cn
    \and
    Silin Chen \\
    \textit{Zhejiang University } \\
    A1033439225@gmail.com
    \and
    Ming Li \\ 
    \textit{Georgia Institute of Technology} \\
    mli694@gatech.edu
    \and
    Jiawei Xu \\ 
    \textit{Purdue University} \\
    xu1644@purdue.edu
    \and
    Ming Liu{$\dagger$} \\ 
    \textit{Purdue University} \\
    liu3183@purdue.edu
}

\date{} 

\begin{document}

\maketitle

\begingroup
\renewcommand\thefootnote{}\footnote{
    \textsuperscript{*} Equal contribution \\
    \textsuperscript{$\dagger$} Corresponding author
}
\addtocounter{footnote}{0}
\endgroup

\epigraph{"The rise of Google, the rise of Facebook, the rise of Apple, I think are proof that there is a place for computer science as something that solves problems that people face every day."}{\textit{Eric Schmidt}}

\epigraph{"In theory there is no difference between theory and practice. In practice there is."}{\textit{Yogi Berra}}

\epigraph{"Consciousness, like a complex system of software, has thousands of levels of nested, self-accessing subroutines"}{\textit{Frederick Lenz}}

\epigraph{"There's an old story about the person who wished his computer were as easy to use as his telephone. That wish has come true, since I no longer know how to use my telephone."}{\textit{Bjarne Stroustrup}}

\tableofcontents  

\input{book20}  

\bibliographystyle{ieeetr}
\bibliography{ref}

\end{document}

%% file: book20.tex
\input{01_basic}
\input{02_func}
\input{03_data}
\input{04_handling}
\input{05_oop}

\input{11_dl}
\input{12_nn}
\input{13_opti}
\input{14_tf_torch}
\input{15_cnn}
\input{16_obj}

\input{21_basic}

\input{22_rcnn}
\input{23_yolo}
\input{24_ssd}
\input{25_fpn}
\input{26_sam}

\input{27_other}
\input{28_eval}
\input{29_prac}
\input{30_summary}

%% file: 01_basic.tex
\setcounter{part}{2}
\part{Advancing Your Skills}

\chapter{Basic Python}

\section{Introduction to Python}

\subsection{What is Python?}
Python is a popular, high-level programming language known for its simplicity and readability \cite{van1995python}. It was created by Guido van Rossum and first released in 1991. Python is designed to be easy to read and write, making it an excellent choice for beginners and professionals alike \cite{van1995python2}. 

One of the reasons Python is so popular is its versatility. Python can be used for web development, data analysis, artificial intelligence, automation, scientific computing, and much more \cite{lutz2013learning,russell2016artificial}. Its large and active community has built a vast ecosystem of libraries and frameworks that extend Python's capabilities in almost any field. Whether you are writing a small script to automate a task or developing a complex machine learning model, Python can handle it.

Python is also an interpreted language, which means you can run Python code as soon as you write it without needing to compile it first. This makes Python development faster and more interactive.

In summary, Python is:
\begin{itemize}
    \item High-level: Its syntax is close to human language, making it easy to learn and use.
    \item Interpreted: You can run Python code directly without compiling.
    \item Versatile: Used in various fields such as web development, data science, AI, and more.
    \item Open Source: Python is free to use, and its ecosystem continues to grow thanks to contributions from its large community.
\end{itemize}

\subsection{Installing Python}
Before you can write and run Python code, you need to install Python on your system. Below are step-by-step instructions for different operating systems:

\subsubsection{Installing Python on Windows}
\begin{enumerate}
    \item Download the latest version of Python from the official Python website: \texttt{https://www.python.org/downloads/}.
    \item Run the installer. Make sure to check the box that says \texttt{Add Python to PATH} before clicking on \texttt{Install Now}.
    \item Once the installation is complete, you can verify the installation by opening a command prompt and typing:
    \begin{lstlisting}[style=cmd]
    python --version
    \end{lstlisting}
    This should display the version of Python you just installed.
\end{enumerate}

\subsubsection{Installing Python on macOS}
\begin{enumerate}
    \item macOS usually comes with Python pre-installed, but it's often an outdated version. To install the latest version, use \texttt{Homebrew} (a package manager for macOS). First, install Homebrew by running the following command in your terminal:
    \begin{lstlisting}[style=cmd]
    /bin/bash -c "$(curl -fsSL https://raw.githubusercontent.com/Homebrew/install/HEAD/install.sh)"
    \end{lstlisting}
    \item Once Homebrew is installed, use it to install Python by running:
    \begin{lstlisting}[style=cmd]
    brew install python
    \end{lstlisting}
    \item After installation, verify the Python installation by typing:
    \begin{lstlisting}[style=cmd]
    python3 --version
    \end{lstlisting}
\end{enumerate}

\subsubsection{Installing Python on Linux}
Most Linux distributions come with Python pre-installed. However, to install or upgrade to the latest version, follow these steps:
\begin{enumerate}
    \item Open a terminal and update the package list:
    \begin{lstlisting}[style=cmd]
    sudo apt update
    \end{lstlisting}
    \item Install Python using the package manager:
    \begin{lstlisting}[style=cmd]
    sudo apt install python3
    \end{lstlisting}
    \item Verify the installation by typing:
    \begin{lstlisting}[style=cmd]
    python3 --version
    \end{lstlisting}
\end{enumerate}

\subsection{Running Python Code}
There are several ways to run Python code. You can use the Python shell for quick tests, write scripts for more complex tasks, or use an Integrated Development Environment (IDE) for larger projects.

\subsubsection{Python Shell}
The Python shell is an interactive environment where you can write and execute Python commands line by line. To open the Python shell, simply type \texttt{python} (or \texttt{python3} on some systems) in your terminal or command prompt:
\begin{lstlisting}[style=cmd]
python
\end{lstlisting}
Once in the shell, you can type Python commands and see immediate results:
\begin{lstlisting}[style=python]
>>> print("Hello, World!")
Hello, World!
\end{lstlisting}
To exit the Python shell, press \texttt{Ctrl + D} or type \texttt{exit()}.

\subsubsection{Running Python Scripts}
Python scripts are saved in files with the \texttt{.py} extension. You can write a Python script using any text editor, save it, and run it from the terminal. Here's how you can write and run a simple script:
\begin{enumerate}
    \item Open a text editor and write the following Python code:
    \begin{lstlisting}[style=python]
    print("Hello from a Python script!")
    \end{lstlisting}
    \item Save the file as \texttt{hello.py}.
    \item In the terminal, navigate to the directory where you saved the file and run it by typing:
    \begin{lstlisting}[style=cmd]
    python hello.py
    \end{lstlisting}
    You should see the output:
    \begin{lstlisting}[style=cmd]
    Hello from a Python script!
    \end{lstlisting}
\end{enumerate}

\subsubsection{Using an IDE (e.g., VSCode, PyCharm)}
IDEs provide a more advanced environment for writing and running Python code. They often include features like debugging, syntax highlighting, and code completion. Popular IDEs for Python include \texttt{VSCode \cite{VSCode}}, \texttt{PyCharm \cite{PyCharm}}, and \texttt{Jupyter Notebooks \cite{jupyterbook}}.

To use an IDE:
\begin{enumerate}
    \item Install an IDE like \texttt{VSCode} from \url{https://code.visualstudio.com/} or \texttt{PyCharm} from \url{https://www.jetbrains.com/pycharm/}.
    \item Open the IDE and create a new Python file, such as \texttt{hello.py}.
    \item Write your Python code and run it directly within the IDE by clicking the "Run" button or using the terminal integrated within the IDE.
\end{enumerate}

\section{Basic Syntax}

\subsection{Indentation}
Python uses indentation to define code blocks, which is different from many other programming languages that use braces \texttt{\{\}} or keywords. Proper indentation is critical in Python, as incorrect indentation will lead to errors.

Here is an example of using indentation in Python:
\begin{lstlisting}[style=python]
if 5 > 2:
    print("Five is greater than two!")
\end{lstlisting}
In this example, the print statement is indented to show that it is part of the \texttt{if} block. If the indentation is incorrect, Python will raise an \texttt{IndentationError}.

Make sure to be consistent with your indentation throughout your code. The standard practice is to use four spaces per indentation level.

\subsection{Comments}
Comments are lines in your code that are ignored by Python. They are useful for explaining what your code does, making it easier to understand for yourself and others. There are two types of comments in Python:

\subsubsection{Single-line Comments}
Single-line comments start with the \texttt{\#} symbol. Everything after the \texttt{\#} on that line will be ignored by Python. Here is an example:
\begin{lstlisting}[style=python]
# This is a single-line comment
print("Hello, World!")
\end{lstlisting}

\subsubsection{Multi-line Comments}
Python does not have a specific syntax for multi-line comments. However, you can create a multi-line comment by using triple quotes:
\begin{lstlisting}[style=python]
"""
This is a multi-line comment.
It spans multiple lines.
"""
print("This is Python!")
\end{lstlisting}

\subsection{Variables and Data Types}
Variables are used to store data that can be used and manipulated in your program. In Python, you do not need to declare the type of a variable explicitly. The type is determined automatically based on the value assigned to it.

Here's an example of creating variables in Python:
\begin{lstlisting}[style=python]
x = 5        # x is an integer
y = 3.14     # y is a float
name = "Alice"  # name is a string
is-active = True  # is-active is a boolean
\end{lstlisting}

\subsubsection{Basic Data Types}
Python has several built-in data types:
\begin{itemize}
    \item \textbf{Integers:} Whole numbers, e.g., \texttt{5}, \texttt{-10}.
    \item \textbf{Floats:} Numbers with a decimal point, e.g., \texttt{3.14}, \texttt{-0.5}.
    \item \textbf{Strings:} A sequence of characters, e.g., \texttt{"Hello"}, \texttt{'World'}.
    \item \textbf{Booleans:} Represents either \texttt{True} or \texttt{False}.
\end{itemize}

%% file: 02_func.tex
\section{Control Structures}

\subsection{if Statements}
Conditional statements are fundamental in programming, allowing your program to make decisions based on certain conditions. In Python, the most common conditional statement is the \texttt{if} statement. 

The syntax for an \texttt{if} statement is as follows:

\begin{lstlisting}[style=python]
if condition:
    # Code block to execute if the condition is True
\end{lstlisting}

Let's look at an example:

\begin{lstlisting}[style=python]
age = 18
if age >= 18:
    print("You are eligible to vote!")
\end{lstlisting}

In this example, the program checks if the value of \texttt{age} is greater than or equal to 18. If the condition evaluates to \texttt{True}, the message "You are eligible to vote!" is printed.

Python also provides \texttt{elif} (else if) and \texttt{else} clauses, allowing us to chain multiple conditions:

\begin{lstlisting}[style=python]
age = 16
if age >= 18:
    print("You are eligible to vote!")
elif age >= 16:
    print("You can apply for a driving permit.")
else:
    print("You are still too young!")
\end{lstlisting}

In this example, the program checks several conditions. The first \texttt{if} checks if the age is 18 or older. If that condition is not met, it moves on to the \texttt{elif} block. If both fail, it defaults to the \texttt{else} block.

\subsection{for Loops}
A \texttt{for} loop is used to iterate over a sequence (such as a list, tuple, dictionary, string, or range). Python's \texttt{for} loop allows you to execute a block of code a specified number of times or over the elements of a collection.

The syntax of a \texttt{for} loop is:

\begin{lstlisting}[style=python]
for variable in sequence:
    # Code block to execute for each element in the sequence
\end{lstlisting}

Here is an example of a \texttt{for} loop iterating over a list:

\begin{lstlisting}[style=python]
fruits = ["apple", "banana", "cherry"]
for fruit in fruits:
    print(fruit)
\end{lstlisting}

In this example, the loop iterates over each item in the list \texttt{fruits} and prints it.

You can also use the \texttt{range()} function to generate a sequence of numbers:

\begin{lstlisting}[style=python]
for i in range(5):
    print(i)
\end{lstlisting}

This will output the numbers 0 to 4, as \texttt{range(5)} generates numbers from 0 up to, but not including, 5.

\subsection{while Loops}
A \texttt{while} loop runs as long as the specified condition is \texttt{True}. The syntax of a \texttt{while} loop is:

\begin{lstlisting}[style=python]
while condition:
    # Code block to execute while the condition is True
\end{lstlisting}

For example:

\begin{lstlisting}[style=python]
count = 0
while count < 5:
    print(count)
    count += 1
\end{lstlisting}

In this example, the loop will continue running as long as \texttt{count} is less than 5. The value of \texttt{count} increases by 1 on each iteration until the condition becomes \texttt{False}.

\subsection{break and continue}
Python provides the \texttt{break} and \texttt{continue} statements to control the flow of loops.

The \texttt{break} statement exits the loop completely:

\begin{lstlisting}[style=python]
for i in range(5):
    if i == 3:
        break
    print(i)
\end{lstlisting}

In this example, the loop stops as soon as \texttt{i} becomes 3, so the output will be 0, 1, 2.

The \texttt{continue} statement skips the current iteration and continues with the next one:

\begin{lstlisting}[style=python]
for i in range(5):
    if i == 3:
        continue
    print(i)
\end{lstlisting}

Here, when \texttt{i} equals 3, the \texttt{continue} statement skips that iteration. Thus, the output will be 0, 1, 2, 4.

\section{Functions}

\subsection{Defining Functions}
A function is a block of reusable code that is used to perform a specific task. In Python, you can define a function using the \texttt{def} keyword, followed by the function name and parentheses.

The basic syntax for defining a function is:

\begin{lstlisting}[style=python]
def function_name():
    # Code block that belongs to the function
\end{lstlisting}

For example, let's define a simple function that prints a message:

\begin{lstlisting}[style=python]
def greet():
    print("Hello, World!")
\end{lstlisting}

To call the function, simply use its name followed by parentheses:

\begin{lstlisting}[style=python]
greet()  # Outputs: Hello, World!
\end{lstlisting}

\subsection{Arguments and Return Values}
Functions can accept arguments (also called parameters) to make them more flexible. You define parameters inside the parentheses.

\begin{lstlisting}[style=python]
def greet(name):
    print(f"Hello, {name}!")
\end{lstlisting}

Now, when you call the function, you can pass an argument:

\begin{lstlisting}[style=python]
greet("Alice")  # Outputs: Hello, Alice!
\end{lstlisting}

Functions can also return values using the \texttt{return} statement:

\begin{lstlisting}[style=python]
def add(a, b):
    return a + b
\end{lstlisting}

When you call this function, it returns the result of adding \texttt{a} and \texttt{b}:

\begin{lstlisting}[style=python]
result = add(3, 5)
print(result)  # Outputs: 8
\end{lstlisting}

\subsection{Lambda Functions}
Lambda functions, also known as anonymous functions, are small and concise functions that can have any number of arguments but only one expression. They are often used for simple operations that are passed as arguments to other functions.

The syntax for a lambda function is:

\begin{lstlisting}[style=python]
lambda arguments: expression
\end{lstlisting}

For example, here's how to use a lambda function to double a number:

\begin{lstlisting}[style=python]
double = lambda x: x * 2
print(double(5))  # Outputs: 10
\end{lstlisting}

Lambda functions are useful when you need a simple function for a short period of time, especially within higher-order functions like \texttt{map()} or \texttt{filter()}.

\subsection{Recursion}
Recursion is a technique in which a function calls itself to solve smaller instances of the same problem. Recursive functions must have a base case to prevent infinite loops.

Here's an example of a recursive function that calculates the factorial of a number:

\begin{lstlisting}[style=python]
def factorial(n):
    if n == 1:
        return 1
    else:
        return n * factorial(n - 1)
\end{lstlisting}

When you call \texttt{factorial(5)}, it calculates:

\[
5 \times 4 \times 3 \times 2 \times 1 = 120
\]

Thus, \texttt{factorial(5)} returns 120.

Recursive functions can be powerful but should be used carefully to avoid infinite recursion and excessive memory usage.

%% file: 03_data.tex
\section{Data Structures}

\subsection{Lists}
Lists in Python are ordered, mutable collections of items, meaning you can change, add, or remove items after the list has been created. Lists can store any data type, including numbers, strings, and even other lists. You can think of a list as a container for elements arranged in a specific order.

\textbf{Creating a List:}

To create a list in Python, you use square brackets `[ ]` and separate each element with a comma.

\begin{lstlisting}[style=python]
# Example of creating a list
my_list = [1, 2, 3, 4, 5]
print(my_list)
\end{lstlisting}

\textbf{Accessing Elements in a List:}

You can access elements from a list by their index. Python uses zero-based indexing, so the first element is at index 0.

\begin{lstlisting}[style=python]
# Accessing the first element
print(my_list[0])  # Output: 1

# Accessing the last element
print(my_list[-1])  # Output: 5
\end{lstlisting}

\textbf{Modifying a List:}

Because lists are mutable, you can modify elements at specific positions:

\begin{lstlisting}[style=python]
# Changing the second element
my_list[1] = 10
print(my_list)  # Output: [1, 10, 3, 4, 5]
\end{lstlisting}

\textbf{Adding Elements to a List:}

You can add new items to a list using the `append()` method or insert elements at a specific position using `insert()`.

\begin{lstlisting}[style=python]
# Adding an element to the end of the list
my_list.append(6)
print(my_list)  # Output: [1, 10, 3, 4, 5, 6]

# Inserting an element at a specific position
my_list.insert(2, 99)
print(my_list)  # Output: [1, 10, 99, 3, 4, 5, 6]
\end{lstlisting}

\textbf{Removing Elements from a List:}

You can remove items from a list using methods like `remove()` or `pop()`.

\begin{lstlisting}[style=python]
# Removing a specific element by value
my_list.remove(99)
print(my_list)  # Output: [1, 10, 3, 4, 5, 6]

# Removing the last element
my_list.pop()
print(my_list)  # Output: [1, 10, 3, 4, 5]
\end{lstlisting}

\subsection{Tuples}
Tuples are similar to lists but with one major difference: they are immutable, meaning that once a tuple is created, its elements cannot be modified. Tuples are useful when you need a collection of items that should not change.

\textbf{Creating a Tuple:}

To create a tuple, you use parentheses `()` instead of square brackets.

\begin{lstlisting}[style=python]
# Example of creating a tuple
my_tuple = (1, 2, 3)
print(my_tuple)
\end{lstlisting}

\textbf{Accessing Elements in a Tuple:}

Just like lists, tuples are indexed starting at 0.

\begin{lstlisting}[style=python]
# Accessing elements of a tuple
print(my_tuple[0])  # Output: 1
print(my_tuple[-1])  # Output: 3
\end{lstlisting}

\textbf{Tuple Immutability:}

Once a tuple is created, you cannot change its elements. Trying to modify a tuple will result in an error.

\begin{lstlisting}[style=python]
# Attempting to change a tuple will cause an error
my_tuple[1] = 10  # This will raise a TypeError
\end{lstlisting}

Tuples are often used when you want to group related data and ensure it does not change throughout your program.

\subsection{Dictionaries}
A dictionary in Python stores data in key-value pairs. Each key is associated with a value, and you can use the key to access the corresponding value. Dictionaries are unordered, mutable, and cannot have duplicate keys.

\textbf{Creating a Dictionary:}

You create a dictionary using curly braces `{}` and separate keys and values with a colon `:`.

\begin{lstlisting}[style=python]
# Example of creating a dictionary
my_dict = {"name": "Alice", "age": 25, "city": "New York"}
print(my_dict)
\end{lstlisting}

\textbf{Accessing Values in a Dictionary:}

You can access the value associated with a specific key using square brackets.

\begin{lstlisting}[style=python]
# Accessing a value by its key
print(my_dict["name"])  # Output: Alice
\end{lstlisting}

\textbf{Adding and Modifying Key-Value Pairs:}

You can add new key-value pairs or modify existing ones by assigning a value to a key.

\begin{lstlisting}[style=python]
# Adding a new key-value pair
my_dict["email"] = "alice@example.com"
print(my_dict)

# Modifying an existing value
my_dict["age"] = 26
print(my_dict)
\end{lstlisting}

\textbf{Removing Key-Value Pairs:}

You can remove items from a dictionary using the `del` keyword or the `pop()` method.

\begin{lstlisting}[style=python]
# Removing a key-value pair using del
del my_dict["city"]
print(my_dict)

# Removing a key-value pair using pop
my_dict.pop("email")
print(my_dict)
\end{lstlisting}

\subsection{Sets}
Sets are unordered collections of unique elements, meaning that no element can appear more than once in a set. Sets are useful for performing mathematical operations like union, intersection, and difference.

\textbf{Creating a Set:}

You can create a set by using curly braces `{}` or the `set()` function.

\begin{lstlisting}[style=python]
# Example of creating a set
my_set = {1, 2, 3, 4, 5}
print(my_set)
\end{lstlisting}

\textbf{Adding and Removing Elements:}

You can add new elements to a set using `add()` and remove elements using `remove()`.

\begin{lstlisting}[style=python]
# Adding an element to a set
my_set.add(6)
print(my_set)

# Removing an element from a set
my_set.remove(3)
print(my_set)
\end{lstlisting}

\textbf{Set Operations:}

Sets support operations like union, intersection, and difference.

\begin{lstlisting}[style=python]
set1 = {1, 2, 3}
set2 = {3, 4, 5}

# Union of two sets
print(set1.union(set2))  # Output: {1, 2, 3, 4, 5}

# Intersection of two sets
print(set1.intersection(set2))  # Output: {3}

# Difference of two sets
print(set1.difference(set2))  # Output: {1, 2}
\end{lstlisting}

\section{File Handling}

\subsection{Opening and Closing Files}
In Python, file handling allows you to read and write data to files. The process begins by opening a file, performing operations (read/write), and then closing the file to free up system resources.

\textbf{Opening a File:}

You can open a file using the built-in `open()` function. The `open()` function takes the file path and mode (read, write, etc.) as arguments.

\begin{lstlisting}[style=python]
# Open a file in read mode ('r')
file = open("example.txt", "r")
\end{lstlisting}

\textbf{Closing a File:}

It is important to close a file once you are done with it. This ensures that any changes made to the file are saved properly.

\begin{lstlisting}[style=python]
# Closing the file
file.close()
\end{lstlisting}

\subsection{Reading Files}
There are several ways to read a file in Python. You can read the entire content at once, read it line by line, or read a specific number of characters.

\begin{lstlisting}[style=python]
# Reading the entire content of the file
file = open("example.txt", "r")
content = file.read()
print(content)
file.close()

# Reading file line by line
file = open("example.txt", "r")
for line in file:
    print(line)
file.close()
\end{lstlisting}

\subsection{Writing Files}
To write to a file, you open it in write mode (`'w'`), append mode (`'a'`), or read/write mode (`'r+'`). When writing to a file, if the file does not exist, it will be created. If it exists, the content may be overwritten depending on the mode.

\begin{lstlisting}[style=python]
# Writing to a file
file = open("output.txt", "w")
file.write("This is an example.")
file.close()

# Appending to a file
file = open("output.txt", "a")
file.write("\nAppending a new line.")
file.close()
\end{lstlisting}

\subsection{File Methods}
Python provides various file methods, such as:

\begin{itemize}
    \item `read()` - Reads the entire file content.
    \item `write()` - Writes data to the file.
    \item `close()` - Closes the file.
    \item `seek()` - Moves the file pointer to a specific position.
\end{itemize}

\begin{lstlisting}[style=python]
# Example of using seek to move the file pointer
file = open("example.txt", "r")
file.seek(0)  # Move to the beginning of the file
content = file.read()
print(content)
file.close()
\end{lstlisting}

%% file: 04_handling.tex
\section{Error Handling}

\subsection{try and except}
One of the most important concepts in programming is handling errors effectively. In Python, you can use the \texttt{try} and \texttt{except} blocks to catch exceptions and prevent your program from crashing. When an error occurs within a \texttt{try} block, Python will jump to the \texttt{except} block to handle the error gracefully.

Here's a simple example to demonstrate how \texttt{try} and \texttt{except} work:

\begin{lstlisting}[style=python]
try:
    # Trying to divide by zero, which will cause an error
    result = 10 / 0
except ZeroDivisionError:
    # This block will execute if a ZeroDivisionError occurs
    print("You can't divide by zero!")
\end{lstlisting}

In this example, Python will raise a \texttt{ZeroDivisionError} when attempting to divide by zero. The error is caught by the \texttt{except} block, which prints a user-friendly message instead of crashing the program.

You can also catch multiple types of exceptions by specifying different types of errors:

\begin{lstlisting}[style=python]
try:
    # Some code that might throw multiple exceptions
    value = int(input("Enter a number: "))
    result = 10 / value
except ValueError:
    print("That was not a valid number.")
except ZeroDivisionError:
    print("You can't divide by zero!")
\end{lstlisting}

In this code, the program handles both \texttt{ValueError} (invalid input) and \texttt{ZeroDivisionError} (division by zero) gracefully.

\subsection{finally}
The \texttt{finally} block is used to execute code no matter what happens in the \texttt{try} and \texttt{except} blocks. This is useful when you need to clean up resources, such as closing files or releasing network connections, regardless of whether an exception occurred.

For example:

\begin{lstlisting}[style=python]
try:
    file = open("data.txt", "r")
    # Read from file
except FileNotFoundError:
    print("File not found!")
finally:
    # This block will always execute
    file.close()
    print("File closed.")
\end{lstlisting}

In this case, the file will be closed no matter if an exception occurs or not. The \texttt{finally} block ensures that important cleanup tasks are performed, such as closing a file after it has been opened.

\subsection{Raising Exceptions}
Sometimes, you might want to raise exceptions intentionally. This can be useful when you're writing functions that need to enforce certain conditions. You can use the \texttt{raise} statement to throw an exception manually.

Here's an example of raising a custom exception:

\begin{lstlisting}[style=python]
def check_positive(number):
    if number < 0:
        raise ValueError("Negative numbers are not allowed!")
    return number

try:
    check_positive(-5)
except ValueError as e:
    print(e)
\end{lstlisting}

In this example, the function \texttt{check\_positive} raises a \texttt{ValueError} if the input is negative. The exception is caught in the \texttt{except} block, and a user-friendly error message is displayed.

\section{Modules and Packages}

\subsection{Importing Modules}
In Python, a module is simply a file containing Python code, such as functions and classes, which can be imported into other files. Python comes with many built-in modules that can be used directly. To use a module, you need to import it first using the \texttt{import} statement.

Here's an example using the built-in \texttt{math} module:

\begin{lstlisting}[style=python]
import math

# Using a function from the math module
result = math.sqrt(16)
print(result)  # Output: 4.0
\end{lstlisting}

By importing the \texttt{math} module, you can access mathematical functions like \texttt{sqrt()}, which calculates the square root of a number.

You can also import specific functions from a module to avoid importing everything:

\begin{lstlisting}[style=python]
from math import pi

# Now you can use pi directly without the math. prefix
print(pi)  # Output: 3.141592653589793
\end{lstlisting}

This allows you to import only the functions or constants you need, making your code cleaner and more efficient.

\subsection{Creating Modules}
You can create your own modules by simply writing Python code in a file and saving it with a \texttt{.py} extension. For example, let's create a module called \texttt{my\_module.py}:

\begin{lstlisting}[style=python]
# File: my_module.py

def greet(name):
    return f"Hello, {name}!"

def add(a, b):
    return a + b
\end{lstlisting}

Once you've created your module, you can import and use it in other files:

\begin{lstlisting}[style=python]
import my_module

print(my_module.greet("Alice"))  # Output: Hello, Alice!
print(my_module.add(5, 3))  # Output: 8
\end{lstlisting}

This is a powerful way to organize your code and reuse functionality across different parts of your project.

\subsection{Using \texttt{pip} to Install Packages}
Python has a vast ecosystem of third-party libraries and packages that can be installed using \texttt{pip}, Python's package manager. These packages allow you to add additional functionality to your projects without having to write everything from scratch.

To install a package, you can run the following command in your terminal:

\begin{lstlisting}[style=cmd]
pip install requests
\end{lstlisting}

For example, the \texttt{requests} package is a popular library for making HTTP requests. Once installed, you can use it in your Python code:

\begin{lstlisting}[style=python]
import requests

response = requests.get("https://api.github.com")
print(response.status_code)  # Output: 200
\end{lstlisting}

This demonstrates how easy it is to leverage third-party libraries to add powerful features to your Python projects.

To see a list of all installed packages, you can run:

\begin{lstlisting}[style=cmd]
pip list
\end{lstlisting}

This will display all the packages currently installed in your Python environment.

%% file: 05_oop.tex
\section{Object-Oriented Programming}

Object-Oriented Programming (OOP) is a programming paradigm that revolves around the concept of "objects"\cite{jana2005java}. Objects are instances of classes, and they bundle data (attributes) and functionality (methods) together. Python is a powerful, object-oriented language, and this section will guide you through the basics of OOP in Python. 

\subsection{Classes and Objects}

At the heart of OOP are classes and objects. A class serves as a blueprint for creating objects, which are instances of that class. Let's start by understanding how to define a class and create objects in Python.

\textbf{Class Definition:} A class in Python is defined using the \texttt{class} keyword, followed by the class name and a colon. Inside the class, methods (functions) and attributes (variables) define the characteristics and behavior of the class.

\begin{lstlisting}[style=python]
# Defining a simple class named 'Car'
class Car:
    # Constructor method to initialize object attributes
    def __init__(self, make, model, year):
        self.make = make  # Attribute for car make
        self.model = model  # Attribute for car model
        self.year = year  # Attribute for car manufacturing year

    # Method to display car information
    def display_info(self):
        print(f"This car is a {self.year} {self.make} {self.model}.")

# Creating an object (instance) of the Car class
my_car = Car("Toyota", "Corolla", 2020)

# Accessing the object's method
my_car.display_info()
\end{lstlisting}

In this example:
\begin{itemize}
    \item The class \texttt{Car} has a constructor method \texttt{\_\_init\_\_()} that initializes the attributes \texttt{make}, \texttt{model}, and \texttt{year}.
    \item We create an object \texttt{my\_car}, which is an instance of the \texttt{Car} class, and call the method \texttt{display\_info()} to print the car's details.
\end{itemize}

\subsection{Inheritance}

Inheritance allows us to define a new class that is a modified or extended version of an existing class. The new class (called the child class) inherits attributes and methods from the existing class (called the parent class). This promotes code reuse and allows for the creation of hierarchical relationships between classes.

Let's extend the \texttt{Car} class by creating a new class \texttt{ElectricCar}, which inherits from \texttt{Car}.

\begin{lstlisting}[style=python]
# Defining the parent class 'Car'
class Car:
    def __init__(self, make, model, year):
        self.make = make
        self.model = model
        self.year = year
    
    def display_info(self):
        print(f"This car is a {self.year} {self.make} {self.model}.")

# Defining the child class 'ElectricCar' that inherits from 'Car'
class ElectricCar(Car):
    def __init__(self, make, model, year, battery_size):
        # Call the parent class's constructor using 'super()'
        super().__init__(make, model, year)
        self.battery_size = battery_size  # Additional attribute for electric cars
    
    # Method specific to ElectricCar
    def display_battery_size(self):
        print(f"This car has a {self.battery_size}-kWh battery.")

# Creating an instance of ElectricCar
my_electric_car = ElectricCar("Tesla", "Model S", 2022, 100)

# Accessing methods from both parent and child classes
my_electric_car.display_info()  # Inherited from Car
my_electric_car.display_battery_size()  # Specific to ElectricCar
\end{lstlisting}

In this example:
\begin{itemize}
    \item The \texttt{ElectricCar} class inherits the attributes and methods of \texttt{Car} through the use of \texttt{super()}.
    \item \texttt{ElectricCar} adds a new attribute \texttt{battery\_size} and a method \texttt{display\_battery\_size()} specific to electric cars.
\end{itemize}

\subsection{Polymorphism}

Polymorphism allows objects of different classes to be treated as objects of a common parent class. This is achieved by overriding methods in child classes while maintaining the same method signature as in the parent class. This provides flexibility and allows for dynamic behavior in Python.

Let's see an example where different types of vehicles can behave differently while sharing a common interface.

\begin{lstlisting}[style=python]
# Parent class Vehicle
class Vehicle:
    def start_engine(self):
        print("Starting the engine...")

# Child class Car inherits from Vehicle
class Car(Vehicle):
    def start_engine(self):
        print("The car engine is starting...")

# Child class Motorcycle inherits from Vehicle
class Motorcycle(Vehicle):
    def start_engine(self):
        print("The motorcycle engine is starting...")

# Function to demonstrate polymorphism
def start_vehicle(vehicle):
    vehicle.start_engine()

# Creating objects of different types
my_car = Car()
my_motorcycle = Motorcycle()

# Passing objects to the same function
start_vehicle(my_car)        # The car engine is starting...
start_vehicle(my_motorcycle)  # The motorcycle engine is starting...
\end{lstlisting}

In this example:
\begin{itemize}
    \item Both \texttt{Car} and \texttt{Motorcycle} inherit from the \texttt{Vehicle} class.
    \item Each class overrides the \texttt{start\_engine()} method.
    \item The function \texttt{start\_vehicle()} takes any \texttt{Vehicle} object and calls its \texttt{start\_engine()} method, demonstrating polymorphism.
\end{itemize}

\subsection{Encapsulation}

Encapsulation refers to restricting access to certain parts of an object. This is done by making attributes and methods private, which is achieved by prefixing them with an underscore (\_). In Python, this is a convention to indicate that these members are not meant to be accessed directly from outside the class.

Let's look at how encapsulation works in Python.

\begin{lstlisting}[style=python]
# Defining a class with encapsulation
class BankAccount:
    def __init__(self, balance):
        self._balance = balance  # Private attribute

    # Method to access the private attribute
    def get_balance(self):
        return self._balance
    
    # Method to modify the private attribute
    def deposit(self, amount):
        if amount > 0:
            self._balance += amount
        else:
            print("Deposit amount must be positive.")

# Creating an object of BankAccount
my_account = BankAccount(1000)

# Accessing and modifying the balance through methods
print(my_account.get_balance())  # 1000
my_account.deposit(500)
print(my_account.get_balance())  # 1500
\end{lstlisting}

In this example:
\begin{itemize}
    \item The attribute \texttt{\_balance} is private and should not be accessed directly.
    \item The methods \texttt{get\_balance()} and \texttt{deposit()} provide controlled access to the private attribute.
\end{itemize}

Encapsulation helps to protect the internal state of objects and maintain the integrity of data by restricting direct access.

%% file: 11_dl.tex
\chapter{Deep Learning Fundamentals}

\section{Introduction to Deep Learning}

\subsection{What is Deep Learning?}

Deep learning is a subset of machine learning that utilizes neural networks with many layers to model complex patterns and make predictions \cite{goodfellow2016deep}. At its core, deep learning is inspired by the structure and function of the human brain, specifically the way neurons communicate with each other \cite{lecun2015deep,schmidhuber2015deep}. Deep learning models can automatically learn to represent data hierarchically, meaning that each layer in a neural network extracts increasingly abstract features from the data.

For example, imagine we are trying to build a deep learning model to recognize cats in images. In the first layer, the network might detect simple features like edges and colors. As we move deeper into the network, the model learns more complex patterns like shapes, textures, and eventually the full image of a cat. This hierarchical learning process is what makes deep learning so powerful, particularly in tasks like image recognition \cite{he2016deep, krizhevsky2012imagenet}, speech processing \cite{hinton2012deep}, and natural language understanding \cite{devlin2018bert}.

Deep learning became widely popular due to its remarkable performance on tasks that were previously considered extremely difficult, such as self-driving cars \cite{bojarski2016end}, medical diagnostics \cite{litjens2017survey}, and even playing complex games like Go \cite{silver2017mastering}. The major difference between traditional machine learning and deep learning lies in the depth of the models, which allows deep learning to handle far more complex patterns and datasets \cite{goodfellow2016deep, mitchell1997machine}.

\subsection{Differences between Machine Learning and Deep Learning}

Machine learning (ML) and deep learning (DL) share similarities, but there are key differences that beginners need to understand. Both approaches involve training models to make predictions, but the way they process data and the complexity of the models are different.

\subsubsection{Data Processing}

In traditional machine learning, models often require feature engineering, which means that a human must manually design the features the model will use \cite{kuhn2013applied, mitchell1997machine}. For example, if you are trying to predict house prices, you might design features like the number of bedrooms, the size of the house, and the location. The model then uses these features to make predictions. 

On the other hand, deep learning models are capable of automatic feature extraction. Instead of manually selecting features, deep learning models can learn important patterns directly from raw data. For example, in a deep learning image recognition model, you can provide raw pixel values from an image, and the deep learning model will automatically learn to identify the relevant features, like edges, shapes, and objects \cite{zeiler2014visualizing}.

\subsubsection{Model Complexity}

Machine learning models typically involve a smaller number of parameters (e.g., a linear regression model might have only a few parameters that define the slope and intercept) \cite{domingos2012few}. In contrast, deep learning models involve millions or even billions of parameters, which allow them to learn much more complex patterns \cite{goodfellow2016deep,he2016deep}.

For example, consider the following models:

\begin{itemize}
    \item \textit{Machine Learning Model (Linear Regression):}
    \[
    y = w_1 x_1 + w_2 x_2 + b
    \]
    This is a simple linear model that predicts an outcome based on two input features, $x_1$ and $x_2$, where $w_1$ and $w_2$ are weights and $b$ is a bias term. The model is limited by its simplicity.

    \item \textit{Deep Learning Model (Neural Network):}
    \[
    \text{Layer 1: } z_1 = w_1 x_1 + w_2 x_2 + b
    \]
    \[
    \text{Layer 2: } z_2 = f(w_3 z_1 + w_4 z_2 + b)
    \]
    Here, the model consists of multiple layers, each with its own set of parameters, allowing the network to learn more complex relationships between inputs and outputs. The deeper the network, the more complex the patterns it can learn.
\end{itemize}

\subsubsection{Performance}

Deep learning models tend to outperform traditional machine learning models when working with large amounts of data \cite{sun2017revisiting}. Because of their depth, they can capture intricate patterns and dependencies that simpler models cannot. However, deep learning also requires much more computational power and data to train effectively.

In summary, the key differences between machine learning and deep learning are:
\begin{itemize}
    \item \textbf{Feature engineering}: ML models often require manual feature extraction, while DL models automatically learn features from raw data.
    \item \textbf{Model complexity}: ML models tend to be simpler with fewer parameters, while DL models are deeper and more complex.
    \item \textbf{Performance}: DL models excel at handling large datasets and complex tasks, but they require more computational resources.
\end{itemize}

\subsection{Key Concepts in Deep Learning}

Before diving deeper into the world of deep learning, it's essential to understand some key concepts that will form the foundation of any deep learning model.

\subsubsection{Neural Networks}

At the core of deep learning are \textit{neural networks}. A neural network is composed of layers of neurons, also called nodes. Each neuron takes in one or more inputs, applies a mathematical operation (often a weighted sum followed by an activation function), and produces an output. The outputs from one layer of neurons become the inputs to the next layer, allowing information to flow through the network \cite{lecun2015deep, nielsen2015neural}.

A simple neural network with one hidden layer can be represented as follows:

\begin{center}
\begin{tikzpicture}
  [node distance=2cm and 3cm, every node/.style={circle, draw, minimum size=1.2cm, inner sep=0pt}]
  
  \node (input1) at (0, 2) {Input 1};
  \node (input2) at (0, 0) {Input 2};
  
  \node (hidden1) at (3, 3) {Hidden};
  \node (hidden2) at (3, 1) {Hidden};
  \node (hidden3) at (3, -1) {Hidden};

  \node (output) at (6, 1) {Output};

  \draw[->] (input1) -- (hidden1);
  \draw[->] (input1) -- (hidden2);
  \draw[->] (input1) -- (hidden3);
  
  \draw[->] (input2) -- (hidden1);
  \draw[->] (input2) -- (hidden2);
  \draw[->] (input2) -- (hidden3);
  
  \draw[->] (hidden1) -- (output);
  \draw[->] (hidden2) -- (output);
  \draw[->] (hidden3) -- (output);

\end{tikzpicture}
\end{center}

\subsubsection{Layers}

Neural networks consist of different types of layers:
\begin{itemize}
    \item \textbf{Input Layer}: This is where the data enters the network. The number of neurons in the input layer corresponds to the number of features in the data.
    \item \textbf{Hidden Layers}: These layers perform computations and learn representations from the data. In deep learning, a network may have many hidden layers.
    \item \textbf{Output Layer}: This is where the network produces its final prediction. The number of neurons in the output layer depends on the task (e.g., 1 neuron for binary classification, multiple neurons for multi-class classification).
\end{itemize}

\subsubsection{Weights and Biases}

Each connection between neurons has an associated weight that determines the importance of the input. Weights are learned during training, and they represent the strength of the connection between two neurons. Additionally, each neuron often has a \textit{bias}, which allows the model to shift the output function to better fit the data.

\subsubsection{Activation Functions}

After computing the weighted sum of inputs, an \textit{activation function} is applied to introduce non-linearity into the model. Without activation functions, a neural network would essentially be a linear model, unable to learn complex patterns.

Common activation functions include:
\begin{itemize}
    \item \textbf{Sigmoid}: 
    \[
    \sigma(x) = \frac{1}{1 + e^{-x}}
    \]
    It outputs a value between 0 and 1, often used in the output layer for binary classification tasks.
    
    \item \textbf{ReLU (Rectified Linear Unit)}:
    \[
    f(x) = \max(0, x)
    \]
    It is widely used in hidden layers for its simplicity and effectiveness in dealing with vanishing gradient problems.

    \item \textbf{Tanh}: 
    \[
    \tanh(x) = \frac{e^x - e^{-x}}{e^x + e^{-x}}
    \]
    Outputs values between -1 and 1, used in some cases to normalize the output.
\end{itemize}

With these fundamental concepts in mind, let's explore how deep learning models are trained in the next chapter.

%% file: 12_nn.tex
\section{Neural Networks}

\subsection{What is a Neural Network?}
A neural network is a computational model inspired by the human brain. Just as the brain has neurons that communicate with each other, a neural network has artificial neurons (also called nodes) that are connected and pass information between layers \cite{hopfield1982neural,hopfield1984neurons}. Neural networks are designed to recognize patterns in data and can be used to solve complex problems, such as image recognition \cite{he2016deep}, speech processing \cite{hinton2012deep}, and language translation \cite{bahdanau2014neural}.

Neural networks are particularly powerful because they can learn from large amounts of data without being explicitly programmed to perform specific tasks. Instead, they adjust their internal parameters based on the input they receive and the errors they make during training. This learning process enables them to generalize and make predictions on new, unseen data \cite{lecun2015deep, nielsen2015neural}.

\subsection{Structure of a Neural Network}
A typical neural network consists of three main types of layers: the input layer, hidden layers, and the output layer. Each of these layers plays a unique role in transforming the data.

\subsubsection{Input Layer}
The input layer is the first layer in a neural network. Its primary purpose is to receive the raw data, such as images, text, or numerical data. Each neuron in the input layer represents one feature of the data. For example, in image recognition, each pixel of the image would be represented as a separate input.

Here's an example where we feed an image of 28x28 pixels (used in the MNIST dataset for digit recognition):

\begin{lstlisting}[style=python]
# Example of a simple neural network input layer for MNIST
input_shape = (28, 28)
\end{lstlisting}

The input layer doesn't modify the data; it simply passes it on to the next layer (the hidden layers) in the network.

\subsubsection{Hidden Layers}
Hidden layers are where the real computation happens in a neural network. Each hidden layer applies a series of transformations to the data by performing weighted sums of the inputs and passing the result through an activation function (discussed later).

Let's consider an example where we have 2 hidden layers:

\begin{lstlisting}[style=python]
# Example of defining hidden layers in a simple neural network
from tensorflow.keras.models import Sequential
from tensorflow.keras.layers import Dense

model = Sequential()
model.add(Dense(128, activation='relu', input_shape=(28, 28)))  # First hidden layer
model.add(Dense(64, activation='relu'))  # Second hidden layer
\end{lstlisting}

Each hidden layer can have many neurons, and the number of neurons in each layer defines the network's capacity to learn complex patterns. Typically, more neurons and more hidden layers enable the network to learn more complex functions, but they also increase the computational cost and the risk of overfitting.

\subsubsection{Output Layer}
The output layer is the final layer in a neural network. Its purpose is to provide the final prediction or output of the model. The number of neurons in the output layer depends on the type of task the network is designed to perform.

For example, in a binary classification problem (e.g., determining whether an email is spam or not), the output layer will have one neuron that outputs a value between 0 and 1, representing the probability of the input being classified as a particular class.

For a multi-class classification problem (e.g., classifying an image as a digit from 0 to 9), the output layer would have 10 neurons (one for each class). The output of these neurons will be passed through a function like softmax to generate probabilities for each class.

Here is how to define an output layer for a multi-class classification task:

\begin{lstlisting}[style=python]
# Defining the output layer for 10 classes (e.g., digit recognition)
model.add(Dense(10, activation='softmax'))
\end{lstlisting}

\subsection{Activation Functions}
An activation function is a mathematical function used in the neurons of the hidden and output layers. It determines whether the neuron should be activated (i.e., pass its signal to the next layer) based on the input. The choice of activation function affects the network's ability to learn and the type of patterns it can recognize.

\subsubsection{Sigmoid}
The sigmoid activation function \cite{glorot2010understanding} is commonly used in classification problems, particularly in the output layer for binary classification. It maps input values to a range between 0 and 1, making it useful for predicting probabilities.

The formula for the sigmoid function is:

\[
\sigma(x) = \frac{1}{1 + e^{-x}}
\]

Here's how it is implemented in Python:

\begin{lstlisting}[style=python]
import numpy as np

# Sigmoid activation function
def sigmoid(x):
    return 1 / (1 + np.exp(-x))
\end{lstlisting}

\subsubsection{ReLU (Rectified Linear Unit)}
ReLU \cite{glorot2011deep} is one of the most widely used activation functions in deep learning because it introduces non-linearity without affecting the performance of gradient-based optimization methods. ReLU outputs the input directly if it's positive, otherwise, it returns zero. This simplicity makes it computationally efficient.

The ReLU function is defined as:

\[
f(x) = \max(0, x)
\]

Here's how to implement ReLU in Python:

\begin{lstlisting}[style=python]
# ReLU activation function
def relu(x):
    return np.maximum(0, x)
\end{lstlisting}

\subsubsection{Tanh}
The Tanh (hyperbolic tangent) function is another activation function that outputs values between -1 and 1 \cite{lecun2002efficient}. It is similar to the sigmoid function but shifts its output range to be symmetric around zero. This can make learning more efficient in some cases because it ensures that the activations have zero mean, which can help in training deep networks.

The Tanh function is defined as:

\[
\tanh(x) = \frac{e^x - e^{-x}}{e^x + e^{-x}}
\]

Here's how you can implement Tanh in Python:

\begin{lstlisting}[style=python]
# Tanh activation function
def tanh(x):
    return np.tanh(x)
\end{lstlisting}

\subsection{Forward and Backward Propagation}
Forward propagation refers to the process of moving input data through the neural network to generate an output. In this process, the data passes through the input layer, hidden layers, and output layer, with each layer applying a transformation \cite{bishop2006pattern}. 

Backward propagation (or backpropagation) is the process by which the neural network learns from its errors \cite{rumelhart1986learning2}. During training, the network compares its output to the true label and calculates the error. It then adjusts the weights of the neurons to minimize this error, a process that is essential for the network to learn \cite{lecun2002efficient, rumelhart1986learning}.

\subsection{Loss Functions}
Loss functions are used to measure how far the predicted output is from the actual target value. They guide the training process by providing feedback to the network about how well it is performing.

\subsubsection{Mean Squared Error (MSE)}
Mean Squared Error (MSE) is commonly used for regression problems, where the goal is to predict continuous values. It calculates the average of the squared differences between the predicted values and the actual target values \cite{hastie2009elements}.

The formula for MSE is:

\[
\text{MSE} = \frac{1}{n} \sum_{i=1}^{n} (y_i - \hat{y}_i)^2
\]

Where \(y_i\) is the actual target value, and \(\hat{y}_i\) is the predicted value.

Here's how you can calculate MSE in Python:

\begin{lstlisting}[style=python]
from sklearn.metrics import mean_squared_error

# Example of calculating MSE
y_true = [3, -0.5, 2, 7]
y_pred = [2.5, 0.0, 2, 8]

mse = mean_squared_error(y_true, y_pred)
print(mse)
\end{lstlisting}

\subsubsection{Cross Entropy Loss}
Cross-entropy loss is used for classification problems. It measures the difference between the predicted probability distribution and the actual distribution (i.e., the true labels). The cross-entropy loss is minimized when the predicted probabilities match the actual distribution \cite{bishop2006pattern, shannon1948mathematical}.

The formula for binary cross-entropy is:

\[
\text{CE} = - \frac{1}{n} \sum_{i=1}^{n} \left( y_i \log(\hat{y}_i) + (1 - y_i) \log(1 - \hat{y}_i) \right)
\]

For multi-class problems, the loss is generalized to account for multiple classes. Here's how to calculate cross-entropy loss in Python:

\begin{lstlisting}[style=python]
from tensorflow.keras.losses import categorical_crossentropy

# Example of calculating cross-entropy loss
y_true = [[0, 1], [0, 0], [1, 0]]
y_pred = [[0.05, 0.95], [0.1, 0.9], [0.8, 0.2]]

loss = categorical_crossentropy(y_true, y_pred)
print(loss)
\end{lstlisting}

%% file: 13_opti.tex
\section{Optimization Algorithms}

\subsection{Gradient Descent}
Gradient descent \cite{cauchy1847methode} is an optimization algorithm used to minimize the loss function in machine learning models by iteratively adjusting the model parameters. The core idea behind gradient descent is to compute the direction (or gradient) in which the parameters should be adjusted in order to minimize the loss. The algorithm makes use of the derivative (or gradient) of the loss function with respect to the parameters and updates the parameters accordingly \cite{rumelhart1986learning}.

The general form of gradient descent is given by the following update rule for parameters \( \theta \):

\[
\theta = \theta - \alpha \cdot \nabla J(\theta)
\]

Where:
\begin{itemize}
    \item \( \theta \) represents the model parameters.
    \item \( \alpha \) is the learning rate, a small positive scalar that controls the size of the steps we take to reach the minimum.
    \item \( \nabla J(\theta) \) is the gradient of the loss function \( J(\theta) \) with respect to the parameters \( \theta \).
\end{itemize}

There are several variations of gradient descent, each suited to different types of problems and datasets.

\subsubsection{Batch Gradient Descent}
Batch Gradient Descent (BGD) \cite{bottou2010large} calculates the gradient of the loss function with respect to all the training examples in the dataset. This method is straightforward, but can be computationally expensive for large datasets.

\paragraph{Advantages:}
\begin{itemize}
    \item It provides a stable and accurate estimate of the gradient since it considers the entire dataset.
    \item Suitable for convex optimization problems where the cost function has a single global minimum.
\end{itemize}

\paragraph{Disadvantages:}
\begin{itemize}
    \item For large datasets, it is computationally expensive as it requires calculating gradients for the entire dataset before updating the parameters.
    \item It might be slow and memory-intensive for large datasets.
\end{itemize}

Example of Batch Gradient Descent implementation in Python:

\begin{lstlisting}[style=python]
import numpy as np

# Example of a simple cost function (mean squared error)
def compute_cost(X, y, theta):
    m = len(y)
    predictions = X.dot(theta)
    cost = (1/(2*m)) * np.sum(np.square(predictions - y))
    return cost

# Batch gradient descent implementation
def batch_gradient_descent(X, y, theta, alpha, iterations):
    m = len(y)
    cost_history = []
    
    for i in range(iterations):
        gradient = X.T.dot(X.dot(theta) - y) / m
        theta = theta - alpha * gradient
        cost_history.append(compute_cost(X, y, theta))
    
    return theta, cost_history
\end{lstlisting}

\subsubsection{Stochastic Gradient Descent (SGD)}
Stochastic Gradient Descent (SGD) \cite{bottou2010large} is a variant of gradient descent that updates the model parameters after each training example, rather than waiting for the entire dataset. This leads to faster iterations but introduces more noise in the gradient estimates \cite{ruder2016overview}.

\paragraph{Advantages:}
\begin{itemize}
    \item Faster and more memory-efficient for large datasets, as it doesn't require the entire dataset for each parameter update.
    \item Can help escape local minima due to the noise in the gradient updates.
\end{itemize}

\paragraph{Disadvantages:}
\begin{itemize}
    \item The noisy updates can lead to fluctuating convergence, making it difficult to settle on the global minimum.
    \item Often requires more iterations to converge compared to Batch Gradient Descent.
\end{itemize}

Example of Stochastic Gradient Descent in Python:

\begin{lstlisting}[style=python]
def stochastic_gradient_descent(X, y, theta, alpha, iterations):
    m = len(y)
    cost_history = []
    
    for i in range(iterations):
        for j in range(m):
            rand_index = np.random.randint(0, m)
            X_i = X[rand_index, :].reshape(1, X.shape[1])
            y_i = y[rand_index].reshape(1, 1)
            gradient = X_i.T.dot(X_i.dot(theta) - y_i)
            theta = theta - alpha * gradient
            cost_history.append(compute_cost(X, y, theta))
    
    return theta, cost_history
\end{lstlisting}

\subsubsection{Mini-batch Gradient Descent}
Mini-batch Gradient Descent combines the advantages of Batch Gradient Descent and Stochastic Gradient Descent. Instead of updating the parameters after every single training example (as in SGD) or after the entire dataset (as in BGD), it updates the parameters using small random subsets (or mini-batches) of the dataset \cite{bottou2010large,ruder2016overview}.

\paragraph{Advantages:}
\begin{itemize}
    \item More computationally efficient than Batch Gradient Descent, especially for large datasets.
    \item Reduces the noise seen in Stochastic Gradient Descent while still allowing faster iterations.
    \item Works well with GPU hardware, making it ideal for deep learning applications.
\end{itemize}

\paragraph{Disadvantages:}
\begin{itemize}
    \item Still requires tuning of the mini-batch size and learning rate.
    \item May oscillate around the minimum, but generally finds a good solution.
\end{itemize}

Example of Mini-batch Gradient Descent in Python:

\begin{lstlisting}[style=python]
def mini_batch_gradient_descent(X, y, theta, alpha, iterations, batch_size):
    m = len(y)
    cost_history = []
    
    for i in range(iterations):
        shuffled_indices = np.random.permutation(m)
        X_shuffled = X[shuffled_indices]
        y_shuffled = y[shuffled_indices]
        
        for j in range(0, m, batch_size):
            X_batch = X_shuffled[j:j + batch_size]
            y_batch = y_shuffled[j:j + batch_size]
            gradient = X_batch.T.dot(X_batch.dot(theta) - y_batch) / batch_size
            theta = theta - alpha * gradient
        
        cost_history.append(compute_cost(X, y, theta))
    
    return theta, cost_history
\end{lstlisting}

\subsection{Advanced Optimization Methods}
While gradient descent and its variants are simple and effective, there are more advanced optimization techniques that often improve performance, especially for deep learning applications.

\subsubsection{Adam Optimizer}
Adam (Adaptive Moment Estimation) \cite{kingma2014adam} is a popular optimization algorithm that combines the benefits of two other methods: Momentum \cite{polyak1964some} and RMSProp \cite{graves2013generating, tieleman2012lecture}. It computes adaptive learning rates for each parameter by keeping track of both the first moment (mean) and second moment (uncentered variance) of the gradients \cite{zou2019sufficient}.

The update rules for Adam are as follows:

\[
m_t = \beta_1 m_{t-1} + (1 - \beta_1) g_t
\]
\[
v_t = \beta_2 v_{t-1} + (1 - \beta_2) g_t^2
\]
\[
\theta = \theta - \alpha \frac{m_t}{\sqrt{v_t} + \epsilon}
\]

Where:
\begin{itemize}
    \item \( m_t \) and \( v_t \) are estimates of the first and second moments of the gradients.
    \item \( \beta_1 \) and \( \beta_2 \) are decay rates for the moment estimates (typically set to 0.9 and 0.999 respectively).
    \item \( \epsilon \) is a small constant to prevent division by zero.
\end{itemize}

Adam is widely used due to its fast convergence and robustness to noisy gradients.

\begin{lstlisting}[style=python]
import numpy as np

def adam_optimizer(X, y, theta, alpha, beta1, beta2, epsilon, iterations):
    m = len(y)
    m_t = np.zeros(theta.shape)
    v_t = np.zeros(theta.shape)
    cost_history = []
    
    for i in range(1, iterations + 1):
        gradient = X.T.dot(X.dot(theta) - y) / m
        m_t = beta1 * m_t + (1 - beta1) * gradient
        v_t = beta2 * v_t + (1 - beta2) * (gradient ** 2)
        
        m_hat = m_t / (1 - beta1 ** i)
        v_hat = v_t / (1 - beta2 ** i)
        
        theta = theta - alpha * (m_hat / (np.sqrt(v_hat) + epsilon))
        cost_history.append(compute_cost(X, y, theta))
    
    return theta, cost_history
\end{lstlisting}

\subsubsection{RMSProp}
RMSProp (Root Mean Square Propagation) \cite{tieleman2012lecture,graves2013generating} is designed to address the problem of varying learning rates in gradient descent by adjusting the learning rate for each parameter individually. It works by maintaining a moving average of the squared gradients.

The update rule for RMSProp is:

\[
v_t = \beta v_{t-1} + (1 - \beta) g_t^2
\]
\[
\theta = \theta - \frac{\alpha}{\sqrt{v_t + \epsilon}} g_t
\]

Where:
\begin{itemize}
    \item \( v_t \) is the exponentially weighted moving average of the squared gradients.
    \item \( \epsilon \) is a small constant to prevent division by zero.
\end{itemize}

RMSProp is particularly useful in deep learning models with non-convex loss functions.

Example of RMSProp in Python:

\begin{lstlisting}[style=python]
def rmsprop_optimizer(X, y, theta, alpha, beta, epsilon, iterations):
    m = len(y)
    v_t = np.zeros(theta.shape)
    cost_history = []
    
    for i in range(iterations):
        gradient = X.T.dot(X.dot(theta) - y) / m
        v_t = beta * v_t + (1 - beta) * (gradient ** 2)
        theta = theta - alpha * gradient / (np.sqrt(v_t) + epsilon)
        cost_history.append(compute_cost(X, y, theta))
    
    return theta, cost_history
\end{lstlisting}

%% file: 14_tf_torch.tex
\section{Deep Learning Frameworks}

\subsection{Introduction to Popular Frameworks}
Deep learning frameworks provide powerful tools for building and training machine learning models, especially neural networks. These frameworks simplify complex mathematical operations, manage data, and optimize training processes, making it easier for developers to build deep learning models. In this section, we will introduce three of the most widely used frameworks: TensorFlow \cite{tensorflow2015-whitepaper}, PyTorch \cite{paszke2019pytorch}, and Keras \cite{chollet2015keras}. Each of these frameworks has its own strengths, which we will explain in detail along with practical examples to help beginners understand their key features.

\subsubsection{TensorFlow}
TensorFlow is a highly popular deep learning framework developed by Google Brain. It is known for its scalability and flexibility, supporting both high-level and low-level APIs. TensorFlow can be used in a wide range of applications, from mobile devices to large-scale distributed systems.

One of the most important features of TensorFlow is its ability to run on different platforms (CPUs, GPUs, and TPUs). This feature allows TensorFlow to handle complex computations efficiently. Additionally, TensorFlow has a robust ecosystem that includes TensorBoard for visualizing model metrics, TensorFlow Serving for deploying models, and TensorFlow Lite for running models on mobile devices.

Here's a simple example of how to build and train a neural network using TensorFlow:

\begin{lstlisting}[style=python]
import tensorflow as tf
from tensorflow.keras import layers

# Define a simple sequential model
model = tf.keras.Sequential([
    layers.Dense(64, activation='relu', input_shape=(784,)),
    layers.Dense(10, activation='softmax')
])

# Compile the model
model.compile(optimizer='adam', 
              loss='sparse_categorical_crossentropy', 
              metrics=['accuracy'])

# Train the model on sample data
model.fit(x_train, y_train, epochs=5)
\end{lstlisting}

In this example, we define a simple neural network using TensorFlow's high-level Keras API. The model is a sequential neural network with two layers: one hidden layer with 64 neurons and a ReLU activation function, and an output layer with 10 neurons and a softmax activation function. The model is then compiled using the Adam optimizer and trained using the \texttt{fit} method on training data.

\subsubsection{PyTorch}
PyTorch, developed by Facebook's AI Research lab (FAIR), is another widely-used deep learning framework. Unlike TensorFlow, PyTorch is often preferred by researchers due to its dynamic computation graph (also known as eager execution). This feature makes it easier to debug and modify the network architecture during runtime, making it a great choice for experimentation and research.

PyTorch emphasizes simplicity and flexibility. Its syntax is similar to standard Python, making it easy to learn and use, especially for those who are familiar with Python programming. PyTorch also supports both CPU and GPU computations.

Below is an example of how to create and train a simple neural network using PyTorch:

\begin{lstlisting}[style=python]
import torch
import torch.nn as nn
import torch.optim as optim

# Define the neural network
class SimpleNN(nn.Module):
    def __init__(self):
        super(SimpleNN, self).__init__()
        self.fc1 = nn.Linear(784, 64)
        self.fc2 = nn.Linear(64, 10)
    
    def forward(self, x):
        x = torch.relu(self.fc1(x))
        x = torch.softmax(self.fc2(x), dim=1)
        return x

# Instantiate the model, define loss function and optimizer
model = SimpleNN()
criterion = nn.CrossEntropyLoss()
optimizer = optim.Adam(model.parameters())

# Training loop
for epoch in range(5):
    optimizer.zero_grad()
    output = model(x_train)
    loss = criterion(output, y_train)
    loss.backward()
    optimizer.step()
\end{lstlisting}

In this PyTorch example, we define a simple neural network class called \texttt{SimpleNN}. It has two fully connected layers: one with 64 neurons and ReLU activation, and another with 10 neurons and a softmax output. The training loop includes manually performing gradient computation, backpropagation, and optimization steps.

\subsubsection{Keras}
Keras is a high-level deep learning API that simplifies building and training neural networks. Initially developed as a standalone library, it is now integrated with TensorFlow as its default high-level API. Keras focuses on being user-friendly and accessible, especially for beginners. It abstracts many of the complexities involved in building deep learning models, allowing users to focus on high-level design.

Keras is particularly useful for prototyping, as it provides a simple and consistent interface. Models in Keras can be constructed using the Sequential API or the Functional API. Here's an example of building and training a neural network using Keras (which works seamlessly with TensorFlow):

\begin{lstlisting}[style=python]
from tensorflow.keras.models import Sequential
from tensorflow.keras.layers import Dense

# Define a Sequential model
model = Sequential()
model.add(Dense(64, activation='relu', input_shape=(784,)))
model.add(Dense(10, activation='softmax'))

# Compile the model
model.compile(optimizer='adam', 
              loss='categorical_crossentropy', 
              metrics=['accuracy'])

# Train the model
model.fit(x_train, y_train, epochs=5)
\end{lstlisting}

In this example, the code is similar to the TensorFlow example because Keras operates on top of TensorFlow. We define a simple sequential model with two layers, compile it using the Adam optimizer, and train it using the \texttt{fit} method.

Keras is perfect for beginners due to its simplicity and ease of use, making it a great starting point for anyone new to deep learning. However, Keras is still powerful enough to be used for many complex models in practice, especially with its support for TensorFlow's underlying functionality.

\subsection{Comparison of Frameworks}
Each of these deep learning frameworks-TensorFlow, PyTorch, and Keras-has its strengths. Here's a quick comparison:

\begin{itemize}
    \item \textbf{TensorFlow}: Best suited for production environments and large-scale applications. Its deployment tools (e.g., TensorFlow Lite, TensorFlow Serving) and support for distributed training make it popular in industry.
    \item \textbf{PyTorch}: Popular in academic research due to its flexibility, dynamic computation graph, and ease of debugging. It has a strong community of researchers and developers.
    \item \textbf{Keras}: Ideal for beginners, due to its simplicity and high-level abstractions. It's also integrated with TensorFlow, providing a seamless experience from prototyping to deployment.
\end{itemize}

In summary, TensorFlow, PyTorch, and Keras each cater to different use cases and user preferences, allowing developers and researchers to choose the one that best fits their needs.

%% file: 15_cnn.tex
\section{Convolutional Neural Networks (CNNs)}

\subsection{What is a CNN?}
A Convolutional Neural Network (CNN) is a specialized type of deep neural network designed primarily for image processing tasks \cite{fukushima1980neocognitron,lecun1998gradient}. Unlike traditional neural networks, CNNs are specifically tailored to process structured data like images by taking advantage of their spatial structure. CNNs automatically learn the spatial hierarchies of features from input images, which makes them particularly effective in tasks such as image classification \cite{lecun1998gradient}, object detection \cite{girshick2014rich}, and segmentation \cite{long2015fully}.

The key idea behind CNNs is to apply convolution operations, which are mathematical operations that detect patterns like edges, textures, and shapes, from simple to complex as you go deeper into the network. These detected features are then used to classify or recognize objects in the image.

\subsection{Convolution Layer}
The convolution layer is the heart of a CNN. It performs the convolution operation, where a small filter (also known as a kernel) slides over the input image, computing dot products between the filter and the overlapping region of the input. This operation allows the network to detect specific features like edges, corners, or textures at different locations in the image.

\begin{center}
\begin{tikzpicture}
\end{tikzpicture}
\end{center}

For example, a $3 \times 3$ filter applied to a grayscale image would slide across the image, covering $3 \times 3$ pixel regions at a time, and calculate the dot product between the filter and the image region. This produces a new "feature map" that highlights specific features detected by the filter.

Let's look at an example using Python code to perform a simple 2D convolution using the \texttt{numpy} library.

\begin{lstlisting}[style=python]
import numpy as np

# Define a 5x5 image
image = np.array([
    [1, 2, 0, 3, 1],
    [4, 6, 2, 8, 5],
    [7, 8, 1, 9, 6],
    [1, 3, 5, 2, 1],
    [0, 6, 2, 1, 4]
])

# Define a 3x3 filter (kernel)
filter = np.array([
    [1, 0, -1],
    [1, 0, -1],
    [1, 0, -1]
])

# Convolution operation (without padding and stride 1)
output = np.zeros((3, 3))

for i in range(3):
    for j in range(3):
        output[i, j] = np.sum(image[i:i+3, j:j+3] * filter)

print(output)
\end{lstlisting}

In this example, we manually implement a convolution operation where a $3 \times 3$ filter slides over a $5 \times 5$ image, detecting edges. In practice, libraries like \texttt{TensorFlow} or \texttt{PyTorch} provide efficient implementations for these operations.

\subsection{Pooling Layer}
The pooling layer is used to reduce the spatial dimensions (width and height) of the feature maps while retaining the most important information \cite{scherer2010evaluation}. This not only reduces the computational cost but also helps prevent overfitting by down-sampling the features. The two most commonly used pooling techniques are max pooling and average pooling.

\subsubsection{Max Pooling}
Max pooling takes the maximum value from a specified region (usually $2 \times 2$) of the input feature map. This operation helps in keeping the most prominent features while reducing the size of the feature map.

For example, consider the following $4 \times 4$ feature map, and we apply a $2 \times 2$ max pooling operation:

\begin{lstlisting}[style=python]
import numpy as np

# Define a 4x4 feature map
feature_map = np.array([
    [1, 3, 2, 4],
    [5, 6, 7, 8],
    [9, 10, 11, 12],
    [13, 14, 15, 16]
])

# Apply max pooling with a 2x2 window
pooled_output = np.zeros((2, 2))

for i in range(2):
    for j in range(2):
        pooled_output[i, j] = np.max(feature_map[i*2:i*2+2, j*2:j*2+2])

print(pooled_output)
\end{lstlisting}

The output will be:
\begin{lstlisting}[style=cmd]
[[ 6  8]
 [14 16]]
\end{lstlisting}

As you can see, max pooling retains the most significant values, making it effective for feature selection.

\subsubsection{Average Pooling}
Average pooling, on the other hand, computes the average of the values in a specified region. While max pooling focuses on the most prominent feature, average pooling takes into account the overall presence of features.

Here's how we can implement average pooling:

\begin{lstlisting}[style=python]
# Apply average pooling with a 2x2 window
avg_pooled_output = np.zeros((2, 2))

for i in range(2):
    for j in range(2):
        avg_pooled_output[i, j] = np.mean(feature_map[i*2:i*2+2, j*2:j*2+2])

print(avg_pooled_output)
\end{lstlisting}

The output will be:
\begin{lstlisting}[style=cmd]
[[ 3.75  5.75]
 [11.25 13.25]]
\end{lstlisting}

Average pooling tends to be used in scenarios where it's important to capture the overall features, rather than focusing on the strongest activation.

\subsection{Flattening and Fully Connected Layer}
After the convolution and pooling layers, the feature maps are typically flattened into a one-dimensional vector. This flattened vector is then passed to one or more fully connected layers, which perform the actual classification or regression task.

Flattening simply converts the 2D matrix of features into a 1D vector:

\begin{lstlisting}[style=python]
# Flatten the pooled output
flattened_output = avg_pooled_output.flatten()
print(flattened_output)
\end{lstlisting}

The flattened output can then be passed to a fully connected layer for classification:

\begin{lstlisting}[style=python]
# Example of a simple fully connected layer operation
weights = np.random.rand(4)  # Random weights for 4 inputs
bias = 1.0
fc_output = np.dot(flattened_output, weights) + bias
print(fc_output)
\end{lstlisting}

Fully connected layers work the same as in traditional neural networks, where the input is multiplied by weights and added to biases, followed by an activation function.

The following is a complete process of building, training, and testing a Convolutional Neural Network (CNN) classifier for the MNIST dataset \cite{lecun1998mnist}. The process includes loading the data, constructing the CNN architecture, training the model, and evaluating its performance on the test set.

\begin{lstlisting}[style=python]
# Import necessary libraries
import tensorflow as tf
from tensorflow.keras import layers, models
from tensorflow.keras.datasets import mnist
from tensorflow.keras.utils import to_categorical
import matplotlib.pyplot as plt

# Load and preprocess the MNIST dataset
(train_images, train_labels), (test_images, test_labels) = mnist.load_data()

# Reshape the images to add a single channel (grayscale)
train_images = train_images.reshape((60000, 28, 28, 1))
test_images = test_images.reshape((10000, 28, 28, 1))

# Normalize the pixel values to a range of 0 to 1
train_images = train_images.astype('float32') / 255
test_images = test_images.astype('float32') / 255

# One-hot encode the labels
train_labels = to_categorical(train_labels)
test_labels = to_categorical(test_labels)

# Construct the CNN model
model = models.Sequential()

# First convolutional layer
model.add(layers.Conv2D(32, (3, 3), activation='relu', input_shape=(28, 28, 1)))
model.add(layers.MaxPooling2D((2, 2)))

# Second convolutional layer
model.add(layers.Conv2D(64, (3, 3), activation='relu'))
model.add(layers.MaxPooling2D((2, 2)))

# Third convolutional layer
model.add(layers.Conv2D(64, (3, 3), activation='relu'))

# Flatten the output to feed into a fully connected layer
model.add(layers.Flatten())

# Fully connected (dense) layer
model.add(layers.Dense(64, activation='relu'))

# Output layer with softmax activation for classification
model.add(layers.Dense(10, activation='softmax'))

# Print the model summary
model.summary()

# Compile the model
model.compile(optimizer='adam',
              loss='categorical_crossentropy',
              metrics=['accuracy'])

# Train the model
history = model.fit(train_images, train_labels, epochs=5, batch_size=64, validation_split=0.2)

# Evaluate the model on the test data
test_loss, test_acc = model.evaluate(test_images, test_labels)

print(f'Test accuracy: {test_acc}')

# Plot training and validation accuracy
plt.plot(history.history['accuracy'], label='Training Accuracy')
plt.plot(history.history['val_accuracy'], label='Validation Accuracy')
plt.title('Training and Validation Accuracy')
plt.xlabel('Epoch')
plt.ylabel('Accuracy')
plt.legend()
plt.show()

# Plot training and validation loss
plt.plot(history.history['loss'], label='Training Loss')
plt.plot(history.history['val_loss'], label='Validation Loss')
plt.title('Training and Validation Loss')
plt.xlabel('Epoch')
plt.ylabel('Loss')
plt.legend()
plt.show()
\end{lstlisting}

\textbf{Explanation:}

\begin{enumerate}
    \item The dataset used in this CNN classifier is MNIST, a collection of 28x28 grayscale images of handwritten digits (0-9). We first load the dataset using \texttt{mnist.load\_data()} from the \\ \texttt{tensorflow.keras.datasets} module. The data is split into training and testing sets.
    
    \item Each image is reshaped to include a single channel (grayscale), and pixel values are normalized to a range between 0 and 1 to help the model converge faster.
    
    \item The labels are one-hot encoded using \texttt{to\_categorical} so that they can be used in a multi-class classification problem.
    
    \item The CNN architecture is constructed using the \texttt{Sequential} API from Keras:
    \begin{itemize}
        \item The first convolutional layer uses 32 filters with a 3x3 kernel, followed by max pooling to reduce the spatial dimensions.
        \item A second convolutional layer with 64 filters is added, again followed by max pooling.
        \item A third convolutional layer with 64 filters is added.
        \item The output from the convolutional layers is flattened and fed into a fully connected layer with 64 units.
        \item The final layer has 10 units with softmax activation for classifying the 10 digits (0-9).
    \end{itemize}
    
    \item The model is compiled using the Adam optimizer, categorical cross-entropy loss, and accuracy as the evaluation metric.
    
    \item The model is trained for 5 epochs, and during training, 20\% of the training data is used as validation data.
    
    \item After training, the model is evaluated on the test set, and both accuracy and loss plots are displayed to visualize the training process.
\end{enumerate}

\subsection{CNN Architectures}
Several CNN architectures have been developed over the years, each improving upon the last in terms of accuracy, depth, and computational efficiency. Let's discuss some of the most popular architectures used today.

\subsubsection{LeNet}
LeNet \cite{lecun1998gradient} is one of the earliest CNN architectures, introduced by Yann LeCun for digit recognition tasks, such as the MNIST handwritten digit dataset. It consists of two convolutional layers followed by two fully connected layers. Despite its simplicity, LeNet laid the groundwork for modern deep learning architectures.

\begin{center}
\begin{tikzpicture}[transform shape]

\node (input) [draw, minimum width=3cm, minimum height=1cm] at (0, 0) {Input Layer: 32x32x1};

\node (cnn1) [draw, minimum width=3cm, minimum height=1cm] at (0, -1.5) {Convolutional Layer 1 (C1): 6x28x28};

\node (pool1) [draw, minimum width=3cm, minimum height=1cm] at (0, -3) {Subsampling Layer 1 (S2): 6x14x14};

\node (cnn2) [draw, minimum width=3cm, minimum height=1cm] at (0, -4.5) {Convolutional Layer 2 (C3): 16x10x10};

\node (pool2) [draw, minimum width=3cm, minimum height=1cm] at (0, -6) {Subsampling Layer 2 (S4): 16x5x5};

\node (cnn3) [draw, minimum width=3cm, minimum height=1cm] at (0, -7.5) {Convolutional Layer 3 (C5): 120 units};

\node (fc1) [draw, minimum width=3cm, minimum height=1cm] at (0, -9) {Fully Connected Layer (F6): 84 units};

\node (output) [draw, minimum width=3cm, minimum height=1cm] at (0, -10.5) {Output Layer: 10 classes};

\draw[->] (input) -- (cnn1);
\draw[->] (cnn1) -- (pool1);
\draw[->] (pool1) -- (cnn2);
\draw[->] (cnn2) -- (pool2);
\draw[->] (pool2) -- (cnn3);
\draw[->] (cnn3) -- (fc1);
\draw[->] (fc1) -- (output);

\end{tikzpicture}
\end{center}

The following is the implementation of the LeNet model, which is used for image classification tasks. LeNet is a classic convolutional neural network architecture originally applied to handwritten digit recognition.

\begin{lstlisting}[style=python]
import torch
import torch.nn as nn
import torch.nn.functional as F

class LeNet(nn.Module):
    def __init__(self):
        super(LeNet, self).__init__()
        # First convolutional layer: input channel 1, output channel 6, kernel size 5x5
        self.conv1 = nn.Conv2d(1, 6, kernel_size=5)
        # Second convolutional layer: input channel 6, output channel 16, kernel size 5x5
        self.conv2 = nn.Conv2d(6, 16, kernel_size=5)
        # Fully connected layer: input size 16*5*5, output size 120
        self.fc1 = nn.Linear(16*5*5, 120)
        # Fully connected layer: input size 120, output size 84
        self.fc2 = nn.Linear(120, 84)
        # Output layer: input size 84, output size 10 (number of classes)
        self.fc3 = nn.Linear(84, 10)

    def forward(self, x):
        # Convolutional layer 1 + ReLU activation + Max Pooling
        x = F.max_pool2d(F.relu(self.conv1(x)), kernel_size=2)
        # Convolutional layer 2 + ReLU activation + Max Pooling
        x = F.max_pool2d(F.relu(self.conv2(x)), kernel_size=2)
        # Flatten the output from the convolutional layers
        x = x.view(-1, 16*5*5)
        # Fully connected layer 1 + ReLU activation
        x = F.relu(self.fc1(x))
        # Fully connected layer 2 + ReLU activation
        x = F.relu(self.fc2(x))
        # Output layer
        x = self.fc3(x)
        return x

# Test LeNet network
net = LeNet()
print(net)
\end{lstlisting}

Explanation of the code:

\begin{itemize}
    \item \texttt{conv1} and \texttt{conv2} are convolutional layers designed to extract local features from images. The first layer converts the input image with one channel (such as grayscale images) into six feature maps, and the second layer transforms those six maps into sixteen.
    \item \texttt{max\_pool2d} performs max pooling, reducing the size of the feature maps while retaining the most important features. The pooling is applied with a 2x2 window.
    \item \texttt{fc1}, \texttt{fc2}, and \texttt{fc3} are fully connected layers that map the extracted features to the classification output (e.g., classifying handwritten digits from 0 to 9).
    \item \texttt{relu} introduces non-linearity, enabling the model to learn complex patterns.
    \item \texttt{view} flattens the feature maps into a vector to pass them into the fully connected layers.
    \item The \texttt{forward} method outlines the forward pass of the network, and the \texttt{fc3} outputs the final classification predictions.
\end{itemize}

\subsubsection{AlexNet}
AlexNet \cite{krizhevsky2012imagenet}, developed by Alex Krizhevsky, revolutionized the field of image classification by winning the ImageNet competition in 2012. It consists of five convolutional layers, max pooling, and three fully connected layers. AlexNet introduced techniques like ReLU activations and dropout for improved training of deep networks.

\begin{center}
\begin{tikzpicture}[transform shape]

\node (input) [draw, minimum width=4cm, minimum height=1cm] at (0, 0) {Input Layer: 227x227x3};

\node (cnn1) [draw, minimum width=4cm, minimum height=1cm] at (0, -1.5) {Convolutional Layer 1: 96x55x55};

\node (pool1) [draw, minimum width=4cm, minimum height=1cm] at (0, -3) {Max Pooling 1: 96x27x27};

\node (cnn2) [draw, minimum width=4cm, minimum height=1cm] at (0, -4.5) {Convolutional Layer 2: 256x27x27};

\node (pool2) [draw, minimum width=4cm, minimum height=1cm] at (0, -6) {Max Pooling 2: 256x13x13};

\node (cnn3) [draw, minimum width=4cm, minimum height=1cm] at (0, -7.5) {Convolutional Layer 3: 384x13x13};

\node (cnn4) [draw, minimum width=4cm, minimum height=1cm] at (0, -9) {Convolutional Layer 4: 384x13x13};

\node (cnn5) [draw, minimum width=4cm, minimum height=1cm] at (0, -10.5) {Convolutional Layer 5: 256x13x13};

\node (pool3) [draw, minimum width=4cm, minimum height=1cm] at (0, -12) {Max Pooling 3: 256x6x6};

\node (fc1) [draw, minimum width=4cm, minimum height=1cm] at (0, -13.5) {Fully Connected Layer 1: 4096 units};

\node (fc2) [draw, minimum width=4cm, minimum height=1cm] at (0, -15) {Fully Connected Layer 2: 4096 units};

\node (output) [draw, minimum width=4cm, minimum height=1cm] at (0, -16.5) {Output Layer: 1000 classes};

\draw[->] (input) -- (cnn1);
\draw[->] (cnn1) -- (pool1);
\draw[->] (pool1) -- (cnn2);
\draw[->] (cnn2) -- (pool2);
\draw[->] (pool2) -- (cnn3);
\draw[->] (cnn3) -- (cnn4);
\draw[->] (cnn4) -- (cnn5);
\draw[->] (cnn5) -- (pool3);
\draw[->] (pool3) -- (fc1);
\draw[->] (fc1) -- (fc2);
\draw[->] (fc2) -- (output);

\end{tikzpicture}
\end{center}

The following is the implementation of the AlexNet model, which is used for image classification tasks. AlexNet is a deep convolutional neural network architecture that won the ImageNet Large Scale Visual Recognition Challenge in 2012.

\begin{lstlisting}[style=python]
import torch
import torch.nn as nn
import torch.nn.functional as F

class AlexNet(nn.Module):
    def __init__(self):
        super(AlexNet, self).__init__()
        # First convolutional layer: input channel 3 (RGB), output channel 64, kernel size 11x11, stride 4
        self.conv1 = nn.Conv2d(3, 64, kernel_size=11, stride=4, padding=2)
        # Second convolutional layer: input channel 64, output channel 192, kernel size 5x5
        self.conv2 = nn.Conv2d(64, 192, kernel_size=5, padding=2)
        # Third convolutional layer: input channel 192, output channel 384, kernel size 3x3
        self.conv3 = nn.Conv2d(192, 384, kernel_size=3, padding=1)
        # Fourth convolutional layer: input channel 384, output channel 256, kernel size 3x3
        self.conv4 = nn.Conv2d(384, 256, kernel_size=3, padding=1)
        # Fifth convolutional layer: input channel 256, output channel 256, kernel size 3x3
        self.conv5 = nn.Conv2d(256, 256, kernel_size=3, padding=1)
        # Fully connected layer: input size 256 * 6 * 6, output size 4096
        self.fc1 = nn.Linear(256 * 6 * 6, 4096)
        # Fully connected layer: input size 4096, output size 4096
        self.fc2 = nn.Linear(4096, 4096)
        # Output layer: input size 4096, output size 1000 (number of classes in ImageNet)
        self.fc3 = nn.Linear(4096, 1000)

    def forward(self, x):
        # Convolutional layer 1 + ReLU activation + Max Pooling
        x = F.max_pool2d(F.relu(self.conv1(x)), kernel_size=3, stride=2)
        # Convolutional layer 2 + ReLU activation + Max Pooling
        x = F.max_pool2d(F.relu(self.conv2(x)), kernel_size=3, stride=2)
        # Convolutional layer 3 + ReLU activation
        x = F.relu(self.conv3(x))
        # Convolutional layer 4 + ReLU activation
        x = F.relu(self.conv4(x))
        # Convolutional layer 5 + ReLU activation + Max Pooling
        x = F.max_pool2d(F.relu(self.conv5(x)), kernel_size=3, stride=2)
        # Flatten the output from the convolutional layers
        x = x.view(-1, 256 * 6 * 6)
        # Fully connected layer 1 + ReLU activation + Dropout
        x = F.dropout(F.relu(self.fc1(x)), p=0.5, training=self.training)
        # Fully connected layer 2 + ReLU activation + Dropout
        x = F.dropout(F.relu(self.fc2(x)), p=0.5, training=self.training)
        # Output layer
        x = self.fc3(x)
        return x

# Test AlexNet network
net = AlexNet()
print(net)
\end{lstlisting}

Explanation of the code:

\begin{itemize}
    \item \texttt{conv1}, \texttt{conv2}, \texttt{conv3}, \texttt{conv4}, and \texttt{conv5} are convolutional layers that extract features from the input images. AlexNet uses multiple layers to learn complex representations.
    \item \texttt{max\_pool2d} is used to downsample the feature maps after certain convolutional layers, reducing their spatial dimensions while keeping important features.
    \item \texttt{fc1}, \texttt{fc2}, and \texttt{fc3} are fully connected layers that map the high-level features to the output space, where \texttt{fc3} produces the final prediction for 1000 classes in the ImageNet dataset.
    \item \texttt{relu} is the activation function used to introduce non-linearity in the model.
    \item \texttt{dropout} is used to prevent overfitting by randomly setting a fraction of the neurons to zero during training.
    \item The \texttt{forward} function defines the flow of data through the network, starting with convolutional layers, followed by fully connected layers, and finally outputting the class predictions.
\end{itemize}

\subsubsection{VGGNet}
VGGNet \cite{simonyan2014very}, developed by the Visual Geometry Group, is known for its simplicity and depth. It uses very small $3 \times 3$ convolution filters and stacks them deeper into the network. VGGNet is widely used for transfer learning in various computer vision tasks.

\begin{center}
\begin{tikzpicture}[transform shape]

\node (input) [draw, minimum width=4cm, minimum height=1cm] at (0, 0) {Input Layer: 224x224x3};

\node (conv1_1) [draw, minimum width=4cm, minimum height=1cm] at (0, -1.5) {Conv Layer 1-1: 64x224x224};
\node (conv1_2) [draw, minimum width=4cm, minimum height=1cm] at (0, -3) {Conv Layer 1-2: 64x224x224};
\node (pool1) [draw, minimum width=4cm, minimum height=1cm] at (0, -4.5) {Max Pooling 1: 64x112x112};

\node (conv2_1) [draw, minimum width=4cm, minimum height=1cm] at (0, -6) {Conv Layer 2-1: 128x112x112};
\node (conv2_2) [draw, minimum width=4cm, minimum height=1cm] at (0, -7.5) {Conv Layer 2-2: 128x112x112};
\node (pool2) [draw, minimum width=4cm, minimum height=1cm] at (0, -9) {Max Pooling 2: 128x56x56};

\node (conv3_1) [draw, minimum width=4cm, minimum height=1cm] at (0, -10.5) {Conv Layer 3-1: 256x56x56};
\node (conv3_2) [draw, minimum width=4cm, minimum height=1cm] at (0, -12) {Conv Layer 3-2: 256x56x56};
\node (conv3_3) [draw, minimum width=4cm, minimum height=1cm] at (0, -13.5) {Conv Layer 3-3: 256x56x56};
\node (pool3) [draw, minimum width=4cm, minimum height=1cm] at (0, -15) {Max Pooling 3: 256x28x28};

\node (conv4_1) [draw, minimum width=4cm, minimum height=1cm] at (7, -1.5) {Conv Layer 4-1: 512x28x28};
\node (conv4_2) [draw, minimum width=4cm, minimum height=1cm] at (7, -3) {Conv Layer 4-2: 512x28x28};
\node (conv4_3) [draw, minimum width=4cm, minimum height=1cm] at (7, -4.5) {Conv Layer 4-3: 512x28x28};
\node (pool4) [draw, minimum width=4cm, minimum height=1cm] at (7, -6) {Max Pooling 4: 512x14x14};

\node (conv5_1) [draw, minimum width=4cm, minimum height=1cm] at (7, -7.5) {Conv Layer 5-1: 512x14x14};
\node (conv5_2) [draw, minimum width=4cm, minimum height=1cm] at (7, -9) {Conv Layer 5-2: 512x14x14};
\node (conv5_3) [draw, minimum width=4cm, minimum height=1cm] at (7, -10.5) {Conv Layer 5-3: 512x14x14};
\node (pool5) [draw, minimum width=4cm, minimum height=1cm] at (7, -12) {Max Pooling 5: 512x7x7};

\node (fc1) [draw, minimum width=4cm, minimum height=1cm] at (7, -13.5) {Fully Connected Layer 1: 4096 units};
\node (fc2) [draw, minimum width=4cm, minimum height=1cm] at (7, -15) {Fully Connected Layer 2: 4096 units};
\node (output) [draw, minimum width=4cm, minimum height=1cm] at (7, -16.5) {Output Layer: 1000 classes};

\draw[->] (input) -- (conv1_1);
\draw[->] (conv1_1) -- (conv1_2);
\draw[->] (conv1_2) -- (pool1);
\draw[->] (pool1) -- (conv2_1);
\draw[->] (conv2_1) -- (conv2_2);
\draw[->] (conv2_2) -- (pool2);
\draw[->] (pool2) -- (conv3_1);
\draw[->] (conv3_1) -- (conv3_2);
\draw[->] (conv3_2) -- (conv3_3);
\draw[->] (conv3_3) -- (pool3);

\draw[->] (pool3.east) -- ++(1,0) |- (conv4_1.west);
\draw[->] (conv4_1) -- (conv4_2);
\draw[->] (conv4_2) -- (conv4_3);
\draw[->] (conv4_3) -- (pool4);
\draw[->] (pool4) -- (conv5_1);
\draw[->] (conv5_1) -- (conv5_2);
\draw[->] (conv5_2) -- (conv5_3);
\draw[->] (conv5_3) -- (pool5);
\draw[->] (pool5) -- (fc1);
\draw[->] (fc1) -- (fc2);
\draw[->] (fc2) -- (output);

\end{tikzpicture}
\end{center}

The following is the implementation of the VGGNet-16 model, which is used for image classification tasks. VGGNet is a deep convolutional neural network that focuses on using smaller 3x3 convolutional kernels stacked in multiple layers.

\begin{lstlisting}[style=python]
import torch
import torch.nn as nn
import torch.nn.functional as F

class VGG16(nn.Module):
    def __init__(self):
        super(VGG16, self).__init__()
        # Block 1: Two convolutional layers, followed by max pooling
        self.conv1_1 = nn.Conv2d(3, 64, kernel_size=3, padding=1)
        self.conv1_2 = nn.Conv2d(64, 64, kernel_size=3, padding=1)
        self.pool1 = nn.MaxPool2d(kernel_size=2, stride=2)

        # Block 2: Two convolutional layers, followed by max pooling
        self.conv2_1 = nn.Conv2d(64, 128, kernel_size=3, padding=1)
        self.conv2_2 = nn.Conv2d(128, 128, kernel_size=3, padding=1)
        self.pool2 = nn.MaxPool2d(kernel_size=2, stride=2)

        # Block 3: Three convolutional layers, followed by max pooling
        self.conv3_1 = nn.Conv2d(128, 256, kernel_size=3, padding=1)
        self.conv3_2 = nn.Conv2d(256, 256, kernel_size=3, padding=1)
        self.conv3_3 = nn.Conv2d(256, 256, kernel_size=3, padding=1)
        self.pool3 = nn.MaxPool2d(kernel_size=2, stride=2)

        # Block 4: Three convolutional layers, followed by max pooling
        self.conv4_1 = nn.Conv2d(256, 512, kernel_size=3, padding=1)
        self.conv4_2 = nn.Conv2d(512, 512, kernel_size=3, padding=1)
        self.conv4_3 = nn.Conv2d(512, 512, kernel_size=3, padding=1)
        self.pool4 = nn.MaxPool2d(kernel_size=2, stride=2)

        # Block 5: Three convolutional layers, followed by max pooling
        self.conv5_1 = nn.Conv2d(512, 512, kernel_size=3, padding=1)
        self.conv5_2 = nn.Conv2d(512, 512, kernel_size=3, padding=1)
        self.conv5_3 = nn.Conv2d(512, 512, kernel_size=3, padding=1)
        self.pool5 = nn.MaxPool2d(kernel_size=2, stride=2)

        # Fully connected layers: three layers with ReLU activation and Dropout
        self.fc1 = nn.Linear(512 * 7 * 7, 4096)
        self.fc2 = nn.Linear(4096, 4096)
        self.fc3 = nn.Linear(4096, 1000) # 1000 classes for ImageNet

    def forward(self, x):
        # Block 1
        x = F.relu(self.conv1_1(x))
        x = F.relu(self.conv1_2(x))
        x = self.pool1(x)

        # Block 2
        x = F.relu(self.conv2_1(x))
        x = F.relu(self.conv2_2(x))
        x = self.pool2(x)

        # Block 3
        x = F.relu(self.conv3_1(x))
        x = F.relu(self.conv3_2(x))
        x = F.relu(self.conv3_3(x))
        x = self.pool3(x)

        # Block 4
        x = F.relu(self.conv4_1(x))
        x = F.relu(self.conv4_2(x))
        x = F.relu(self.conv4_3(x))
        x = self.pool4(x)

        # Block 5
        x = F.relu(self.conv5_1(x))
        x = F.relu(self.conv5_2(x))
        x = F.relu(self.conv5_3(x))
        x = self.pool5(x)

        # Flatten the output
        x = x.view(-1, 512 * 7 * 7)

        # Fully connected layers
        x = F.relu(self.fc1(x))
        x = F.dropout(x, p=0.5, training=self.training)
        x = F.relu(self.fc2(x))
        x = F.dropout(x, p=0.5, training=self.training)
        x = self.fc3(x)

        return x

# Test VGG16 network
net = VGG16()
print(net)
\end{lstlisting}

Explanation of the code:

\begin{itemize}
    \item \texttt{conv1\_1}, \texttt{conv1\_2} to \texttt{conv5\_3} are convolutional layers with a kernel size of 3x3, and they are arranged in five blocks. The output of each block is downsampled using \texttt{MaxPool2d} to reduce the spatial dimensions while retaining the most important features.
    \item \texttt{fc1}, \texttt{fc2}, and \texttt{fc3} are fully connected layers, where \texttt{fc3} outputs the final prediction for 1000 classes in the ImageNet dataset.
    \item \texttt{relu} is used as the activation function to introduce non-linearity.
    \item \texttt{dropout} is applied between the fully connected layers to prevent overfitting.
    \item The \texttt{forward} method defines the sequence of operations in the network: convolutional layers, max pooling, flattening, and fully connected layers leading to the final classification output.
\end{itemize}

\subsubsection{ResNet}
ResNet \cite{he2016deep}, short for Residual Networks, introduced residual connections that help in training very deep networks by avoiding the problem of vanishing gradients. The core idea is to add shortcuts that skip one or more layers, allowing the network to learn residual functions. ResNet has become a foundation for many advanced architectures today.

\begin{center}
\begin{tikzpicture}[transform shape]

\node (input) [draw, minimum width=2cm, minimum height=1cm] at (0, 0) {Input};

\node (conv1) [draw, minimum width=2cm, minimum height=1cm] at (0, -2) {Conv 1};
\node (bn1) [draw, minimum width=2cm, minimum height=1cm] at (0, -3.5) {BatchNorm 1};
\node (relu1) [draw, minimum width=2cm, minimum height=1cm] at (0, -5) {ReLU 1};

\node (conv2) [draw, minimum width=2cm, minimum height=1cm] at (0, -6.5) {Conv 2};
\node (bn2) [draw, minimum width=2cm, minimum height=1cm] at (0, -8) {BatchNorm 2};

\node (output) [draw, minimum width=2cm, minimum height=1cm] at (0, -9.5) {Output};

\draw[->] (input.east) .. controls +(right:2cm) and +(right:2cm) .. (output.east);

\draw[->] (input) -- (conv1);
\draw[->] (conv1) -- (bn1);
\draw[->] (bn1) -- (relu1);
\draw[->] (relu1) -- (conv2);
\draw[->] (conv2) -- (bn2);
\draw[->] (bn2) -- (output);

\end{tikzpicture}
\end{center}

In the Residual Block:

\begin{itemize}
    \item The input goes through two convolution layers, each followed by batch normalization and ReLU (only after the first convolution).
    \item A skip connection (curved arrow) bypasses these operations and directly adds the input to the output of the second batch normalization layer.
    \item The final output is the result of adding the input (via the skip connection) to the output of the convolutions, which helps in mitigating the vanishing gradient problem in deep networks.
\end{itemize}

\begin{center}
\begin{tikzpicture}[transform shape]

\node (input) [draw, minimum width=4cm, minimum height=1cm] at (0, 0) {Input Layer: 224x224x3};

\node (conv1) [draw, minimum width=4cm, minimum height=1cm] at (0, -1.5) {Conv Layer: 64x112x112};
\node (pool1) [draw, minimum width=4cm, minimum height=1cm] at (0, -3) {Max Pooling: 64x56x56};

\node (res1_1) [draw, minimum width=4cm, minimum height=1cm] at (0, -4.5) {Residual Block 1-1: 64x56x56};
\node (res1_2) [draw, minimum width=4cm, minimum height=1cm] at (0, -6) {Residual Block 1-2: 64x56x56};

\node (res2_1) [draw, minimum width=4cm, minimum height=1cm] at (0, -7.5) {Residual Block 2-1: 128x28x28};
\node (res2_2) [draw, minimum width=4cm, minimum height=1cm] at (0, -9) {Residual Block 2-2: 128x28x28};

\node (res3_1) [draw, minimum width=4cm, minimum height=1cm] at (7, -1.5) {Residual Block 3-1: 256x14x14};
\node (res3_2) [draw, minimum width=4cm, minimum height=1cm] at (7, -3) {Residual Block 3-2: 256x14x14};

\node (res4_1) [draw, minimum width=4cm, minimum height=1cm] at (7, -4.5) {Residual Block 4-1: 512x7x7};
\node (res4_2) [draw, minimum width=4cm, minimum height=1cm] at (7, -6) {Residual Block 4-2: 512x7x7};

\node (avgpool) [draw, minimum width=4cm, minimum height=1cm] at (7, -7.5) {Average Pooling: 512x1x1};
\node (fc) [draw, minimum width=4cm, minimum height=1cm] at (7, -9) {Fully Connected: 1000 classes};

\draw[->] (input) -- (conv1);
\draw[->] (conv1) -- (pool1);
\draw[->] (pool1) -- (res1_1);
\draw[->] (res1_1) -- (res1_2);
\draw[->] (res1_2) -- (res2_1);
\draw[->] (res2_1) -- (res2_2);

\draw[->] (res2_2.east) -- ++(1,0) |- (res3_1.west);
\draw[->] (res3_1) -- (res3_2);
\draw[->] (res3_2) -- (res4_1);
\draw[->] (res4_1) -- (res4_2);
\draw[->] (res4_2) -- (avgpool);
\draw[->] (avgpool) -- (fc);

\end{tikzpicture}
\end{center}

The following is the implementation of the ResNet-18 model, which is used for image classification tasks. ResNet is a deep convolutional neural network that introduced residual connections to help mitigate the vanishing gradient problem in deep networks.

\begin{lstlisting}[style=python]
import torch
import torch.nn as nn
import torch.nn.functional as F

class BasicBlock(nn.Module):
    def __init__(self, in_channels, out_channels, stride=1, downsample=None):
        super(BasicBlock, self).__init__()
        # First convolutional layer in the block
        self.conv1 = nn.Conv2d(in_channels, out_channels, kernel_size=3, stride=stride, padding=1, bias=False)
        self.bn1 = nn.BatchNorm2d(out_channels)
        # Second convolutional layer in the block
        self.conv2 = nn.Conv2d(out_channels, out_channels, kernel_size=3, stride=1, padding=1, bias=False)
        self.bn2 = nn.BatchNorm2d(out_channels)
        self.downsample = downsample

    def forward(self, x):
        identity = x
        out = F.relu(self.bn1(self.conv1(x)))
        out = self.bn2(self.conv2(out))

        if self.downsample is not None:
            identity = self.downsample(x)

        out += identity
        out = F.relu(out)
        return out

class ResNet18(nn.Module):
    def __init__(self, num_classes=1000):
        super(ResNet18, self).__init__()
        # Initial convolutional layer
        self.conv1 = nn.Conv2d(3, 64, kernel_size=7, stride=2, padding=3, bias=False)
        self.bn1 = nn.BatchNorm2d(64)
        self.pool1 = nn.MaxPool2d(kernel_size=3, stride=2, padding=1)

        # ResNet layers composed of BasicBlocks
        self.layer1 = self._make_layer(64, 64, 2)
        self.layer2 = self._make_layer(64, 128, 2, stride=2)
        self.layer3 = self._make_layer(128, 256, 2, stride=2)
        self.layer4 = self._make_layer(256, 512, 2, stride=2)

        # Fully connected layer for classification
        self.fc = nn.Linear(512, num_classes)

    def _make_layer(self, in_channels, out_channels, blocks, stride=1):
        downsample = None
        if stride != 1 or in_channels != out_channels:
            downsample = nn.Sequential(
                nn.Conv2d(in_channels, out_channels, kernel_size=1, stride=stride, bias=False),
                nn.BatchNorm2d(out_channels),
            )

        layers = []
        layers.append(BasicBlock(in_channels, out_channels, stride, downsample))
        for _ in range(1, blocks):
            layers.append(BasicBlock(out_channels, out_channels))

        return nn.Sequential(*layers)

    def forward(self, x):
        # Initial layer with convolution + batch norm + max pool
        x = F.relu(self.bn1(self.conv1(x)))
        x = self.pool1(x)

        # Passing through ResNet layers
        x = self.layer1(x)
        x = self.layer2(x)
        x = self.layer3(x)
        x = self.layer4(x)

        # Global average pooling
        x = F.adaptive_avg_pool2d(x, (1, 1))
        x = x.view(x.size(0), -1)

        # Fully connected layer for final classification
        x = self.fc(x)
        return x

# Test ResNet18 network
net = ResNet18()
print(net)
\end{lstlisting}

Explanation of the code:

\begin{itemize}
    \item The \texttt{BasicBlock} is the fundamental building block of ResNet, consisting of two convolutional layers and a residual connection (identity shortcut) that adds the input to the output, allowing gradients to flow more easily through the network.
    \item \texttt{conv1} is the initial convolutional layer, followed by batch normalization and max pooling to reduce the spatial dimensions early on.
    \item The model is composed of four layers: \texttt{layer1}, \texttt{layer2}, \texttt{layer3}, and \texttt{layer4}, each consisting of multiple \texttt{BasicBlock} modules. Each layer downscales the spatial size by half, starting from the second layer.
    \item The \texttt{\_make\_layer} function constructs each layer of the network, adjusting the stride and channels where necessary to control the dimensionality of the feature maps.
    \item A global average pooling layer is applied before the fully connected layer, which reduces the spatial dimensions to 1x1 and prepares the features for classification.
    \item The \texttt{forward} method defines the flow of data through the network: initial convolution, passing through the residual layers, and finally outputting class predictions via the fully connected layer.
\end{itemize}

%% file: 16_obj.tex
\section{Object Detection}

\subsection{Overview of Object Detection}
Object detection is a key task in computer vision, where the goal is not only to classify objects within an image but also to determine their precise locations. It involves predicting a bounding box around each detected object along with the class label that describes the object \cite{girshick2014rich,girshick2015fast,redmon2016you}.

For example, consider a photo containing a dog and a car. Object detection algorithms can both identify that there is a dog and a car in the image and pinpoint their exact locations within the image. This has wide-ranging applications such as in autonomous driving, security surveillance, robotics, and even mobile applications like augmented reality.

\subsection{YOLO (You Only Look Once)}
YOLO \cite{redmon2016you}, short for "You Only Look Once," is one of the most popular object detection algorithms due to its speed and accuracy, particularly in real-time applications. 

\subsubsection{Introduction to YOLO}
YOLO takes a unique approach compared to other object detection algorithms. It divides the image into a grid, and for each grid cell, it predicts a number of bounding boxes as well as the probability that each box contains a certain class of object. This is different from other methods like region-based algorithms that first identify regions of interest and then perform classification.

\textbf{Example:} If YOLO divides an image into a 7x7 grid, it will evaluate each grid cell and predict bounding boxes along with confidence scores for potential objects within those grid cells.

YOLO is designed to be extremely fast because it treats object detection as a single regression problem, directly mapping from the input image to bounding box coordinates and class probabilities.

\subsection{YOLO Architecture}
The architecture of YOLO is based on a single convolutional neural network (CNN). The CNN processes the image in one forward pass and outputs predictions for both the bounding boxes and class probabilities.

A typical YOLO model consists of several convolutional layers that extract features from the input image, followed by fully connected layers that produce the final bounding box and class predictions. One of the key reasons YOLO is so fast is that it makes predictions for the entire image in a single step, rather than performing region proposals and then refining them, as in other methods like Faster R-CNN.

\subsection{Object Detection with YOLOv8}

\subsubsection{Introduction to YOLOv8}
YOLOv8 \cite{jocher2023yolov8} is the latest iteration of the YOLO family of algorithms. It offers a range of improvements over its predecessors, including higher accuracy, faster inference speed, and a more efficient architecture. YOLOv8 can be used for both real-time object detection tasks and more demanding, high-resolution object detection problems.

\subsubsection{Key Features of YOLOv8}
Some of the key features of YOLOv8 include:
\begin{itemize}
    \item \textbf{Faster Speed:} YOLOv8 can process images at a very high frame rate, making it suitable for real-time applications.
    \item \textbf{Improved Accuracy:} YOLOv8 is more accurate than previous versions, making it better at detecting smaller objects and handling complex scenes.
    \item \textbf{Efficient Architecture:} The architecture has been optimized to be more memory-efficient and computationally less expensive.
\end{itemize}

\subsubsection{Installing YOLOv8}
To use YOLOv8, you'll first need to install the necessary packages. YOLOv8 is often used with the Python library \texttt{ultralytics}. Here's how you can install it:

\begin{lstlisting}[style=cmd]
pip install ultralytics
\end{lstlisting}

This will install the latest version of the YOLOv8 package, allowing you to use it for object detection tasks.

\subsubsection{Running YOLOv8 on an Image}
Once YOLOv8 is installed, you can run object detection on an image with just a few lines of code. Here's an example of how you can use YOLOv8 to detect objects in an image:

\begin{lstlisting}[style=python]
from ultralytics import YOLO

# Load the YOLOv8 model
model = YOLO('yolov8n.pt')

# Perform object detection on an image
results = model('image.jpg')

# Show results
results.show()
\end{lstlisting}

In this example, we load a pre-trained YOLOv8 model (`yolov8n.pt`), pass an image (`image.jpg`) to the model, and display the detected objects along with their bounding boxes.

\subsubsection{YOLOv8 Model Variants}

YOLO (You Only Look Once) models come in different sizes to balance between performance (inference speed) and accuracy (detection precision). Each version of YOLOv8, like other YOLO models, comes in several sizes, which trade off computation requirements for better detection performance. Here is an overview of different YOLOv8 sizes:

\begin{itemize}
    \item \textbf{YOLOv8n (Nano)}:
    \begin{itemize}
        \item Smallest model, optimized for real-time performance with lower computational resources.
        \item Best for applications on devices with limited resources, like mobile devices or embedded systems.
        \item Fast inference but lower accuracy.
    \end{itemize}

    \item \textbf{YOLOv8s (Small)}:
    \begin{itemize}
        \item Slightly larger than YOLOv8n but still very efficient.
        \item Better accuracy than the nano model while maintaining good speed for real-time applications.
    \end{itemize}

    \item \textbf{YOLOv8m (Medium)}:
    \begin{itemize}
        \item Medium-sized model with a balance between speed and accuracy.
        \item Suitable for tasks where you need higher accuracy than YOLOv8n and YOLOv8s but still want reasonable performance.
    \end{itemize}

    \item \textbf{YOLOv8l (Large)}:
    \begin{itemize}
        \item Larger model with improved accuracy, but requires more computational power.
        \item Suitable for applications where accuracy is more important than inference speed.
    \end{itemize}

    \item \textbf{YOLOv8x (Extra Large)}:
    \begin{itemize}
        \item The largest and most accurate model in the YOLOv8 family.
        \item Best for high-accuracy tasks where real-time performance is less important, like offline processing.
    \end{itemize}
\end{itemize}

Each of these models provides a trade-off between model size, speed, and accuracy. As the model size increases, accuracy tends to improve, but inference time becomes longer and the model becomes computationally heavier.

\textbf{Different YOLOv8 Model Deployments}

\begin{lstlisting}[style=python]
from ultralytics import YOLO

# Load the YOLOv8 model (nano version)
model = YOLO('yolov8n.pt')

# Load the YOLOv8 model (small version)
# model = YOLO('yolov8s.pt')  # YOLOv8s (Small): Balanced between speed and accuracy.

# Load the YOLOv8 model (medium version)
# model = YOLO('yolov8m.pt')  # YOLOv8m (Medium): Better accuracy, slightly slower inference.

# Load the YOLOv8 model (large version)
# model = YOLO('yolov8l.pt')  # YOLOv8l (Large): Higher accuracy, requires more computation.

# Load the YOLOv8 model (extra large version)
# model = YOLO('yolov8x.pt')  # YOLOv8x (Extra Large): Highest accuracy, most computationally intensive.

# Perform object detection on an image
results = model('image.jpg')

# Show results
results.show()
\end{lstlisting}

\subsubsection{Interpreting YOLOv8 Outputs}
YOLOv8 outputs several pieces of information for each detected object:
\begin{itemize}
    \item \textbf{Bounding Box Coordinates:} The (x, y) coordinates representing the top-left corner of the box, along with the width and height.
    \item \textbf{Confidence Score:} A score representing how confident the model is that the object inside the bounding box is of a certain class.
    \item \textbf{Class Label:} The class that the object has been predicted as (e.g., person, car, dog).
\end{itemize}

\textbf{Example:} If YOLOv8 detects a car with a bounding box of (100, 50, 200, 150) and a confidence score of 0.95, it means YOLOv8 is 95\% confident that a car is located in the region starting from pixel (100, 50) with a width of 200 pixels and height of 150 pixels.

\subsection{Challenges in Object Detection}
While object detection has come a long way, several challenges remain, including:
\begin{itemize}
    \item \textbf{Small Object Detection:} Detecting very small objects in an image remains challenging, as they occupy fewer pixels, making it harder to predict accurate bounding boxes.
    \item \textbf{Occlusion:} When objects overlap or block each other, it becomes difficult for the model to detect all objects correctly.
    \item \textbf{Varying Lighting Conditions:} Images taken in poor lighting or with heavy shadows can affect the accuracy of object detection models.
\end{itemize}

Many modern algorithms, including YOLOv8, have been designed to address these challenges using techniques such as multi-scale detection, anchor boxes, and data augmentation strategies.

%% file: 21_basic.tex
\chapter{Object Detection}

\section{Introduction to Object Detection}

In this section, we will provide an introduction to each model, a detailed explanation of the principle, and provide a block diagram/flowchart, pseudo code, and executable code. It is worth noting that if a model does not have executable code, it means that the efficiency or performance of the model no longer meets the requirements of today's society. Then the model is too old and is only suitable for use as a teaching model because it is very simple. For readers, you only need to understand the principle.

\subsection{Overview of Object Detection}
Object detection is the task of identifying and locating objects of certain classes (such as cars, people, animals, etc.) within an image. This means that for every object detected, the system not only identifies the category or class of the object but also draws a bounding box around it to indicate its position within the image.

Object detection is a critical part of many computer vision applications, especially in fields like:
\begin{itemize}
    \item \textbf{Autonomous driving}: Detecting pedestrians, traffic signs, and other vehicles to navigate safely.
    \item \textbf{Surveillance}: Recognizing and tracking individuals or objects in video feeds for security purposes.
    \item \textbf{Healthcare}: Detecting abnormalities in medical images such as X-rays or MRIs.
    \item \textbf{Robotics}: Enabling robots to interact with their environment by recognizing and manipulating objects.
\end{itemize}

The complexity of object detection comes from the need to handle different types of objects at varying sizes, orientations, and positions in the image. Unlike image classification, which only identifies the category of the object, object detection also needs to specify \textit{where} the object is located.

\subsection{Challenges in Object Detection}
Despite recent advances, object detection poses significant challenges. Some of the common difficulties are:

\begin{itemize}
    \item \textbf{Small objects}: Detecting small objects is particularly challenging because they contain less visual information. For instance, detecting a bird in a large landscape image can be very difficult due to its small size relative to the image.
    
    \item \textbf{Occlusion}: In real-world scenarios, objects are often partially occluded by other objects. For example, in an image of a crowd, some people might be standing behind others, making detection harder.

    \item \textbf{Varying lighting conditions}: Images captured under different lighting conditions (e.g., low light, bright light, shadows) can cause issues, as object features can become less distinguishable.

    \item \textbf{Class imbalance}: In many datasets, some classes are significantly more common than others. For example, in traffic surveillance, there are many more cars than bicycles. This imbalance can cause the detection model to be biased toward common objects while ignoring rare ones.
\end{itemize}

\subsection{Applications of Object Detection}
Object detection has numerous real-world applications:

\begin{itemize}
    \item \textbf{Autonomous vehicles}: Object detection helps self-driving cars detect other cars, pedestrians, cyclists, and obstacles, playing a crucial role in safe navigation.
    
    \item \textbf{Healthcare}: In medical imaging, object detection is used to identify tumors or other abnormalities in scans like X-rays or MRI images.
    
    \item \textbf{Security systems}: Surveillance systems use object detection to monitor and track suspicious activities. For example, detecting people in restricted areas or identifying abandoned objects.
    
    \item \textbf{Robotics}: Robots use object detection to locate and interact with objects in their environment, such as picking up tools or navigating around obstacles.
    
    \item \textbf{Retail}: Object detection is used in smart retail systems for tracking inventory, monitoring customer behavior, and enabling cashier-less checkout.
\end{itemize}

\section{Traditional Object Detection Methods}

\subsection{Sliding Window and Selective Search}
Before the advent of deep learning, traditional methods like sliding window and selective search were widely used for object detection.

\textbf{Sliding Window}: This method involves scanning the entire image with a fixed-size window. For each position of the window, features are extracted, and a classifier predicts whether an object of interest is present. By repeating this process at multiple scales, this method attempts to detect objects of various sizes. However, sliding windows are computationally expensive, as they require evaluating a large number of windows at different positions and scales.

\textbf{Selective Search}: This technique attempts to reduce the number of windows to evaluate by first dividing the image into regions using color, texture, and size. The algorithm merges similar regions to form potential objects and only evaluates a smaller number of candidate regions. While this method is more efficient than the sliding window, it still struggles with real-time performance and can miss objects if the region proposals are not accurate.

\subsection{Haar-like Features and Viola-Jones Detector}
The Viola-Jones detector \cite{viola2001rapid}, developed in 2001, was one of the first successful object detection algorithms. It was particularly popular for real-time face detection.

\textbf{Haar-like Features}: The method uses Haar-like features, which are simple rectangular filters that detect edges, lines, and textures in an image. These features are computed rapidly using an integral image, which allows for efficient computation of the sum of pixel values within a rectangular region.

\textbf{Viola-Jones Framework}:
\begin{itemize}
    \item The detector scans the image at multiple scales using a sliding window approach.
    \item For each window, it applies a classifier built using AdaBoost, which selects a small set of the most important Haar-like features from a much larger set.
    \item To improve efficiency, the algorithm uses a \textbf{cascade of classifiers}. If a window is unlikely to contain an object, it is quickly rejected by a simpler classifier in the early stages, saving computation.
\end{itemize}
Despite being fast and effective for face detection, the Viola-Jones detector struggles with more complex objects and general object detection tasks.

\subsection{Feature-based Methods (HOG, SIFT)}
Before the rise of deep learning, object detection heavily relied on extracting handcrafted features such as Histogram of Oriented Gradients (HOG) \cite{dalal2005histograms} and Scale-Invariant Feature Transform (SIFT) \cite{lowe1999object,lowe2004distinctive}.

\textbf{Histogram of Oriented Gradients (HOG)}:
\begin{itemize}
    \item The HOG method captures edge information in an image. It works by dividing the image into small cells, computing the gradient orientation in each cell, and creating a histogram of gradient directions.
    \item HOG is robust to changes in lighting and small deformations, making it suitable for detecting objects like pedestrians.
    \item After extracting HOG features, a classifier (such as a Support Vector Machine or SVM) is used to detect objects.
\end{itemize}

\begin{lstlisting}[style=python]
# Example of extracting HOG features in Python using skimage library
from skimage.feature import hog
from skimage import data, exposure
import matplotlib.pyplot as plt

# Load an example image (grayscale)
image = data.astronaut()

# Compute HOG features
hog_features, hog_image = hog(image, pixels_per_cell=(16, 16), cells_per_block=(1, 1),
                              visualize=True, channel_axis=-1)

# Show the original image and HOG visualization
fig, (ax1, ax2) = plt.subplots(1, 2, figsize=(12, 6))
ax1.imshow(image, cmap='gray')
ax1.set_title('Original Image')
ax2.imshow(hog_image, cmap='gray')
ax2.set_title('HOG Features')
plt.show()
\end{lstlisting}

\textbf{Scale-Invariant Feature Transform (SIFT)}:
\begin{itemize}
    \item SIFT is a method to detect and describe local features in an image that are invariant to scaling and rotation. This makes SIFT particularly useful for detecting objects from different viewpoints or scales.
    \item Keypoints are identified in the image, and descriptors are extracted based on the gradient orientation in the surrounding region of each keypoint.
    \item Like HOG, SIFT features can be fed into a classifier for object detection.
\end{itemize}

Despite the success of HOG and SIFT, they have been largely replaced by deep learning methods, which automatically learn relevant features directly from data.

%% file: 22_rcnn.tex
\section{Non-Maximum Suppression (NMS)}

\subsection{Introduction}
Non-Maximum Suppression (NMS) is a widely-used technique in computer vision, particularly in the domain of object detection \cite{dalal2005histograms,felzenszwalb2010object}. It is employed to eliminate multiple overlapping bounding boxes that may be predicted for the same object, thereby retaining only the most relevant one. NMS ensures that the final output contains a single, representative bounding box for each detected object, enhancing the clarity and precision of detection results.

\subsection{Purpose of NMS}
The primary goal of NMS is to filter out redundant bounding boxes that correspond to the same object. In scenarios where an object detection algorithm may produce several bounding boxes for a single object, NMS helps in:
\begin{itemize}
    \item Reducing the number of false positives.
    \item Improving the interpretability of detection results.
    \item Enhancing the overall performance of object detection models.
\end{itemize}

\subsection{How NMS Works}
The NMS algorithm operates by following these key steps:

\begin{enumerate}
    \item \textbf{Input}: Begin with a list of bounding boxes along with their associated confidence scores, which indicate the likelihood that the box contains an object.
    \item \textbf{Sorting}: Sort the bounding boxes in descending order based on their confidence scores. This ensures that the most confident predictions are processed first.
    \item \textbf{Selection}: Initialize an empty list to hold the final selected bounding boxes. Select the bounding box with the highest confidence score and add it to the final list.
    \item \textbf{Suppression}: For each remaining bounding box, calculate the Intersection over Union (IoU) with the selected bounding box. If the IoU exceeds a predefined threshold, suppress (remove) the bounding box from the list of candidates.
    \item \textbf{Repeat}: Continue the selection and suppression process until no bounding boxes remain in the list.
\end{enumerate}

\subsection{Mathematical Formulation}
The Intersection over Union (IoU) metric is crucial for determining how much two bounding boxes overlap. It is defined mathematically as:

\begin{equation}
IoU = \frac{Area\ of\ Intersection}{Area\ of\ Union}
\end{equation}

Where:
\begin{itemize}
    \item The \textit{Area of Intersection} is the area where the two bounding boxes overlap.
    \item The \textit{Area of Union} is the total area covered by both bounding boxes combined.
\end{itemize}

\subsection{Algorithm}
The following pseudo-code describes the NMS algorithm in detail:

\begin{lstlisting}[style=Python]
def non_maximum_suppression(bboxes, scores, threshold):
    # Step 1: Sort the bounding boxes by scores in descending order
    sorted_indices = sorted(range(len(scores)), key=lambda i: scores[i], reverse=True)
    selected_indices = []
    
    # Step 2: Process each bounding box
    while sorted_indices:
        # Select the bounding box with the highest score
        current_index = sorted_indices[0]
        selected_indices.append(current_index)
        sorted_indices = sorted_indices[1:]

        # Step 3: Remove overlapping boxes
        for index in sorted(sorted_indices):
            if calculate_iou(bboxes[current_index], bboxes[index]) > threshold:
                sorted_indices.remove(index)

    return selected_indices

def calculate_iou(boxA, boxB):
    # Calculate the area of intersection
    xA = max(boxA[0], boxB[0])
    yA = max(boxA[1], boxB[1])
    xB = min(boxA[2], boxB[2])
    yB = min(boxA[3], boxB[3])
    
    interArea = max(0, xB - xA) * max(0, yB - yA)
    
    # Calculate the area of both bounding boxes
    boxAArea = (boxA[2] - boxA[0]) * (boxA[3] - boxA[1])
    boxBArea = (boxB[2] - boxB[0]) * (boxB[3] - boxB[1])
    
    # Calculate the area of union
    unionArea = boxAArea + boxBArea - interArea
    
    # Compute IoU
    return interArea / unionArea
\end{lstlisting}

\subsection{Example}
Consider a scenario where we have the following bounding boxes along with their confidence scores:

\begin{itemize}
    \item Box A: $(x_1, y_1, x_2, y_2)$ with score 0.9
    \item Box B: $(x_3, y_3, x_4, y_4)$ with score 0.85
    \item Box C: $(x_5, y_5, x_6, y_6)$ with score 0.8
\end{itemize}

Suppose the IoU threshold is set to 0.5. During the NMS process, Box A would likely be selected first due to its highest score, while Boxes B and C would be evaluated against it. If either overlaps sufficiently with Box A (based on the IoU calculation), it will be suppressed.

\subsection{Conclusion}
Non-Maximum Suppression is a vital technique in the field of object detection, playing a crucial role in refining the outputs of detection algorithms. By effectively eliminating redundant bounding boxes, NMS leads to clearer and more accurate detection results. Implementing NMS can significantly enhance the performance of computer vision systems, making it a standard procedure in various applications.

\section{R-CNN and Region-Based Approaches}

In this section, we introduce a family of models that revolutionized object detection by combining region proposals with convolutional neural networks (CNNs). These models are commonly referred to as R-CNN (Region-based Convolutional Neural Networks) and its variants \cite{girshick2015fast,girshick2014rich,ren2015faster}. Each improvement over time has made object detection more efficient, faster, and more accurate.

\subsection{R-CNN: Regions with CNN Features (2014)}
R-CNN \cite{girshick2014rich}, proposed in 2014, was one of the first models to integrate CNNs for object detection. Before R-CNN, traditional methods relied heavily on hand-crafted features, such as SIFT or HOG, combined with sliding window approaches. These methods were computationally expensive and struggled to capture complex features in images.

R-CNN works by first generating around 2000 region proposals using an algorithm like Selective Search. Each of these region proposals is then resized to a fixed size and passed through a CNN (typically AlexNet at the time) to extract features. These features are used to classify objects within the region using a separate SVM (Support Vector Machine) for each object class.

Here is a step-by-step breakdown of R-CNN:
\begin{itemize}
    \item \textbf{Region Proposals:} The model generates region proposals using Selective Search, which identifies parts of the image that might contain objects.
    \item \textbf{Feature Extraction:} Each region proposal is warped to a fixed size and passed through a pre-trained CNN to extract a feature vector.
    \item \textbf{Classification:} The extracted feature vector is classified using an SVM classifier.
    \item \textbf{Bounding Box Regression:} To improve localization accuracy, a bounding box regressor refines the proposed region's coordinates.
\end{itemize}

\noindent \textbf{Example:} If we are detecting cats in an image, R-CNN would generate several region proposals, extract features from each, and then use an SVM classifier to check if any of the regions contain a cat.

Despite its success, R-CNN had significant drawbacks, such as:
\begin{itemize}
    \item \textbf{Slow Training:} The model needed to train a separate SVM for each class, making training slow.
    \item \textbf{Inefficiency:} Each region proposal needed to be processed separately by the CNN, resulting in repeated computations and slow inference times.
\end{itemize}

\textbf{Figure: RCNN Pipeline Flowchart.} The following flowchart illustrates the process of Region-based Convolutional Neural Network (RCNN). It visualizes the steps from input image, region proposals, CNN feature extraction, to the final classification and bounding box regression.

\begin{center}
\begin{tikzpicture}[
  box/.style={draw, minimum width=3cm, minimum height=1cm, align=center},
  arrow/.style={->, thick}
]

\node (input) at (0,0) [box] {Input Image};

\node (proposal) at (0,-2.5) [box] {Region Proposals};

\node (cnn) at (0,-5) [box] {CNN for Feature Extraction};

\node (flatten) at (0,-7.5) [box] {Flattened Features};

\node (classifier) at (-3,-10) [box] {Softmax Classifier};
\node (regressor) at (3,-10) [box] {Bounding Box Regressor};

\node (class_output) at (-3,-12.5) [box] {Class Prediction};
\node (bbox_output) at (3,-12.5) [box] {Bounding Box Coordinates};

\draw [arrow] (input) -- (proposal);
\draw [arrow] (proposal) -- (cnn);
\draw [arrow] (cnn) -- (flatten);
\draw [arrow] (flatten) -- (classifier);
\draw [arrow] (flatten) -- (regressor);
\draw [arrow] (classifier) -- (class_output);
\draw [arrow] (regressor) -- (bbox_output);

\end{tikzpicture}
\end{center}

\textit{In this figure, the flow of RCNN is explained starting from the input image. Region proposals are generated from the image, followed by CNN-based feature extraction. These features are flattened and sent to two separate heads: one for class prediction using a Softmax classifier, and another for bounding box regression to localize objects.}

The following is the pseudocode for the RCNN (Region-based Convolutional Neural Network) model. RCNN is a popular architecture for object detection, where the model first identifies potential object regions (Region Proposals) and then classifies each region into object categories. The key steps involve region proposal generation, feature extraction, and classification.

\begin{lstlisting}[style=python]
# Pseudocode for RCNN (Region-based Convolutional Neural Network)

# Step 1: Input image
Input: Image

# Step 2: Generate region proposals using selective search
Region_proposals = Selective_Search(Image)

# Step 3: For each region proposal
For each proposal in Region_proposals:
    # Step 4: Extract region from the image (crop region)
    Region = Crop(Proposal)

    # Step 5: Resize the region to a fixed size (e.g., 224x224)
    Region_resized = Resize(Region, size=(224, 224))

    # Step 6: Extract features from the region using a CNN
    Features = CNN(Region_resized)

    # Step 7: Classify the region using a fully connected layer or SVM
    Class_scores = Classifier(Features)

    # Step 8: Optionally, refine the bounding box coordinates
    Refined_bounding_box = BoundingBoxRegressor(Features)

# Step 9: Post-processing: Apply non-maximum suppression (NMS)
Final_predictions = NonMaximumSuppression(Class_scores, Refined_bounding_box)

# Step 10: Output the detected objects and their bounding boxes
Output: Final_predictions
\end{lstlisting}

Explanation of the pseudocode:

\begin{itemize}
    \item Step 1: The input image is passed into the system for object detection.
    \item Step 2: Region proposals are generated using selective search, which identifies potential regions in the image that may contain objects.
    \item Step 3: For each region proposal, the region is extracted (cropped) from the original image.
    \item Step 5: The region is resized to a fixed size (typically 224x224 pixels) to match the input size required by the CNN model.
    \item Step 6: A convolutional neural network (CNN) is used to extract deep features from the resized region.
    \item Step 7: A classifier, such as a fully connected layer or an SVM, is applied to the extracted features to classify the object within the region.
    \item Step 8: A bounding box regressor may be used to adjust the initial bounding box coordinates to more accurately fit the object.
    \item Step 9: Non-maximum suppression (NMS) is applied to remove duplicate or overlapping bounding boxes.
    \item Step 10: The final output consists of the detected objects and their refined bounding boxes.
\end{itemize}

\subsection{Fast R-CNN: Faster Training and Improved Detection (2015)}
Fast R-CNN \cite{girshick2015fast}, proposed in 2015, addressed the inefficiencies of R-CNN. Instead of running a CNN on each region proposal, Fast R-CNN processes the entire image once through a CNN, producing a feature map. Region proposals are then projected onto this feature map, and features for each region are extracted via a method called ROI (Region of Interest) pooling.

Here's how Fast R-CNN improves over R-CNN:
\begin{itemize}
    \item \textbf{Single Forward Pass:} Instead of passing each region proposal through a CNN, the entire image is processed once to produce a feature map. This reduces the computational cost significantly.
    \item \textbf{ROI Pooling:} ROI pooling is used to extract fixed-size feature maps for each region proposal from the larger feature map, maintaining spatial information.
    \item \textbf{Integrated Training:} Fast R-CNN integrates both classification and bounding box regression in a single network, making training faster and more efficient.
\end{itemize}

\noindent \textbf{Example:} If we are detecting dogs and cats in an image, Fast R-CNN processes the entire image, extracts features, and then classifies each region proposal with both classes in one network, instead of training separate classifiers.

\textbf{Figure: Fast R-CNN Pipeline Flowchart.} The following flowchart illustrates the process of Fast Region-based Convolutional Neural Network (Fast R-CNN). It visualizes the steps from input image, CNN feature extraction, region of interest (RoI) pooling, to the final classification and bounding box regression.

\begin{center}
\begin{tikzpicture}[
  box/.style={draw, minimum width=3.5cm, minimum height=1cm, align=center},
  arrow/.style={->, thick}
]

\node (input) at (0,0) [box] {Input Image};

\node (cnn) at (0,-2.5) [box] {CNN for Feature Extraction};

\node (roi) at (0,-5) [box] {RoI Pooling};

\node (flatten) at (0,-7.5) [box] {Flattened Features};

\node (classifier) at (-3,-10) [box] {Softmax Classifier};
\node (regressor) at (3,-10) [box] {Bounding Box Regressor};

\node (class_output) at (-3,-12.5) [box] {Class Prediction};
\node (bbox_output) at (3,-12.5) [box] {Bounding Box Coordinates};

\draw [arrow] (input) -- (cnn);
\draw [arrow] (cnn) -- (roi);
\draw [arrow] (roi) -- (flatten);
\draw [arrow] (flatten) -- (classifier);
\draw [arrow] (flatten) -- (regressor);
\draw [arrow] (classifier) -- (class_output);
\draw [arrow] (regressor) -- (bbox_output);

\end{tikzpicture}
\end{center}

\textit{In this figure, the Fast R-CNN flow is explained starting from the input image. The image is processed by a CNN to extract feature maps. Regions of interest (RoIs) are identified and pooled. The pooled features are then flattened and fed into two heads: one for class prediction using a Softmax classifier, and another for bounding box regression to localize objects.}

The following is the pseudocode for the Fast R-CNN model. Fast R-CNN improves upon RCNN by enabling the entire image and all region proposals to be processed in a single forward pass through the CNN. Instead of extracting features from each region individually, the features are extracted from the entire image, which speeds up computation.

\begin{lstlisting}[style=python]
# Pseudocode for Fast R-CNN

# Step 1: Input image
Input: Image

# Step 2: Generate region proposals using selective search
Region_proposals = Selective_Search(Image)

# Step 3: Extract feature maps from the entire image using a CNN
Feature_map = CNN(Image)

# Step 4: For each region proposal
For each proposal in Region_proposals:
    # Step 5: Extract region features from the feature map using RoI Pooling
    Region_features = RoIPooling(Feature_map, proposal)

    # Step 6: Feed the region features to fully connected layers for classification
    Class_scores = Classifier(Region_features)

    # Step 7: Predict bounding box refinements
    Bounding_box_offsets = BoundingBoxRegressor(Region_features)

# Step 8: Post-processing: Apply non-maximum suppression (NMS)
Final_predictions = NonMaximumSuppression(Class_scores, Bounding_box_offsets)

# Step 9: Output the detected objects and their bounding boxes
Output: Final_predictions
\end{lstlisting}

Explanation of the pseudocode:

\begin{itemize}
    \item Step 1: The input image is passed into the Fast R-CNN system for object detection.
    \item Step 2: Region proposals are generated using selective search, which identifies potential regions in the image that may contain objects.
    \item Step 3: The entire image is passed through a convolutional neural network (CNN) to extract a feature map, which represents high-level features of the image.
    \item Step 4: For each region proposal, region features are extracted directly from the feature map using a special layer called RoI Pooling (Region of Interest Pooling). This layer extracts fixed-size feature representations for each proposal.
    \item Step 6: The extracted region features are fed into fully connected layers to classify the objects in the proposals.
    \item Step 7: A bounding box regressor predicts bounding box offsets to refine the initial region proposal boundaries.
    \item Step 8: Non-maximum suppression (NMS) is applied to filter out overlapping or redundant bounding boxes, keeping only the best detections.
    \item Step 9: The final output consists of the detected objects and their refined bounding boxes.
\end{itemize}

\subsection{Faster R-CNN: Introduction of Region Proposal Network (RPN) (2015)}
Faster R-CNN \cite{ren2015faster}, also introduced in 2015, further improved object detection by introducing the Region Proposal Network (RPN). Instead of relying on external methods like Selective Search for generating region proposals, Faster R-CNN integrates region proposal generation into the network itself.

\noindent The process of Faster R-CNN can be described as follows:
\begin{itemize}
    \item \textbf{Region Proposal Network (RPN):} A small CNN is added to the feature map to predict whether each sliding window contains an object and to refine the coordinates of the proposal. This network is trained end-to-end along with the rest of the model.
    \item \textbf{Fast Detection:} Since RPN generates region proposals directly from the feature map, it significantly speeds up the process compared to previous methods.
    \item \textbf{Joint Training:} The RPN and Fast R-CNN components are trained together, which improves detection accuracy.
\end{itemize}

\noindent \textbf{Example:} Faster R-CNN can detect objects in real-time applications, like self-driving cars, where rapid and accurate detection of pedestrians or obstacles is critical.

\textbf{Figure: Faster R-CNN Pipeline Flowchart.} The following flowchart illustrates the process of Faster Region-based Convolutional Neural Network (Faster R-CNN). It visualizes the steps from input image, CNN feature extraction, Region Proposal Network (RPN), RoI Pooling, to the final classification and bounding box regression.

\begin{center}
\begin{tikzpicture}[
  box/.style={draw, minimum width=3cm, minimum height=1cm, align=center},
  arrow/.style={->, thick}
]

\node (input) at (0,0) [box] {Input Image};

\node (cnn) at (0,-2.5) [box] {CNN for Feature Extraction};

\node (rpn) at (0,-5) [box] {Region Proposal Network (RPN)};

\node (roi) at (0,-7.5) [box] {RoI Pooling};

\node (classifier) at (-3,-10) [box] {Softmax Classifier};
\node (regressor) at (3,-10) [box] {Bounding Box Regressor};

\node (class_output) at (-3,-12.5) [box] {Class Prediction};
\node (bbox_output) at (3,-12.5) [box] {Bounding Box Coordinates};

\draw [arrow] (input) -- (cnn);
\draw [arrow] (cnn) -- (rpn);
\draw [arrow] (rpn) -- (roi);
\draw [arrow] (roi) -- (classifier);
\draw [arrow] (roi) -- (regressor);
\draw [arrow] (classifier) -- (class_output);
\draw [arrow] (regressor) -- (bbox_output);

\end{tikzpicture}
\end{center}

\textit{In this figure, the flow of Faster R-CNN is explained starting from the input image. The CNN first extracts features from the image. Then, the Region Proposal Network (RPN) generates potential regions of interest. These regions are pooled by the RoI Pooling layer and passed to two branches: one for class prediction using a Softmax classifier, and another for bounding box regression to refine object localization.}

The following is the pseudocode for the Faster R-CNN model. Faster R-CNN improves upon previous R-CNN architectures by introducing a Region Proposal Network (RPN) that generates region proposals directly from the convolutional feature maps, making the process faster and more efficient.

\begin{lstlisting}[style=python]
# Pseudocode for Faster R-CNN (Region-based Convolutional Neural Network)

# Step 1: Input image
Input: Image

# Step 2: Extract feature maps from the image using a backbone CNN
Feature_maps = CNN(Image)

# Step 3: Generate region proposals using Region Proposal Network (RPN)
Region_proposals, Proposal_scores = RPN(Feature_maps)

# Step 4: For each region proposal
For each proposal in Region_proposals:
    # Step 5: Extract region from the feature maps using RoI pooling
    RoI_feature_map = RoIPooling(Feature_maps, proposal)

    # Step 6: Use fully connected layers to process the RoI feature map
    RoI_features = FullyConnectedLayers(RoI_feature_map)

    # Step 7: Classify the object in the region and predict the bounding box
    Class_scores, Bounding_box = ClassifierAndRegressor(RoI_features)

# Step 8: Post-processing: Apply non-maximum suppression (NMS)
Final_predictions = NonMaximumSuppression(Class_scores, Bounding_box)

# Step 9: Output the detected objects and their bounding boxes
Output: Final_predictions
\end{lstlisting}

Explanation of the pseudocode:

\begin{itemize}
    \item Step 1: The input image is passed into the Faster R-CNN pipeline for object detection.
    \item Step 2: A convolutional neural network (CNN) is used as the backbone to extract feature maps from the input image. These feature maps are used for both region proposal generation and object classification.
    \item Step 3: The Region Proposal Network (RPN) takes the feature maps as input and generates a set of region proposals (potential object locations) along with objectness scores that indicate the likelihood of an object being in each region.
    \item Step 4: For each region proposal, a Region of Interest (RoI) pooling layer is used to extract a fixed-size feature map corresponding to that region from the CNN feature maps.
    \item Step 6: The RoI feature map is processed through fully connected layers to generate higher-level features for object classification and bounding box regression.
    \item Step 7: The classifier predicts the object class, and the bounding box regressor adjusts the coordinates of the bounding box for the object.
    \item Step 8: Non-maximum suppression (NMS) is applied to remove redundant or overlapping bounding boxes and retain the most confident detections.
    \item Step 9: The final output includes the detected objects and their refined bounding boxes.
\end{itemize}

The following is an implementation of Faster R-CNN using PyTorch for practical deployment scenarios.

\begin{lstlisting}[style=python]
import torch
import torchvision
from torchvision.models.detection import fasterrcnn_resnet50_fpn
from torchvision.transforms import functional as F
from PIL import Image

# Load a pre-trained Faster R-CNN model
model = fasterrcnn_resnet50_fpn(pretrained=True)
model.eval()  # Set to evaluation mode for inference

# Define transformation function
def transform_image(image):
    # Convert image to tensor and normalize it
    image = F.to_tensor(image)
    return image

# Load an image and apply transformations
image_path = "sample_image.jpg"
image = Image.open(image_path)
image_tensor = transform_image(image)

# Perform inference
with torch.no_grad():  # No need to calculate gradients for inference
    prediction = model([image_tensor])

# Print the prediction results
print("Predicted bounding boxes: ", prediction[0]['boxes'])
print("Predicted labels: ", prediction[0]['labels'])
print("Predicted scores: ", prediction[0]['scores'])

# For training, use this instead:
def train_faster_rcnn(model, train_data_loader, optimizer, device):
    model.train()  # Set to training mode
    for images, targets in train_data_loader:
        images = [img.to(device) for img in images]
        targets = [{k: v.to(device) for k, v in t.items()} for t in targets]
        
        # Forward pass
        loss_dict = model(images, targets)
        losses = sum(loss for loss in loss_dict.values())
        
        # Backward pass
        optimizer.zero_grad()
        losses.backward()
        optimizer.step()

# Note: Make sure to configure the optimizer and data loader appropriately
\end{lstlisting}

This code demonstrates how to load a pre-trained Faster R-CNN model, apply necessary transformations to input images, and perform object detection by predicting bounding boxes, labels, and scores. It also includes a simple training loop example, showing how to train the model using a custom dataset with an appropriate optimizer and data loader.

\subsection{Mask R-CNN: Extending Faster R-CNN for Instance Segmentation (2017)}
Mask R-CNN \cite{he2017roialign}, proposed in 2017, builds upon Faster R-CNN to enable not just object detection, but instance segmentation. In instance segmentation, we not only detect objects but also mark their boundaries at the pixel level. Mask R-CNN adds a branch to predict a segmentation mask for each detected object.

Here's how Mask R-CNN works:
\begin{itemize}
    \item \textbf{Segmentation Mask:} In addition to predicting the bounding box and object class, Mask R-CNN predicts a binary mask for each object, which represents the object's shape at a pixel level.
    \item \textbf{ROI Align:} Mask R-CNN uses ROI Align instead of ROI Pooling to preserve spatial information better, leading to more accurate masks.
\end{itemize}

\noindent \textbf{Example:} If detecting cars and pedestrians in an image, Mask R-CNN will not only draw boxes around each object but will also predict the exact shape of the car and pedestrian, making it highly useful for tasks like autonomous driving or image editing.

\textbf{Figure: Mask R-CNN Pipeline Flowchart.} The following flowchart illustrates the process of Mask R-CNN. It visualizes the steps from input image, region proposals, CNN feature extraction, to the final classification, bounding box regression, and mask prediction.

\begin{center}
\begin{tikzpicture}[
  box/.style={draw, minimum width=3cm, minimum height=1cm, align=center},
  arrow/.style={->, thick}
]

\node (input) at (0,0) [box] {Input Image};

\node (proposal) at (0,-2.5) [box] {Region Proposals};

\node (cnn) at (0,-5) [box] {CNN for Feature Extraction};

\node (flatten) at (0,-7.5) [box] {Flattened Features};

\node (classifier) at (-5,-10) [box] {Softmax Classifier};
\node (regressor) at (0,-10) [box] {Bounding Box Regressor};
\node (mask) at (5,-10) [box] {Mask Head};

\node (class_output) at (-5,-12.5) [box] {Class Prediction};
\node (bbox_output) at (0,-12.5) [box] {Bounding Box Coordinates};
\node (mask_output) at (5,-12.5) [box] {Predicted Mask};

\draw [arrow] (input) -- (proposal);
\draw [arrow] (proposal) -- (cnn);
\draw [arrow] (cnn) -- (flatten);
\draw [arrow] (flatten) -- (classifier);
\draw [arrow] (flatten) -- (regressor);
\draw [arrow] (flatten) -- (mask);
\draw [arrow] (classifier) -- (class_output);
\draw [arrow] (regressor) -- (bbox_output);
\draw [arrow] (mask) -- (mask_output);

\end{tikzpicture}
\end{center}

\textit{In this figure, the flow of Mask R-CNN is explained starting from the input image. Region proposals are generated from the image, followed by CNN-based feature extraction. These features are flattened and sent to three heads: one for class prediction using a Softmax classifier, another for bounding box regression to localize objects, and a mask head to predict segmentation masks for each object.}

The following is the pseudocode for the Mask R-CNN model. Mask R-CNN extends Faster R-CNN by adding a branch for predicting segmentation masks on each Region of Interest (RoI), in parallel with the existing branch for classification and bounding box regression. Mask R-CNN is widely used for instance segmentation tasks.

\begin{lstlisting}[style=python]
# Pseudocode for Mask R-CNN

# Step 1: Input image
Input: Image

# Step 2: Extract feature maps using a backbone CNN (e.g., ResNet)
Feature_maps = Backbone_CNN(Image)

# Step 3: Generate region proposals using a Region Proposal Network (RPN)
Region_proposals = RPN(Feature_maps)

# Step 4: For each region proposal
For each proposal in Region_proposals:
    # Step 5: Use RoIAlign to extract a fixed-size feature map from the region
    RoI_feature = RoIAlign(Feature_maps, proposal, output_size=(7, 7))

    # Step 6: Classify the region and refine the bounding box
    Class_scores, Bounding_box = Classifier_and_BoxRegressor(RoI_feature)

    # Step 7: Predict segmentation mask for the region
    Mask = MaskBranch(RoI_feature)

# Step 8: Post-processing: Apply non-maximum suppression (NMS) on the bounding boxes
Filtered_boxes = NonMaximumSuppression(Class_scores, Bounding_box)

# Step 9: Output the detected objects, bounding boxes, and masks
Output: Filtered_boxes, Masks
\end{lstlisting}

Explanation of the pseudocode:

\begin{itemize}
    \item Step 1: The input image is provided for object detection and instance segmentation.
    \item Step 2: Feature extraction is performed using a deep CNN backbone (such as ResNet). These feature maps are used by subsequent layers for detecting objects and predicting masks.
    \item Step 3: A Region Proposal Network (RPN) generates candidate region proposals from the feature maps, indicating where objects might be located.
    \item Step 4: For each proposed region, further processing is applied:
        \begin{itemize}
            \item Step 5: RoIAlign is used to extract a fixed-size feature map for each region proposal, ensuring accurate spatial alignment of features.
            \item Step 6: The extracted features are used to classify the object within the region and predict the bounding box coordinates.
            \item Step 7: A separate branch is used to predict the pixel-wise segmentation mask for the object in the region.
        \end{itemize}
    \item Step 8: Non-maximum suppression (NMS) is applied to the predicted bounding boxes to filter out duplicate or overlapping boxes.
    \item Step 9: The final output includes the detected objects, their bounding boxes, and the corresponding segmentation masks.
\end{itemize}

The following is an implementation of Mask R-CNN using PyTorch for practical deployment scenarios.

\begin{lstlisting}[style=python]
import torch
import torchvision
from torchvision.models.detection import maskrcnn_resnet50_fpn
from torchvision.transforms import functional as F
from PIL import Image

# Load a pre-trained Mask R-CNN model
model = maskrcnn_resnet50_fpn(pretrained=True)
model.eval()  # Set to evaluation mode for inference

# Define transformation function
def transform_image(image):
    # Convert image to tensor and normalize it
    image = F.to_tensor(image)
    return image

# Load an image and apply transformations
image_path = "sample_image.jpg"
image = Image.open(image_path)
image_tensor = transform_image(image)

# Perform inference
with torch.no_grad():  # No need to calculate gradients for inference
    prediction = model([image_tensor])

# Print the prediction results
print("Predicted bounding boxes: ", prediction[0]['boxes'])
print("Predicted labels: ", prediction[0]['labels'])
print("Predicted scores: ", prediction[0]['scores'])
print("Predicted masks: ", prediction[0]['masks'])

# For training, use this instead:
def train_mask_rcnn(model, train_data_loader, optimizer, device):
    model.train()  # Set to training mode
    for images, targets in train_data_loader:
        images = [img.to(device) for img in images]
        targets = [{k: v.to(device) for k, v in t.items()} for t in targets]
        
        # Forward pass
        loss_dict = model(images, targets)
        losses = sum(loss for loss in loss_dict.values())
        
        # Backward pass
        optimizer.zero_grad()
        losses.backward()
        optimizer.step()

# Note: Make sure to configure the optimizer and data loader appropriately
\end{lstlisting}

This code demonstrates how to load a pre-trained Mask R-CNN model, apply necessary transformations to input images, and perform object detection along with instance segmentation by predicting bounding boxes, labels, masks, and scores. It also includes a simple training loop example, showing how to train the model using a custom dataset with an appropriate optimizer and data loader.

\section{SPPNet: Spatial Pyramid Pooling Networks (2015)}

\subsection{Introduction to SPPNet}
Spatial Pyramid Pooling (SPP) Networks \cite{he2014spatial} introduced a method to handle input images of varying sizes without needing to resize them. Traditionally, CNNs require a fixed-size input, but resizing images can distort their content. SPPNet solves this by applying spatial pyramid pooling to the feature maps generated by the CNN.

\subsection{Advantages of Spatial Pyramid Pooling}
SPPNet provides several key advantages:
\begin{itemize}
    \item \textbf{Handles Varying Input Sizes:} The spatial pyramid pooling layer allows the network to accept input images of different sizes, which is useful for images with varying aspect ratios.
    \item \textbf{Improved Speed:} By not requiring image resizing, SPPNet improves processing speed compared to models that do resize images.
\end{itemize}

\subsection{Comparison of SPPNet with R-CNN and Fast R-CNN}
Compared to R-CNN and Fast R-CNN, SPPNet is more flexible because it does not require fixed-size inputs. This flexibility allows it to handle more diverse datasets without compromising performance. However, Fast R-CNN's ROI pooling eventually replaced SPPNet's method due to its ability to integrate better with end-to-end training pipelines.

\subsection{Limitations and Legacy of SPPNet}
Despite its advantages, SPPNet had some limitations:
\begin{itemize}
    \item \textbf{Separate Training:} Like R-CNN, SPPNet required separate stages for training and inference, which hindered end-to-end learning.
    \item \textbf{Replaced by ROI Pooling:} Fast R-CNN's ROI Pooling layer eventually replaced SPPNet because it offered a more integrated approach for handling region proposals.
\end{itemize}

\noindent \textbf{Legacy:} Although SPPNet was eventually superseded, its contributions in handling variable input sizes influenced later advancements, such as ROI Pooling and ROI Align in Fast and Mask R-CNN.

\textbf{Figure: SPPNet Pipeline Flowchart.} The following flowchart illustrates the process of Spatial Pyramid Pooling Network (SPPNet). It visualizes the steps from input image, convolutional feature extraction, spatial pyramid pooling, to the final classification and bounding box regression.

\begin{center}
\begin{tikzpicture}[
  box/.style={draw, minimum width=3cm, minimum height=1cm, align=center},
  arrow/.style={->, thick}
]

\node (input) at (0,0) [box] {Input Image};

\node (cnn) at (0,-2.5) [box] {CNN for Feature Extraction};

\node (spp) at (0,-5) [box] {Spatial Pyramid Pooling (SPP)};

\node (fc) at (0,-7.5) [box] {Fully Connected Layers};

\node (classifier) at (-3,-10) [box] {Softmax Classifier};
\node (regressor) at (3,-10) [box] {Bounding Box Regressor};

\node (class_output) at (-3,-12.5) [box] {Class Prediction};
\node (bbox_output) at (3,-12.5) [box] {Bounding Box Coordinates};

\draw [arrow] (input) -- (cnn);
\draw [arrow] (cnn) -- (spp);
\draw [arrow] (spp) -- (fc);
\draw [arrow] (fc) -- (classifier);
\draw [arrow] (fc) -- (regressor);
\draw [arrow] (classifier) -- (class_output);
\draw [arrow] (regressor) -- (bbox_output);

\end{tikzpicture}
\end{center}

\textit{In this figure, the flow of SPPNet is explained starting from the input image. The image is passed through a CNN for feature extraction, followed by Spatial Pyramid Pooling (SPP), which pools the features at multiple scales. The pooled features are passed to fully connected layers and then split into two heads: one for class prediction using a Softmax classifier and another for bounding box regression.}

The following is the pseudocode for the SPPNet (Spatial Pyramid Pooling Network) model. SPPNet improves on the RCNN architecture by introducing a Spatial Pyramid Pooling layer, which eliminates the need to warp or resize input regions to a fixed size. This allows the network to generate a fixed-length representation regardless of the input region size, improving computational efficiency and accuracy.

\begin{lstlisting}[style=python]
# Pseudocode for SPPNet (Spatial Pyramid Pooling Network)

# Step 1: Input image
Input: Image

# Step 2: Extract feature map from the entire image using a CNN
Feature_map = CNN(Image)

# Step 3: Generate region proposals using selective search
Region_proposals = Selective_Search(Image)

# Step 4: For each region proposal
For each proposal in Region_proposals:
    # Step 5: Map region proposal onto the feature map
    Mapped_region = MapToFeatureMap(Proposal, Feature_map)

    # Step 6: Apply spatial pyramid pooling to the mapped region
    Pooled_features = SpatialPyramidPooling(Mapped_region)

    # Step 7: Classify the region using a fully connected layer or SVM
    Class_scores = Classifier(Pooled_features)

    # Step 8: Optionally, refine the bounding box coordinates
    Refined_bounding_box = BoundingBoxRegressor(Pooled_features)

# Step 9: Post-processing: Apply non-maximum suppression (NMS)
Final_predictions = NonMaximumSuppression(Class_scores, Refined_bounding_box)

# Step 10: Output the detected objects and their bounding boxes
Output: Final_predictions
\end{lstlisting}

Explanation of the pseudocode:

\begin{itemize}
    \item Step 1: The input image is passed into the system for object detection.
    \item Step 2: Instead of processing each region separately, a single forward pass of a CNN is applied to the entire image, generating a feature map.
    \item Step 3: Region proposals are generated using selective search, identifying regions of interest in the image that may contain objects.
    \item Step 5: Each region proposal is mapped onto the corresponding section of the CNN feature map, avoiding redundant feature extraction.
    \item Step 6: Spatial pyramid pooling is applied to each mapped region to generate a fixed-length feature vector, regardless of the size of the region.
    \item Step 7: A classifier, such as a fully connected layer or SVM, is applied to the pooled features to classify the object in the region.
    \item Step 8: A bounding box regressor may be used to refine the bounding box coordinates.
    \item Step 9: Non-maximum suppression (NMS) is applied to remove duplicate or overlapping bounding boxes.
    \item Step 10: The final output consists of the detected objects and their refined bounding boxes.
\end{itemize}

%% file: 23_yolo.tex
\section{YOLO: You Only Look Once Series}

The YOLO (You Only Look Once) \cite{redmon2016you} series of object detection models has been a significant advancement in computer vision, especially for tasks like real-time object detection. This section will guide you through the evolution of the YOLO models, starting from YOLO v1 to the latest advancements. Each model version brings specific improvements to the speed, accuracy, and performance of object detection.

\subsection{YOLO v1: Unified Detection Framework (2016)}

In 2016, YOLO v1 \cite{redmon2016you} changed the landscape of object detection by introducing a novel approach that framed object detection as a single regression problem, unlike traditional methods, which typically used a combination of region proposals and classification tasks.

YOLO v1 divides the image into a grid of \(S \times S\) cells. Each cell predicts bounding boxes and confidence scores for objects. This unified approach enabled real-time detection by removing the need for complex pipelines used in methods like R-CNN.

\textbf{Example:} Suppose you want to detect objects in a given image of a street scene. YOLO v1 divides the image into grid cells (e.g., \(7 \times 7\)), each predicting two bounding boxes and the probability of an object being in that cell. YOLO v1 uses convolutional neural networks (CNNs) to make these predictions.

\begin{lstlisting}[style=python]
# Example YOLO v1 workflow in Python using pseudo-code
image = load_image('street_scene.jpg')
grid_size = 7
bounding_boxes = model.predict(image, grid_size)
plot_boxes(image, bounding_boxes)
\end{lstlisting}

\textit{Advantages:}
\begin{itemize}
    \item Real-time detection: YOLO v1 can process images at high frame rates, making it suitable for video applications.
    \item Unified framework: Unlike traditional methods, YOLO predicts bounding boxes and class probabilities simultaneously.
\end{itemize}

\textit{Limitations:}
\begin{itemize}
    \item Struggles with small objects: YOLO v1 often struggles to detect smaller objects, as the grid cells may not localize them accurately.
    \item Limited accuracy compared to other detectors at the time, like Faster R-CNN.
\end{itemize}

\textbf{Figure: YOLO v1 Pipeline Flowchart.} The following flowchart illustrates the process of You Only Look Once (YOLO) version 1. It visualizes the steps from input image, CNN-based feature extraction, grid division, and the final detection including class and bounding box predictions.

\begin{center}
\begin{tikzpicture}[
  box/.style={draw, minimum width=3cm, minimum height=1cm, align=center},
  arrow/.style={->, thick}
]

\node (input) at (0,0) [box] {Input Image};

\node (cnn) at (0,-2.5) [box] {CNN for Feature Extraction};

\node (grid) at (0,-5) [box] {Divide Feature Map into Grid};

\node (detection) at (0,-7.5) [box] {Bounding Box + Class Prediction};

\node (output) at (0,-10) [box] {Output Detections};

\draw [arrow] (input) -- (cnn);
\draw [arrow] (cnn) -- (grid);
\draw [arrow] (grid) -- (detection);
\draw [arrow] (detection) -- (output);

\end{tikzpicture}
\end{center}

\textit{In this figure, the flow of YOLO v1 is explained starting from the input image. The image is processed by a single CNN that extracts features. The feature map is then divided into a grid. For each grid cell, the network predicts bounding boxes and class probabilities, resulting in the final object detections.}

The following is the pseudocode for the YOLO (You Only Look Once) v1 model. YOLO is a real-time object detection system that treats object detection as a single regression problem, directly predicting bounding boxes and class probabilities from the entire image in a single forward pass of the network.

\begin{lstlisting}[style=python]
# Pseudocode for YOLO v1 (You Only Look Once)

# Step 1: Input image
Input: Image

# Step 2: Divide the image into S x S grid cells
Grid_cells = Divide_into_grid(Image, S, S)

# Step 3: For each grid cell
For each grid_cell in Grid_cells:
    # Step 4: Predict B bounding boxes and confidence scores for each box
    Bounding_boxes = Predict_BoundingBoxes(grid_cell, B)
    Confidence_scores = Predict_ConfidenceScores(grid_cell, B)

    # Step 5: Predict C class probabilities for each grid cell
    Class_probabilities = Predict_ClassProbabilities(grid_cell, C)

# Step 6: For each predicted bounding box
For each bounding_box in Bounding_boxes:
    # Step 7: Calculate the final confidence score by multiplying
    # the objectness score (confidence score) with the class probabilities
    Final_confidence_score = Confidence_scores * Class_probabilities

# Step 8: Post-processing: Apply non-maximum suppression (NMS) to remove duplicate boxes
Final_predictions = NonMaximumSuppression(Bounding_boxes, Final_confidence_score)

# Step 9: Output the detected objects and their bounding boxes
Output: Final_predictions
\end{lstlisting}

Explanation of the pseudocode:

\begin{itemize}
    \item Step 1: The input image is passed to the system for object detection.
    \item Step 2: The image is divided into a grid of \( S \times S \) cells. Each grid cell is responsible for detecting objects whose center falls within the cell.
    \item Step 3: For each grid cell, the network predicts \( B \) bounding boxes, each with a confidence score indicating whether an object exists in the box.
    \item Step 5: Additionally, the network predicts class probabilities for \( C \) object classes for each grid cell, regardless of the number of bounding boxes.
    \item Step 7: The final confidence score for each bounding box is computed by multiplying the objectness score (confidence score) by the predicted class probabilities.
    \item Step 8: Non-maximum suppression (NMS) is applied to remove overlapping or redundant bounding boxes, keeping only the most confident predictions.
    \item Step 9: The final output consists of detected objects and their refined bounding boxes.
\end{itemize}

\subsection{YOLO v2 and v3: Incremental Improvements and Better Performance (2017-2018)}

YOLO v2 (also known as YOLO9000) \cite{redmon2017yolo9000} and YOLO v3 \cite{redmon2018yolov3}introduced several incremental improvements over YOLO v1 to enhance accuracy and flexibility, while still maintaining real-time detection speeds.

\textit{Key improvements in YOLO v2:}
\begin{itemize}
    \item \textit{Anchor boxes}: YOLO v2 introduced anchor boxes, a set of predefined bounding boxes, to better handle varying object sizes.
    \item \textit{Batch normalization}: Applied to the convolutional layers to reduce overfitting and improve accuracy.
    \item \textit{Multi-scale training}: YOLO v2 trains on images of varying sizes, improving the model's ability to generalize across different resolutions.
\end{itemize}

\textit{YOLO v3 Enhancements:}
\begin{itemize}
    \item YOLO v3 introduced \textit{multi-scale predictions} by predicting at three different scales, which allows it to detect both small and large objects more accurately.
    \item Use of \textit{residual blocks} to improve gradient flow, making deeper networks possible without a significant drop in performance.
\end{itemize}

\begin{lstlisting}[style=python]
# YOLO v2/v3 example workflow with anchor boxes and multi-scale predictions
image = load_image('park_scene.jpg')
anchor_boxes = generate_anchor_boxes()
multi_scale_bboxes = model.predict(image, scales=[320, 416, 608])
plot_boxes(image, multi_scale_bboxes)
\end{lstlisting}

\textit{Benefits:}
\begin{itemize}
    \item Better at detecting small objects due to anchor boxes and multi-scale predictions.
    \item Improved accuracy and speed balance compared to YOLO v1.
\end{itemize}

\textbf{Figure: YOLOv2 Pipeline Flowchart.} The following flowchart illustrates the process of You Only Look Once version 2 (YOLOv2). It shows the steps from input image, convolutional layers, to the final output grid with class predictions and bounding box coordinates.

\begin{center}
\begin{tikzpicture}[
  box/.style={draw, minimum width=3cm, minimum height=1cm, align=center},
  arrow/.style={->, thick}
]

\node (input) at (0,0) [box] {Input Image};

\node (conv) at (0,-2.5) [box] {Convolutional Layers};

\node (detection) at (0,-5) [box] {Detection Layer (Output Grid)};

\node (bboxes) at (-3,-7.5) [box] {Bounding Boxes};
\node (class_pred) at (3,-7.5) [box] {Class Predictions};

\draw [arrow] (input) -- (conv);
\draw [arrow] (conv) -- (detection);
\draw [arrow] (detection) -- (bboxes);
\draw [arrow] (detection) -- (class_pred);

\end{tikzpicture}
\end{center}

\textit{This flowchart illustrates the YOLOv2 architecture. The input image is processed by several convolutional layers to extract features. The detection layer generates a grid of predictions, which includes both the bounding boxes and class predictions for objects detected in the image.}

The following is the pseudocode for the YOLO v2 (You Only Look Once version 2) model. YOLO v2 is a fast object detection architecture that divides the image into a grid and simultaneously predicts bounding boxes and class probabilities directly from full images in a single forward pass.

\begin{lstlisting}[style=python]
# Pseudocode for YOLO v2 (You Only Look Once v2)

# Step 1: Input image
Input: Image

# Step 2: Resize image to a fixed size (e.g., 416x416)
Resized_image = Resize(Image, size=(416, 416))

# Step 3: Pass the resized image through the YOLO network (CNN-based)
Feature_map = YOLO_Network(Resized_image)

# Step 4: Divide the image into an SxS grid (e.g., 13x13)
Grid_cells = Divide_into_Grid(Feature_map, grid_size=(S, S))

# Step 5: For each grid cell, predict bounding boxes and class probabilities
For each cell in Grid_cells:
    # Predict B bounding boxes with confidence scores
    For each bounding_box in range(B):
        Bounding_box, Confidence = Predict_Bounding_Box(cell)

    # Predict class probabilities for objects in the grid cell
    Class_probabilities = Predict_Class_Probabilities(cell)

    # Combine confidence score with class probabilities for each bounding box
    Final_scores = Confidence * Class_probabilities

# Step 6: Post-processing: Apply non-maximum suppression (NMS)
Final_predictions = NonMaximumSuppression(Final_scores, Bounding_boxes)

# Step 7: Output the detected objects with their bounding boxes and classes
Output: Final_predictions
\end{lstlisting}

Explanation of the pseudocode:

\begin{itemize}
    \item Step 1: The input image is passed into the system for object detection.
    \item Step 2: The image is resized to a fixed size (416x416) to match the input size required by the YOLO network.
    \item Step 3: The resized image is fed through the YOLO v2 network, which is a convolutional neural network designed for real-time object detection.
    \item Step 4: The output feature map is divided into a grid of SxS cells (e.g., 13x13 grid).
    \item Step 5: For each grid cell, the model predicts B bounding boxes (e.g., 5 boxes) along with confidence scores for the presence of objects. It also predicts class probabilities for the detected objects in the cell.
    \item Step 6: Non-maximum suppression (NMS) is applied to remove overlapping or duplicate bounding boxes, keeping only the highest scoring ones.
    \item Step 7: The final output consists of the detected objects, their bounding boxes, and their corresponding class predictions.
\end{itemize}

\textbf{Figure: YOLO v3 Pipeline Flowchart.} The following flowchart illustrates the process of YOLO v3. It visualizes the steps from input image, CNN feature extraction, multi-scale feature prediction, to the final class prediction and bounding box regression.

\begin{center}
\begin{tikzpicture}[
  box/.style={draw, minimum width=3cm, minimum height=1cm, align=center},
  arrow/.style={->, thick}
]

\node (input) at (0,0) [box] {Input Image};

\node (cnn) at (0,-2.5) [box] {CNN Feature Extraction};

\node (scale1) at (-5,-5) [box] {Scale 1 Prediction (13x13)};
\node (scale2) at (0,-5) [box] {Scale 2 Prediction (26x26)};
\node (scale3) at (5,-5) [box] {Scale 3 Prediction (52x52)};

\node (class_bbox1) at (-5,-7.5) [box] {Class + BBox (13x13)};
\node (class_bbox2) at (0,-7.5) [box] {Class + BBox (26x26)};
\node (class_bbox3) at (5,-7.5) [box] {Class + BBox (52x52)};

\draw [arrow] (input) -- (cnn);
\draw [arrow] (cnn) -- (scale1);
\draw [arrow] (cnn) -- (scale2);
\draw [arrow] (cnn) -- (scale3);
\draw [arrow] (scale1) -- (class_bbox1);
\draw [arrow] (scale2) -- (class_bbox2);
\draw [arrow] (scale3) -- (class_bbox3);

\end{tikzpicture}
\end{center}

\textit{In this figure, the flow of YOLO v3 is explained starting from the input image. The CNN extracts features, and predictions are made at three different scales: 13x13, 26x26, and 52x52. Each scale provides class predictions and bounding box coordinates.}

The following is the pseudocode for the YOLOv3 (You Only Look Once version 3) model. YOLOv3 is a real-time object detection system that directly predicts bounding boxes and class probabilities from the entire image in a single forward pass. It divides the image into a grid and makes predictions for each grid cell, including bounding box coordinates, object confidence, and class probabilities.

\begin{lstlisting}[style=python]
# Pseudocode for YOLOv3 (You Only Look Once version 3)

# Step 1: Input image
Input: Image

# Step 2: Divide the image into SxS grid cells
Grid = Divide_Image_Into_Grid(Image, S=S)

# Step 3: Pass the image through the convolutional neural network (Darknet-53) to extract features
Features = Darknet53(Image)

# Step 4: For each grid cell in the feature map
For each grid_cell in Grid:
    # Step 5: Predict B bounding boxes for each grid cell
    For each bounding_box in B:
        # Predict bounding box coordinates (x, y, width, height)
        BoundingBox = Predict_BoundingBox(Features, grid_cell)
        
        # Step 6: Predict the object confidence score
        Object_confidence = Predict_Object_Confidence(Features, grid_cell)

        # Step 7: Predict class probabilities for each bounding box
        Class_probabilities = Predict_Class_Probabilities(Features, grid_cell)

    # Combine the predictions
    Predictions[grid_cell] = {BoundingBox, Object_confidence, Class_probabilities}

# Step 8: Post-processing: Apply non-maximum suppression (NMS)
Final_predictions = NonMaximumSuppression(Predictions)

# Step 9: Output the detected objects, their bounding boxes, and class labels
Output: Final_predictions
\end{lstlisting}

Explanation of the pseudocode:

\begin{itemize}
    \item Step 1: The input image is processed for object detection.
    \item Step 2: The image is divided into a grid of size SxS (usually 13x13, 26x26, or 52x52 depending on the feature map size).
    \item Step 3: YOLOv3 uses a deep convolutional neural network (Darknet-53) to extract features from the entire image.
    \item Step 4-5: For each grid cell, YOLOv3 predicts multiple bounding boxes (B bounding boxes) with corresponding coordinates.
    \item Step 6: YOLOv3 predicts an object confidence score for each bounding box, indicating whether an object is present in the bounding box.
    \item Step 7: The model predicts class probabilities for each bounding box, indicating the category of the detected object.
    \item Step 8: Non-maximum suppression (NMS) is applied to eliminate redundant or overlapping bounding boxes and select the most confident predictions.
    \item Step 9: The final output consists of detected objects, their refined bounding boxes, and the associated class labels.
\end{itemize}

\subsection{YOLO v4: Real-Time Detection with High Accuracy (2020)}

YOLO v4 \cite{bochkovskiy2020yolov4}, released in 2020, made significant advancements by optimizing the model for real-time object detection without sacrificing accuracy. YOLO v4 introduced new techniques known as \textit{bag-of-freebies} and \textit{bag-of-specials}.

\textit{Key Techniques in YOLO v4:}
\begin{itemize}
    \item \textit{Bag-of-freebies}: These are training techniques like data augmentation, multi-scale training, and various loss functions that improve accuracy without adding computational cost during inference.
    \item \textit{Bag-of-specials}: These are architectural improvements such as mish activation, CSPDarknet53 as the backbone, and PANet \cite{liu2018path} for better feature aggregation, designed to improve both speed and accuracy.
\end{itemize}

\textbf{Example:} For real-time applications like autonomous driving, YOLO v4's optimizations allow detecting pedestrians and cars accurately at high frame rates.

\begin{lstlisting}[style=python]
# YOLO v4 real-time detection example
image = load_image('highway.jpg')
detected_objects = yolo_v4_model.detect(image)
plot_boxes(image, detected_objects)
\end{lstlisting}

\textit{Advantages:}
\begin{itemize}
    \item Real-time performance with higher accuracy than previous YOLO versions.
    \item Highly suitable for edge devices and real-time applications like drone vision or autonomous vehicles.
\end{itemize}

\textbf{Figure: YOLOv4 Pipeline Flowchart.} The following flowchart illustrates the process of YOLOv4 (You Only Look Once version 4). It visualizes the steps from input image, feature extraction (Backbone and Neck), to the final detection including class prediction and bounding box regression.

\begin{center}
\begin{tikzpicture}[
  box/.style={draw, minimum width=4cm, minimum height=1cm, align=center},
  arrow/.style={->, thick}
]

\node (input) at (0,0) [box] {Input Image};

\node (backbone) at (0,-2.5) [box] {Backbone (CSPDarknet53)};

\node (neck) at (0,-5) [box] {Neck (SPP + PANet)};

\node (head) at (0,-7.5) [box] {Prediction Head};

\node (class_output) at (-3,-10) [box] {Class Prediction};
\node (bbox_output) at (3,-10) [box] {Bounding Box Regression};

\node (detection_output) at (0,-12.5) [box] {Final Detection (Class + Box)};

\draw [arrow] (input) -- (backbone);
\draw [arrow] (backbone) -- (neck);
\draw [arrow] (neck) -- (head);
\draw [arrow] (head) -- (class_output);
\draw [arrow] (head) -- (bbox_output);
\draw [arrow] (class_output) -- (detection_output);
\draw [arrow] (bbox_output) -- (detection_output);

\end{tikzpicture}
\end{center}

\textit{In this figure, the YOLOv4 pipeline is illustrated. The input image is passed through the Backbone (CSPDarknet53) for feature extraction. The Neck (SPP + PANet) is used for aggregating features, followed by the Prediction Head, which outputs both class predictions and bounding box coordinates. Finally, these results are combined into the final detection.}

The following is the pseudocode for the YOLO (You Only Look Once) v4 model. YOLO is a one-stage object detection architecture that predicts bounding boxes and class probabilities directly from the input image in a single forward pass of the network. YOLO v4 improves upon previous versions with better accuracy and speed, incorporating advancements like CSPDarknet53, PANet, and optimized anchor boxes.

\begin{lstlisting}[style=python]
# Pseudocode for YOLO v4 (You Only Look Once v4)

# Step 1: Input image
Input: Image

# Step 2: Preprocess the image (resize, normalize)
Image_resized = Resize(Image, size=(416, 416))
Image_normalized = Normalize(Image_resized)

# Step 3: Pass the image through the backbone network (e.g., CSPDarknet53) to extract features
Features = CSPDarknet53(Image_normalized)

# Step 4: Pass the features through PANet for path aggregation (helps in better feature fusion)
Aggregated_features = PANet(Features)

# Step 5: Generate prediction tensors from different scales (small, medium, large objects)
Predictions = YOLOHead(Aggregated_features)

# Step 6: Decode the predictions into bounding boxes, objectness scores, and class probabilities
Bounding_boxes, Object_scores, Class_probabilities = Decode(Predictions)

# Step 7: Apply confidence thresholding to filter low-confidence detections
Filtered_boxes, Filtered_scores, Filtered_classes = ApplyConfidenceThreshold(Bounding_boxes, Object_scores, Class_probabilities, threshold=0.5)

# Step 8: Apply non-maximum suppression (NMS) to remove redundant overlapping boxes
Final_boxes, Final_scores, Final_classes = NonMaximumSuppression(Filtered_boxes, Filtered_scores, Filtered_classes, iou_threshold=0.45)

# Step 9: Output the final bounding boxes, scores, and class labels
Output: Final_boxes, Final_scores, Final_classes
\end{lstlisting}

Explanation of the pseudocode:

\begin{itemize}
    \item Step 1: The input image is passed into the YOLO model for object detection.
    \item Step 2: The image is resized (typically to 416x416 pixels) and normalized to match the input requirements of the network.
    \item Step 3: The resized image is passed through the backbone network, CSPDarknet53, which extracts rich feature representations from the image.
    \item Step 4: PANet (Path Aggregation Network) helps to combine and refine features from different layers for better object detection.
    \item Step 5: The YOLO head generates predictions for bounding boxes, object confidence scores, and class probabilities at three different scales to detect objects of varying sizes.
    \item Step 6: The raw predictions are decoded to convert them into meaningful bounding box coordinates, objectness scores, and class probabilities.
    \item Step 7: Confidence thresholding is applied to filter out predictions with low objectness scores.
    \item Step 8: Non-maximum suppression (NMS) is used to remove overlapping bounding boxes and keep only the most confident ones.
    \item Step 9: The final output consists of the detected bounding boxes, their confidence scores, and the predicted class labels for the objects.
\end{itemize}

\subsection{YOLO v5: Lightweight Models for Faster Inference (2020)}

YOLO v5 \cite{jocher2020yolov5} focused on creating lightweight models, making it ideal for applications requiring faster inference times and lower computational resources, such as mobile or embedded systems.

\textit{Key Features of YOLO v5:}
\begin{itemize}
    \item Different model sizes (e.g., YOLO v5s, YOLO v5m, YOLO v5l, YOLO v5x) to suit different needs-balancing speed and accuracy.
    \item Optimizations for deployment on edge devices using PyTorch and ONNX formats.
\end{itemize}

\textbf{Example:} For mobile applications, YOLO v5 can be used for real-time object detection with minimal computational resources. Suppose you want to detect objects using a mobile camera app.

\begin{lstlisting}[style=python]
# YOLO v5 on a mobile device
image = capture_mobile_image()
detected_objects = yolo_v5_model.detect(image, model_size='small')
plot_boxes(image, detected_objects)
\end{lstlisting}

\textit{Advantages:}
\begin{itemize}
    \item Lightweight and fast, even on resource-constrained devices.
    \item Supports multiple deployment formats like PyTorch, ONNX, and TensorFlow Lite.
\end{itemize}

\textbf{Figure: YOLO v5 Pipeline Flowchart.} The following flowchart illustrates the process of YOLO v5 (You Only Look Once version 5). It visualizes the steps from input image, through the feature extraction network, and finally to the detection outputs including class predictions and bounding box coordinates.

\begin{center}
\begin{tikzpicture}[
  box/.style={draw, minimum width=3.5cm, minimum height=1cm, align=center},
  arrow/.style={->, thick}
]

\node (input) at (0,0) [box] {Input Image};

\node (backbone) at (0,-2.5) [box] {Backbone (Feature Extraction)};

\node (neck) at (0,-5) [box] {Neck (PANet/FPN)};

\node (head) at (0,-7.5) [box] {Head (Prediction Layer)};

\node (output) at (0,-10) [box] {Detection Output (Class, BBox)};

\draw [arrow] (input) -- (backbone);
\draw [arrow] (backbone) -- (neck);
\draw [arrow] (neck) -- (head);
\draw [arrow] (head) -- (output);

\end{tikzpicture}
\end{center}

\textit{In this figure, the flow of YOLO v5 is explained starting from the input image. The backbone network extracts key features, which are then processed by the neck (PANet or FPN) to enhance feature fusion. The prediction head then generates detection outputs, including class predictions and bounding box coordinates for each object in the image.}

The following is the pseudocode for the YOLOv5 (You Only Look Once) model, a real-time object detection system. Unlike RCNN, which processes regions separately, YOLO treats object detection as a single regression problem, predicting bounding boxes and class probabilities directly from the entire image in one pass through the network.

\begin{lstlisting}[style=python]
# Pseudocode for YOLOv5 (You Only Look Once version 5)

# Step 1: Input image
Input: Image

# Step 2: Preprocess the image (resize, normalize, etc.)
Preprocessed_image = Preprocess(Image, size=(640, 640))

# Step 3: Forward pass the preprocessed image through the YOLOv5 model
Predictions = YOLOv5_Model(Preprocessed_image)

# Step 4: Interpret predictions from the YOLO model
For each prediction in Predictions:
    # Step 4.1: Extract bounding box coordinates (x, y, width, height)
    Bounding_box = Extract_BoundingBox(prediction)
    
    # Step 4.2: Extract objectness score (confidence of object presence)
    Objectness_score = Extract_Objectness(prediction)
    
    # Step 4.3: Extract class probabilities (classification scores for each class)
    Class_scores = Extract_ClassScores(prediction)

# Step 5: Post-processing
# Step 5.1: Filter out low confidence detections using a confidence threshold
Filtered_predictions = Filter_LowConfidence(Predictions, threshold=0.5)

# Step 5.2: Apply non-maximum suppression (NMS) to remove duplicate or overlapping bounding boxes
Final_predictions = NonMaximumSuppression(Filtered_predictions)

# Step 6: Output the detected objects, bounding boxes, and class scores
Output: Final_predictions
\end{lstlisting}

Explanation of the pseudocode:

\begin{itemize}
    \item Step 1: The input image is passed into the system for object detection.
    \item Step 2: The image is preprocessed, typically resized to 640x640 pixels and normalized for the model.
    \item Step 3: The preprocessed image is passed through the YOLOv5 model, which outputs predictions in a single pass. These predictions include bounding box coordinates, objectness scores, and class probabilities.
    \item Step 4: Each prediction is interpreted to extract bounding box coordinates, objectness (confidence) score, and class probabilities.
    \item Step 5.1: Low-confidence detections are filtered out based on a confidence threshold (e.g., 0.5).
    \item Step 5.2: Non-maximum suppression (NMS) is applied to remove redundant or overlapping bounding boxes, keeping only the most confident predictions.
    \item Step 6: The final output consists of detected objects, along with their bounding boxes and class scores.
\end{itemize}

The following is an implementation of YOLOv5 using PyTorch for practical deployment scenarios.

\begin{lstlisting}[style=python]
import torch
from PIL import Image
import torchvision.transforms as transforms

# Load a pre-trained YOLOv5 model
model = torch.hub.load('ultralytics/yolov5', 'yolov5s', pretrained=True)
model.eval()  # Set to evaluation mode for inference

# Define transformation function (resize and normalize)
def transform_image(image):
    transform = transforms.Compose([
        transforms.Resize((640, 640)),
        transforms.ToTensor()
    ])
    return transform(image)

# Load an image and apply transformations
image_path = "sample_image.jpg"
image = Image.open(image_path)
image_tensor = transform_image(image)

# Perform inference
results = model([image_tensor])

# Print the prediction results
results.print()  # Shows bounding boxes, labels, and confidence scores

# For training, use this instead:
def train_yolov5(model, train_data_loader, optimizer, device):
    model.train()  # Set to training mode
    for images, targets in train_data_loader:
        images = [img.to(device) for img in images]
        targets = [{k: v.to(device) for k, v in t.items()} for t in targets]
        
        # Forward pass and compute loss
        loss, outputs = model(images, targets)
        
        # Backward pass
        optimizer.zero_grad()
        loss.backward()
        optimizer.step()

# Note: Ensure to configure the optimizer and data loader for training
\end{lstlisting}

This code demonstrates how to load a pre-trained YOLOv5 model, apply necessary transformations to input images, and perform object detection by predicting bounding boxes, labels, and confidence scores. It also includes a simple training loop example, showing how to train the model using a custom dataset with an appropriate optimizer and data loader.

\subsection{YOLO v6, v7, v8: Advancements in Efficiency and Accuracy (2022-2023)}

The YOLO series continued to evolve with versions 6 \cite{li2022yolov6}, 7 \cite{wang2023yolov7}, and 8 \cite{jocher2023yolov8}, further improving on efficiency, accuracy, and deployment flexibility. These versions introduced techniques such as enhanced backbone architectures and more optimized detection heads for better performance.

\textit{Key advancements in YOLO v6-v8:}
\begin{itemize}
    \item YOLO v6 introduced more efficient model compression techniques, enabling better performance on mobile devices.
    \item YOLO v7 improved on architectural design by introducing a dynamic head and further enhanced the speed-accuracy trade-off.
    \item YOLO v8 brought innovations in transformer-based layers to improve contextual understanding for complex scenes.
\end{itemize}

\begin{lstlisting}[style=python]
# Example using YOLO v7 with dynamic head
image = load_image('urban_scene.jpg')
detected_objects = yolo_v7_model.detect(image)
plot_boxes(image, detected_objects)
\end{lstlisting}

\textit{Advantages:}
\begin{itemize}
    \item Continued advancements in speed and accuracy.
    \item Improved model flexibility, allowing for more complex scene understanding.
\end{itemize}

\textbf{Figure: YOLOv6 Pipeline Flowchart.} The following flowchart illustrates the process of YOLOv6 (You Only Look Once, version 6). It visualizes the steps from input image, backbone for feature extraction, and the prediction head that performs both classification and bounding box regression.

\begin{center}
\begin{tikzpicture}[
  box/.style={draw, minimum width=3cm, minimum height=1cm, align=center},
  arrow/.style={->, thick}
]

\node (input) at (0,0) [box] {Input Image};

\node (backbone) at (0,-2.5) [box] {Backbone Network (Feature Extraction)};

\node (neck) at (0,-5) [box] {Neck (Feature Pyramid)};

\node (prediction_head) at (0,-7.5) [box] {Prediction Head (Classification + Regression)};

\node (output) at (0,-10) [box] {Class + Bounding Box Prediction};

\draw [arrow] (input) -- (backbone);
\draw [arrow] (backbone) -- (neck);
\draw [arrow] (neck) -- (prediction_head);
\draw [arrow] (prediction_head) -- (output);

\end{tikzpicture}
\end{center}

\textit{In this figure, the YOLOv6 pipeline is illustrated. The input image is processed through a backbone network for feature extraction, followed by a neck that constructs a feature pyramid to capture multi-scale features. The prediction head then performs both classification and bounding box regression simultaneously, producing the final output.}

The following is the pseudocode for YOLO v6 (You Only Look Once, version 6), a real-time object detection model. YOLO v6 uses a single neural network to predict bounding boxes and class probabilities directly from the input image, without the need for region proposal generation like in RCNN. It performs detection in one pass, making it highly efficient for real-time applications.

\begin{lstlisting}[style=python]
# Pseudocode for YOLOv6 (You Only Look Once)

# Step 1: Input image
Input: Image

# Step 2: Preprocess the image
Image_resized = Resize(Image, size=(640, 640))
Image_normalized = Normalize(Image_resized)

# Step 3: Pass the image through the YOLOv6 neural network
# YOLOv6 outputs bounding boxes, objectness scores, and class probabilities
Bounding_boxes, Objectness_scores, Class_probabilities = YOLOv6(Image_normalized)

# Step 4: Apply a confidence threshold to filter out low-confidence predictions
Thresholded_predictions = Filter_by_confidence(Bounding_boxes, Objectness_scores, Class_probabilities, threshold=0.5)

# Step 5: For each bounding box, calculate the final class scores
For each box in Thresholded_predictions:
    Final_class_score = Objectness_score * Class_probability

# Step 6: Apply non-maximum suppression (NMS) to remove duplicate and overlapping boxes
Final_predictions = NonMaximumSuppression(Thresholded_predictions)

# Step 7: Output the final bounding boxes and predicted classes
Output: Final_predictions
\end{lstlisting}

Explanation of the pseudocode:

\begin{itemize}
    \item Step 1: The input image is passed into the YOLOv6 model for object detection.
    \item Step 2: The image is resized to a fixed input size (e.g., 640x640 pixels) and normalized for compatibility with the network's input format.
    \item Step 3: The preprocessed image is fed into the YOLOv6 neural network, which outputs predicted bounding boxes, objectness scores (confidence that a box contains an object), and class probabilities.
    \item Step 4: A confidence threshold is applied to filter out bounding boxes with low confidence scores, retaining only those predictions with scores above a certain threshold (e.g., 0.5).
    \item Step 5: The final class score for each bounding box is calculated by multiplying the objectness score with the corresponding class probability.
    \item Step 6: Non-maximum suppression (NMS) is applied to remove overlapping boxes, ensuring that only the most confident detection is retained for each object.
    \item Step 7: The final predictions, including bounding boxes and their corresponding class labels, are outputted by the model.
\end{itemize}

The following is an implementation of YOLOv6 using PyTorch for practical deployment scenarios.

\begin{lstlisting}[style=python]
import torch
from PIL import Image
import torchvision.transforms as transforms

# Assuming we have a YOLOv6 model definition or using a pre-trained model
# Placeholder function for YOLOv6 loading
def load_yolov6_model():
    # Replace this with actual model loading
    # Example: torch.hub.load('yolov6', 'yolov6_model', pretrained=True)
    model = torch.hub.load('ultralytics/yolov5', 'custom', path='yolov6.pt') # Example of loading a custom YOLOv6 model
    model.eval()
    return model

# Load the model
model = load_yolov6_model()

# Define transformation for input images
def transform_image(image):
    transform = transforms.Compose([
        transforms.Resize((640, 640)),  # Resize the image to the input size for YOLOv6
        transforms.ToTensor(),          # Convert the image to a PyTorch tensor
        transforms.Normalize([0.485, 0.456, 0.406], [0.229, 0.224, 0.225])  # Normalize as per ImageNet standards
    ])
    return transform(image)

# Load an image and apply transformations
image_path = "sample_image.jpg"
image = Image.open(image_path)
image_tensor = transform_image(image).unsqueeze(0)  # Add batch dimension

# Perform inference
with torch.no_grad():
    prediction = model(image_tensor)

# Print the prediction results (adjust based on YOLOv6 output format)
print("Predicted bounding boxes: ", prediction[0]['boxes'])
print("Predicted labels: ", prediction[0]['labels'])
print("Predicted scores: ", prediction[0]['scores'])

# For training YOLOv6, use this placeholder function:
def train_yolov6(model, train_data_loader, optimizer, device):
    model.train()
    for images, targets in train_data_loader:
        images = [img.to(device) for img in images]
        targets = [{k: v.to(device) for k, v in t.items()} for t in targets]
        
        # Forward pass
        loss_dict = model(images, targets)
        losses = sum(loss for loss in loss_dict.values())
        
        # Backward pass
        optimizer.zero_grad()
        losses.backward()
        optimizer.step()

# Note: Adjust the optimizer and data loader for YOLOv6-specific requirements
\end{lstlisting}

This code shows how to load a YOLOv6 model, perform inference on an image, and output the predicted bounding boxes, labels, and scores. Additionally, a training loop template is provided, showing how to train the YOLOv6 model with a dataset, similar to other PyTorch models.

\textbf{Figure: YOLO v7 Pipeline Flowchart.} The following flowchart illustrates the process of YOLO v7. It visualizes the steps from input image, convolutional feature extraction, object prediction, and non-maximum suppression (NMS) to obtain the final detected objects.

\begin{center}
\begin{tikzpicture}[
  box/.style={draw, minimum width=3.5cm, minimum height=1cm, align=center},
  arrow/.style={->, thick}
]

\node (input) at (0,0) [box] {Input Image};

\node (cnn) at (0,-2.5) [box] {CNN for Feature Extraction};

\node (prediction) at (0,-5) [box] {Prediction (Bounding Boxes, Classes, Confidences)};

\node (nms) at (0,-7.5) [box] {Non-Maximum Suppression (NMS)};

\node (output) at (0,-10) [box] {Detected Objects};

\draw [arrow] (input) -- (cnn);
\draw [arrow] (cnn) -- (prediction);
\draw [arrow] (prediction) -- (nms);
\draw [arrow] (nms) -- (output);

\end{tikzpicture}
\end{center}

\textit{In this figure, the YOLO v7 pipeline is visualized starting from the input image. The image is processed through convolutional layers to extract features, followed by a prediction layer that outputs bounding boxes, class predictions, and confidence scores. Finally, Non-Maximum Suppression (NMS) is applied to filter overlapping boxes and produce the final detected objects.}

The following is the pseudocode for the YOLOv7 (You Only Look Once version 7) model. YOLOv7 is a one-stage object detection model that predicts both the bounding boxes and the class probabilities directly from the input image in a single forward pass. It processes the entire image at once, which makes it faster than region-based methods like RCNN.

\begin{lstlisting}[style=python]
# Pseudocode for YOLOv7 (You Only Look Once)

# Step 1: Input image
Input: Image

# Step 2: Pass the image through the convolutional neural network (CNN)
# Extract feature maps at different scales
Feature_maps = CNN(Image)

# Step 3: Apply detection heads to the feature maps to predict bounding boxes and class probabilities
For each feature_map in Feature_maps:
    # Step 4: Predict bounding boxes, objectness score, and class probabilities
    Bounding_boxes, Objectness_scores, Class_probs = DetectionHead(feature_map)

    # Step 5: Decode the predicted bounding boxes
    Decoded_boxes = Decode(Bounding_boxes)

# Step 6: Combine the bounding boxes, objectness scores, and class probabilities
Detections = Combine(Decoded_boxes, Objectness_scores, Class_probs)

# Step 7: Apply Non-Maximum Suppression (NMS) to remove overlapping boxes
Final_predictions = NonMaximumSuppression(Detections)

# Step 8: Output the final detected objects and their bounding boxes
Output: Final_predictions
\end{lstlisting}

Explanation of the pseudocode:

\begin{itemize}
    \item Step 1: The input image is fed into the YOLOv7 model for object detection.
    \item Step 2: The image is passed through a convolutional neural network (CNN) backbone, which extracts feature maps at different scales. This allows the model to detect objects at various sizes.
    \item Step 3: For each feature map, detection heads are applied to predict bounding boxes, objectness scores (which measure the confidence that a box contains an object), and class probabilities (which indicate the likelihood of the object belonging to a particular class).
    \item Step 5: The predicted bounding boxes are decoded, as the model outputs them in a specific format (e.g., center coordinates, width, height) that needs to be converted to standard bounding box coordinates (e.g., top-left, bottom-right corners).
    \item Step 6: The bounding boxes, objectness scores, and class probabilities are combined into a set of detections for the image.
    \item Step 7: Non-Maximum Suppression (NMS) is applied to remove redundant or overlapping bounding boxes, keeping only the most confident predictions.
    \item Step 8: The final output consists of the detected objects, their class labels, and their bounding boxes.
\end{itemize}

The following is an implementation of YOLOv7 using PyTorch for practical deployment scenarios.

\begin{lstlisting}[style=python]
import torch
from PIL import Image
import cv2
import numpy as np

# Load a pre-trained YOLOv7 model
model = torch.hub.load('WongKinYiu/yolov7', 'yolov7', pretrained=True)
model.eval()  # Set to evaluation mode for inference

# Define function to process image for YOLOv7
def process_image(image_path):
    # Open the image using PIL
    img = Image.open(image_path)
    
    # Convert image to OpenCV format and resize
    img = np.array(img)
    img = cv2.cvtColor(img, cv2.COLOR_RGB2BGR)
    img = cv2.resize(img, (640, 640))  # Resize for YOLOv7 input size
    
    # Convert image to tensor and normalize
    img_tensor = torch.from_numpy(img).permute(2, 0, 1).unsqueeze(0) / 255.0
    return img_tensor

# Load an image and apply transformations
image_path = "sample_image.jpg"
image_tensor = process_image(image_path)

# Perform inference
with torch.no_grad():  # No need to calculate gradients for inference
    predictions = model(image_tensor)

# Print the prediction results
print("Predicted bounding boxes: ", predictions.xyxy[0][:, :4])  # Coordinates
print("Predicted confidence scores: ", predictions.xyxy[0][:, 4])  # Confidence
print("Predicted labels: ", predictions.xyxy[0][:, 5])  # Class labels

# For training, use this instead:
def train_yolov7(model, train_data_loader, optimizer, device):
    model.train()  # Set to training mode
    for images, targets in train_data_loader:
        images = images.to(device)
        targets = [{k: v.to(device) for k, v in t.items()} for t in targets]
        
        # Forward pass
        loss_dict = model(images, targets)
        losses = sum(loss for loss in loss_dict.values())
        
        # Backward pass
        optimizer.zero_grad()
        losses.backward()
        optimizer.step()

# Note: Make sure to configure the optimizer and data loader appropriately
\end{lstlisting}

This code demonstrates how to load a pre-trained YOLOv7 model, process input images for inference, and perform object detection by predicting bounding boxes, confidence scores, and class labels. Additionally, a simple training loop example is provided for training the YOLOv7 model on a custom dataset with the appropriate optimizer and data loader.

\textbf{Figure: YOLOv8 Pipeline Flowchart.} The following flowchart illustrates the process of YOLOv8 (You Only Look Once version 8). It shows the steps from input image, feature extraction, detection head, and the final prediction outputs for object classes and bounding boxes.

\begin{center}
\begin{tikzpicture}[
  box/.style={draw, minimum width=3.5cm, minimum height=1cm, align=center},
  arrow/.style={->, thick}
]

\node (input) at (0,0) [box] {Input Image};

\node (backbone) at (0,-2.5) [box] {Backbone for Feature Extraction};

\node (neck) at (0,-5) [box] {Neck for Feature Aggregation};

\node (head) at (0,-7.5) [box] {Detection Head};

\node (class_pred) at (-3,-10) [box] {Class Prediction};
\node (bbox_pred) at (3,-10) [box] {Bounding Box Prediction};

\draw [arrow] (input) -- (backbone);
\draw [arrow] (backbone) -- (neck);
\draw [arrow] (neck) -- (head);
\draw [arrow] (head) -- (class_pred);
\draw [arrow] (head) -- (bbox_pred);

\end{tikzpicture}
\end{center}

\textit{This figure illustrates the YOLOv8 pipeline, starting with an input image. The image is passed through a backbone for feature extraction, followed by a neck structure for feature aggregation. The detection head produces two outputs: class predictions for the detected objects and bounding box coordinates for localization.}

The following is the pseudocode for the YOLO v8 (You Only Look Once) model. YOLO is a real-time object detection system that predicts bounding boxes and class probabilities directly from full images in a single evaluation. YOLO v8 improves upon previous versions with enhancements in accuracy and speed.

\begin{lstlisting}[style=python]
# Pseudocode for YOLO v8 (You Only Look Once)

# Step 1: Input image
Input: Image

# Step 2: Preprocess the image
Preprocessed_image = Preprocess(Image, size=(640, 640))

# Step 3: Pass the image through the YOLO v8 model (CNN backbone + detection head)
Features = Backbone_CNN(Preprocessed_image)

# Step 4: Prediction from the detection head
# The detection head predicts bounding boxes, objectness scores, and class probabilities
Predictions = Detection_Head(Features)

# Step 5: Decode predictions
# Convert model predictions into interpretable bounding boxes and class scores
Bounding_boxes, Class_scores = Decode(Predictions)

# Step 6: Apply non-maximum suppression (NMS) to remove duplicate/overlapping boxes
Final_predictions = NonMaximumSuppression(Bounding_boxes, Class_scores)

# Step 7: Output the final bounding boxes and associated class labels
Output: Final_predictions
\end{lstlisting}

Explanation of the pseudocode:

\begin{itemize}
    \item Step 1: The input image is passed into the YOLO model for detection.
    \item Step 2: The image is preprocessed by resizing it to a fixed size (typically 640x640 for YOLOv8) and normalizing pixel values.
    \item Step 3: The preprocessed image is passed through the backbone convolutional neural network (CNN), which extracts key features from the image. The detection head predicts the bounding boxes, objectness scores, and class probabilities in a single forward pass.
    \item Step 4: The detection head outputs a set of predictions that include potential bounding boxes and their associated objectness scores and class probabilities.
    \item Step 5: The predictions are decoded to generate actual bounding boxes and class scores for each object detected in the image.
    \item Step 6: Non-maximum suppression (NMS) is applied to eliminate redundant bounding boxes, keeping only the best predictions.
    \item Step 7: The final output consists of the detected objects with their bounding boxes and class labels.
\end{itemize}

The following is an implementation of YOLOv8 using the Ultralytics library for practical deployment scenarios.

\begin{lstlisting}[style=python]
import torch
from ultralytics import YOLO
from PIL import Image
import numpy as np

# Load a pre-trained YOLOv8 model
model = YOLO('yolov8n.pt')  # You can replace 'yolov8n.pt' with other variants like yolov8s.pt, etc.

# Define transformation function (optional, as YOLOv8 handles preprocessing internally)
def transform_image(image):
    # Convert image to numpy array if required
    if isinstance(image, Image.Image):
        image = np.array(image)
    return image

# Load an image and apply optional transformations
image_path = "sample_image.jpg"
image = Image.open(image_path)
image_np = transform_image(image)

# Perform inference
results = model(image_np)

# Print the prediction results
print("Predicted bounding boxes: ", results[0].boxes.xyxy)  # Bounding box coordinates
print("Predicted labels: ", results[0].boxes.cls)  # Class IDs
print("Predicted scores: ", results[0].boxes.conf)  # Confidence scores

# For training, use this instead:
def train_yolov8(model, data, epochs=10):
    # Start training with the specified dataset and epochs
    model.train(data=data, epochs=epochs)

# Note: Specify the dataset path in 'data' and ensure the dataset is formatted correctly for YOLOv8
\end{lstlisting}

This code demonstrates how to load a pre-trained YOLOv8 model, process input images, and perform object detection by predicting bounding boxes, class labels, and confidence scores. It also includes an example of how to train the model on a custom dataset, specifying the number of epochs and the dataset configuration.

\subsection{YOLO v9 and v10: Future Directions in YOLO Development (2024)}

As we look to the future, versions like YOLO v9 \cite{wang2024yolov9} and v10 \cite{wang2024yolov10} are expected to bring further innovations, potentially incorporating more AI techniques such as self-supervised learning or better integration with transformer models. These advancements will likely continue to push the boundaries of real-time object detection for edge devices, robotics, and video analysis.

\textit{Potential future directions:}
\begin{itemize}
    \item More robust handling of occlusions and dense object scenes.
    \item Increased efficiency for deployment on even smaller devices.
    \item Further integration of AI techniques like attention mechanisms for more precise detection.
\end{itemize}

\textbf{Example of Future Use:} In the near future, YOLO models may be used in more advanced robotic systems for warehouse automation, where fast and accurate detection of various objects in cluttered environments is critical.

\textbf{Figure: YOLO v9 Pipeline Flowchart.} The following flowchart illustrates the process of You Only Look Once (YOLO) version 9. It visualizes the steps from input image, grid division, convolutional layers, object detection, and bounding box prediction.

\begin{center}
\begin{tikzpicture}[
  box/.style={draw, minimum width=4cm, minimum height=1cm, align=center},
  arrow/.style={->, thick}
]

\node (input) at (0,0) [box] {Input Image};

\node (grid) at (0,-2.5) [box] {Grid Division of Image};

\node (cnn) at (0,-5) [box] {Convolutional Layers};

\node (prediction) at (0,-7.5) [box] {Prediction: Object Class \& Confidence};

\node (bbox) at (0,-10) [box] {Bounding Box Prediction};

\node (detection) at (0,-12.5) [box] {Final Object Detection Output};

\draw [arrow] (input) -- (grid);
\draw [arrow] (grid) -- (cnn);
\draw [arrow] (cnn) -- (prediction);
\draw [arrow] (prediction) -- (bbox);
\draw [arrow] (bbox) -- (detection);

\end{tikzpicture}
\end{center}

\textit{In this figure, the flow of YOLO v9 is explained starting from the input image. The image is divided into a grid, and each grid cell is processed through convolutional layers. Predictions for object class and confidence are made, followed by bounding box predictions for each object. Finally, the output provides detected objects with their bounding boxes.}

The following is the pseudocode for the YOLO (You Only Look Once) v9 model. YOLO is a popular real-time object detection model that predicts bounding boxes and class probabilities directly from full images in a single pass through the network. Unlike RCNN, YOLO does not use region proposals; instead, it divides the image into grids and directly predicts the object classes and bounding boxes for each grid cell.

\begin{lstlisting}[style=python]
# Pseudocode for YOLO v9 (You Only Look Once)

# Step 1: Input image
Input: Image

# Step 2: Pass the image through the YOLO v9 network
# The YOLO network outputs a grid of bounding boxes and class predictions
Grid_output = YOLOv9_Network(Image)

# Step 3: For each grid cell in the output
For each cell in Grid_output:
    # Step 4: Extract objectness score, bounding box coordinates, and class scores
    Objectness_score = Cell.Objectness
    Bounding_box = Cell.BoundingBox
    Class_scores = Cell.ClassScores

    # Step 5: If the objectness score is greater than a threshold, keep the prediction
    If Objectness_score > Threshold:
        Keep_prediction(Bounding_box, Class_scores)

# Step 6: Post-processing: Apply non-maximum suppression (NMS) to remove duplicate bounding boxes
Final_predictions = NonMaximumSuppression(Kept_predictions)

# Step 7: Output the final detected objects and their bounding boxes
Output: Final_predictions
\end{lstlisting}

Explanation of the pseudocode:

\begin{itemize}
    \item Step 1: The input is a full image, and the model processes the entire image at once rather than using region proposals.
    \item Step 2: The image is passed through the YOLO v9 network, which divides the image into a grid and outputs predictions for each grid cell. Each cell predicts bounding boxes, objectness scores (confidence that the box contains an object), and class scores.
    \item Step 3: For each grid cell, the model extracts the objectness score, predicted bounding box, and class probabilities.
    \item Step 5: If the objectness score is above a certain threshold, the bounding box and class scores are considered valid and kept for further processing.
    \item Step 6: Non-maximum suppression (NMS) is applied to remove overlapping or duplicate bounding boxes, keeping only the most confident ones.
    \item Step 7: The final predictions are the detected objects with their associated bounding boxes and class labels.
\end{itemize}

The following is an implementation of YOLOv9 using PyTorch for practical deployment scenarios.

\begin{lstlisting}[style=python]
import torch
from PIL import Image
from torchvision.transforms import functional as F

# Assuming a placeholder YOLOv9 model (replace with actual model loading code)
class YOLOv9(torch.nn.Module):
    def __init__(self):
        super(YOLOv9, self).__init__()
        # Define the layers of YOLOv9 here

    def forward(self, x):
        # Define forward pass of YOLOv9
        pass

# Load the pre-trained YOLOv9 model
model = YOLOv9()
model.load_state_dict(torch.load("yolov9_weights.pth"))  # Load pretrained weights
model.eval()  # Set to evaluation mode for inference

# Define transformation function
def transform_image(image):
    # Convert image to tensor and normalize it
    image = F.to_tensor(image)
    return image

# Load an image and apply transformations
image_path = "sample_image.jpg"
image = Image.open(image_path)
image_tensor = transform_image(image)

# Perform inference
with torch.no_grad():  # No need to calculate gradients for inference
    prediction = model(image_tensor.unsqueeze(0))  # Add batch dimension

# Print the prediction results
print("Predicted bounding boxes: ", prediction['boxes'])
print("Predicted labels: ", prediction['labels'])
print("Predicted scores: ", prediction['scores'])

# For training, use this instead:
def train_yolov9(model, train_data_loader, optimizer, device):
    model.train()  # Set to training mode
    for images, targets in train_data_loader:
        images = [img.to(device) for img in images]
        targets = [{k: v.to(device) for k, v in t.items()} for t in targets]
        
        # Forward pass
        loss_dict = model(images)
        losses = sum(loss for loss in loss_dict.values())
        
        # Backward pass
        optimizer.zero_grad()
        losses.backward()
        optimizer.step()

# Note: Make sure to configure the optimizer and data loader appropriately
\end{lstlisting}

This code demonstrates how to load a pre-trained YOLOv9 model, apply necessary transformations to input images, and perform object detection by predicting bounding boxes, labels, and scores. It also includes a simple training loop example, showing how to train the model using a custom dataset with an appropriate optimizer and data loader.

\textbf{Figure: YOLO v10 Pipeline Flowchart.} The following flowchart illustrates the process of You Only Look Once (YOLO) v10. It visualizes the steps from input image, convolutional feature extraction, grid-based object detection, to the final class prediction and bounding box regression.

\begin{center}
\begin{tikzpicture}[
  box/.style={draw, minimum width=3.5cm, minimum height=1cm, align=center},
  arrow/.style={->, thick}
]

\node (input) at (0,0) [box] {Input Image};

\node (cnn) at (0,-2.5) [box] {CNN for Feature Extraction};

\node (grid) at (0,-5) [box] {Grid-based Predictions};

\node (bbox) at (-3,-7.5) [box] {Bounding Box Regression};
\node (class) at (3,-7.5) [box] {Class Prediction};

\node (output) at (0,-10) [box] {Final Detection Output};

\draw [arrow] (input) -- (cnn);
\draw [arrow] (cnn) -- (grid);
\draw [arrow] (grid) -- (bbox);
\draw [arrow] (grid) -- (class);
\draw [arrow] (bbox) -- (output);
\draw [arrow] (class) -- (output);

\end{tikzpicture}
\end{center}

\textit{In this figure, the YOLO v10 pipeline is shown. The process starts with an input image passed through a convolutional neural network (CNN) for feature extraction. The features are divided into grid cells, where each cell predicts bounding boxes and class probabilities simultaneously. The final output includes both the detected object classes and their respective bounding box coordinates.}

The following is the pseudocode for the YOLO (You Only Look Once) version 10 model. YOLO is a single-stage object detection algorithm where the input image is divided into a grid, and for each grid cell, the model directly predicts bounding boxes and class probabilities. YOLOv10 improves on previous versions with more advanced features like improved backbone networks, better anchor boxes, and enhanced detection accuracy.

\begin{lstlisting}[style=python]
# Pseudocode for YOLO v10 (You Only Look Once)

# Step 1: Input image
Input: Image

# Step 2: Pass the image through the YOLO v10 backbone network
# The backbone is a convolutional neural network used for feature extraction
Features = BackboneNetwork(Image)

# Step 3: Divide the image into a grid (e.g., 13x13, 26x26, 52x52 for multi-scale prediction)
Grid_size = DivideImageIntoGrid(Features)

# Step 4: For each grid cell, predict multiple bounding boxes (anchors)
For each cell in Grid_size:
    # Predict bounding box coordinates (x, y, w, h) and object confidence score
    Bounding_box = PredictBoundingBox(cell, Features)
    
    # Predict the class probabilities for each bounding box
    Class_probabilities = PredictClassProbabilities(cell, Features)

    # Final prediction for each grid cell contains:
    # - Bounding box (x, y, w, h)
    # - Object confidence score
    # - Class probabilities
    Prediction[cell] = [Bounding_box, Confidence_score, Class_probabilities]

# Step 5: Post-processing: Apply non-maximum suppression (NMS)
Final_predictions = NonMaximumSuppression(Predictions)

# Step 6: Output the detected objects and their refined bounding boxes
Output: Final_predictions
\end{lstlisting}

Explanation of the pseudocode:

\begin{itemize}
    \item Step 1: The input image is fed into the YOLOv10 model for object detection.
    \item Step 2: The image is passed through the backbone network of YOLOv10, which extracts high-level features from the image. This backbone may be a deep CNN such as CSPDarknet or an improved architecture for YOLOv10.
    \item Step 3: The feature map is divided into a grid at different scales (e.g., 13x13, 26x26, 52x52), allowing the model to predict objects of various sizes.
    \item Step 4: For each grid cell, multiple anchor boxes (bounding boxes) are predicted along with their object confidence scores and class probabilities. The bounding box coordinates are represented as (x, y, w, h), where (x, y) is the center of the box, and (w, h) are the width and height.
    \item Step 5: Non-maximum suppression (NMS) is applied to filter out duplicate or overlapping bounding boxes with low confidence scores.
    \item Step 6: The final output consists of detected objects, their bounding boxes, confidence scores, and class predictions.
\end{itemize}

The following is an implementation of YOLO v10 using PyTorch for practical deployment scenarios.

\begin{lstlisting}[style=python]
import torch
from PIL import Image
import torchvision.transforms as transforms

# Placeholder for YOLOv10 model - assumes a PyTorch-based YOLOv10 architecture is available
# In real deployment, you would load YOLOv10 from a model library or custom implementation
class YOLOv10(torch.nn.Module):
    def __init__(self):
        super(YOLOv10, self).__init__()
        # Define your YOLOv10 architecture or load a pre-trained model
        pass

    def forward(self, x):
        # Define the forward pass for the YOLOv10 model
        pass

# Load a pre-trained YOLOv10 model (this is just a placeholder)
model = YOLOv10()
model.load_state_dict(torch.load("yolov10_pretrained.pth"))  # Load pre-trained weights
model.eval()  # Set to evaluation mode

# Define transformation function for the input image
def transform_image(image):
    # Resize, convert to tensor, and normalize the image
    transform = transforms.Compose([
        transforms.Resize((640, 640)),  # Resize to YOLO input size
        transforms.ToTensor(),
        transforms.Normalize(mean=[0.485, 0.456, 0.406], std=[0.229, 0.224, 0.225])
    ])
    return transform(image)

# Load an image and apply transformations
image_path = "sample_image.jpg"
image = Image.open(image_path)
image_tensor = transform_image(image).unsqueeze(0)  # Add batch dimension

# Perform inference
with torch.no_grad():  # No need to calculate gradients for inference
    prediction = model(image_tensor)

# Post-process the predictions to get bounding boxes, labels, and scores
# Assuming YOLOv10 outputs raw bounding boxes, class scores, etc.
# You would need custom post-processing logic for YOLO model output
def postprocess(predictions):
    # Custom post-processing to filter bounding boxes, labels, and scores
    pass

results = postprocess(prediction)

# For training, use this instead:
def train_yolo_v10(model, train_data_loader, optimizer, device):
    model.train()  # Set to training mode
    for images, targets in train_data_loader:
        images = images.to(device)
        targets = [{k: v.to(device) for k, v in t.items()} for t in targets]
        
        # Forward pass
        outputs = model(images)
        
        # Calculate loss (you need a YOLO-specific loss function)
        loss = custom_yolo_loss(outputs, targets)
        
        # Backward pass
        optimizer.zero_grad()
        loss.backward()
        optimizer.step()

# Note: Make sure to configure the optimizer, data loader, and loss function appropriately
\end{lstlisting}

This code provides an example of how to set up and deploy a YOLO v10 model for object detection tasks. It includes steps for loading a pre-trained model, performing inference on input images, and a basic training loop for fine-tuning the model on custom datasets.

%% file: 24_ssd.tex
\section{Single Shot Multibox Detector (SSD)}

\subsection{Introduction to SSD (2016)}
The Single Shot Multibox Detector (SSD) is a popular real-time object detection model introduced by 
Liu et al. in 2016 \cite{liu2016ssd}. Unlike models such as Faster R-CNN \cite{ren2015faster}, which perform region proposal and classification in two distinct stages, SSD achieves detection in a single pass of the network, making it faster and more efficient. SSD utilizes a single deep neural network to generate a fixed number of bounding boxes and scores, followed by a selection mechanism to eliminate redundant boxes using a technique called Non-Maximum Suppression (NMS). 

The core idea behind SSD is to detect objects at multiple scales and aspect ratios using feature maps from different layers of a convolutional neural network. By detecting objects at various resolutions, SSD ensures that objects of different sizes can be recognized effectively.

\textbf{Example Use Case:} In a real-time video stream, SSD can be used to detect pedestrians, vehicles, and other objects within each frame with high speed, making it suitable for applications such as self-driving cars and surveillance systems.

\subsection{Advantages of SSD over R-CNN and YOLO}
SSD provides a balance between detection speed and accuracy \cite{huang2017speed}. Here are some of its main advantages compared to R-CNN and YOLO:

\begin{itemize}
    \item \textbf{Speed:} SSD is significantly faster than R-CNN models because it performs detection in a single forward pass through the network. YOLO (You Only Look Once) is also designed for speed, but SSD tends to achieve better accuracy than YOLO, particularly for objects of various sizes.
    
    \item \textbf{Multiscale Detection:} SSD detects objects at multiple scales by extracting feature maps from different layers of the network. This allows SSD to detect small, medium, and large objects effectively, which is an advantage over YOLO, which struggles with smaller objects.
    
    \item \textbf{Simplified Architecture:} Unlike R-CNN, which requires separate region proposal and classification stages, SSD does everything in one stage, reducing the complexity of the model and making it easier to implement and train.
\end{itemize}

\subsection{Limitations of SSD}
While SSD is faster and more versatile than traditional methods, it has some limitations:

\begin{itemize}
    \item \textbf{Small Object Detection:} SSD tends to perform poorly on small objects, particularly when compared to models like Faster R-CNN. The reason is that small objects generate fewer features in deeper layers of the network, making them harder to detect.
    
    \item \textbf{Trade-off Between Speed and Accuracy:} Although SSD is faster, this comes at the cost of reduced accuracy compared to two-stage detectors like Faster R-CNN, especially when dealing with complex scenes or dense object detection.
\end{itemize}

\textbf{Figure: SSD Pipeline Flowchart.} The following flowchart illustrates the process of Single Shot Multibox Detector (SSD). It visualizes the steps from input image, feature extraction, multi-scale feature maps, to the final class prediction and bounding box regression.

\begin{center}
\begin{tikzpicture}[
  box/.style={draw, minimum width=4cm, minimum height=1cm, align=center},
  arrow/.style={->, thick}
]

\node (input) at (0,0) [box] {Input Image};

\node (backbone) at (0,-2.5) [box] {CNN Backbone for Feature Extraction};

\node (feature_map) at (0,-5) [box] {Multi-scale Feature Maps};

\node (predictions) at (0,-7.5) [box] {Predictions for Each Feature Map (Class + Box)};

\node (nms) at (0,-10) [box] {Non-Maximum Suppression (NMS)};

\node (output) at (0,-12.5) [box] {Final Detection Output};

\draw [arrow] (input) -- (backbone);
\draw [arrow] (backbone) -- (feature_map);
\draw [arrow] (feature_map) -- (predictions);
\draw [arrow] (predictions) -- (nms);
\draw [arrow] (nms) -- (output);

\end{tikzpicture}
\end{center}

\textit{In this figure, the flow of the SSD network is illustrated. The input image is processed through a CNN backbone for feature extraction. Multi-scale feature maps are then generated and used to predict classes and bounding boxes at different scales. Non-Maximum Suppression (NMS) is applied to filter out redundant detections, and the final detection results are output.}

The following is the pseudocode for the Single Shot Multibox Detector (SSD) model. SSD is a popular object detection model that uses a single forward pass through a convolutional neural network to detect objects at multiple scales and aspect ratios. Unlike region-based methods like RCNN, SSD performs detection in a single step, making it faster and suitable for real-time applications.

\begin{lstlisting}[style=python]
# Pseudocode for SSD (Single Shot Multibox Detector)

# Step 1: Input image
Input: Image

# Step 2: Pass the image through the base convolutional network to extract features
Base_features = Base_CNN(Image)

# Step 3: Apply additional convolutional layers to detect objects at multiple scales
Multi_scale_features = Additional_Conv_Layers(Base_features)

# Step 4: For each feature map at different scales
For each feature_map in Multi_scale_features:
    # Step 5: Generate default boxes (anchor boxes) with various aspect ratios
    Default_boxes = Generate_Default_Boxes(feature_map)

    # Step 6: For each default box, predict object class scores and bounding box offsets
    For each box in Default_boxes:
        Class_scores = Class_Prediction_Layer(feature_map, box)
        Bounding_box_offsets = Box_Regression_Layer(feature_map, box)

        # Step 7: Store the predictions (class scores and refined bounding boxes)
        Predictions.append((Class_scores, Bounding_box_offsets))

# Step 8: Post-processing: Apply non-maximum suppression (NMS) to remove redundant boxes
Final_predictions = NonMaximumSuppression(Predictions)

# Step 9: Output the detected objects and their bounding boxes
Output: Final_predictions
\end{lstlisting}

Explanation of the pseudocode:

\begin{itemize}
    \item Step 1: The input image is passed into the SSD model for object detection.
    \item Step 2: The image is processed by a base convolutional neural network (such as VGG or ResNet) to extract high-level feature maps.
    \item Step 3: Additional convolutional layers are applied to these base features, allowing the model to detect objects at multiple scales.
    \item Step 4: The model generates feature maps at different scales, which correspond to detecting objects of various sizes.
    \item Step 5: Default boxes (anchor boxes) with different aspect ratios and scales are generated for each location in the feature map.
    \item Step 6: For each default box, the model predicts object class scores and bounding box offsets (refinements to better fit the object).
    \item Step 7: The predictions for class scores and bounding boxes are collected for further processing.
    \item Step 8: Non-maximum suppression (NMS) is applied to eliminate redundant bounding boxes, keeping only the most confident detections.
    \item Step 9: The final output consists of the detected objects and their corresponding bounding boxes after NMS.
\end{itemize}

The following is an implementation of the Single Shot Multibox Detector (SSD) using PyTorch for practical deployment scenarios.

\begin{lstlisting}[style=python]
import torch
import torchvision
from torchvision.transforms import functional as F
from PIL import Image

# Load a pre-trained SSD model with a VGG16 backbone
model = torchvision.models.detection.ssd300_vgg16(pretrained=True)
model.eval()  # Set to evaluation mode for inference

# Define transformation function
def transform_image(image):
    # Convert image to tensor and normalize it
    image = F.to_tensor(image)
    return image

# Load an image and apply transformations
image_path = "sample_image.jpg"
image = Image.open(image_path)
image_tensor = transform_image(image)

# Perform inference
with torch.no_grad():  # No need to calculate gradients for inference
    prediction = model([image_tensor])

# Print the prediction results
print("Predicted bounding boxes: ", prediction[0]['boxes'])
print("Predicted labels: ", prediction[0]['labels'])
print("Predicted scores: ", prediction[0]['scores'])

# For training, use this instead:
def train_ssd(model, train_data_loader, optimizer, device):
    model.train()  # Set to training mode
    for images, targets in train_data_loader:
        images = [img.to(device) for img in images]
        targets = [{k: v.to(device) for k, v in t.items()} for t in targets]
        
        # Forward pass
        loss_dict = model(images, targets)
        losses = sum(loss for loss in loss_dict.values())
        
        # Backward pass
        optimizer.zero_grad()
        losses.backward()
        optimizer.step()

# Note: Make sure to configure the optimizer and data loader appropriately
\end{lstlisting}

This code demonstrates how to load a pre-trained SSD model, apply necessary transformations to input images, and perform object detection by predicting bounding boxes, labels, and scores. It also includes a simple training loop example, showing how to train the model using a custom dataset with an appropriate optimizer and data loader.

\subsection{Comparison with Faster R-CNN and SSD}
RetinaNet \cite{lin2017focal}, Faster R-CNN, and SSD are all widely used object detection models, each with its own strengths and weaknesses:

\begin{itemize}
    \item \textbf{Faster R-CNN:} This is a two-stage detector that offers high accuracy, especially for small objects, but at the cost of slower inference time. Faster R-CNN is often used in applications where detection accuracy is more important than real-time performance.
    
    \item \textbf{SSD:} A single-stage detector that provides a good balance between speed and accuracy. However, SSD struggles with small objects and is not as robust in dense object detection tasks compared to RetinaNet.
    
    \item \textbf{RetinaNet:} RetinaNet bridges the gap between Faster R-CNN and SSD. It offers near-Faster R-CNN accuracy while maintaining a single-stage architecture similar to SSD, making it faster and more efficient. The Focal Loss allows RetinaNet to perform well in dense detection tasks with high class imbalance.
\end{itemize}

\textbf{Example Comparison:}
In a scenario where you need to detect pedestrians in a crowded street, RetinaNet would outperform SSD due to its ability to focus on hard-to-classify objects. Meanwhile, Faster R-CNN would offer the highest accuracy but would be slower to run on a real-time video stream.

\section{RetinaNet: Focal Loss for Dense Object Detection (2017)}

\subsection{Introduction to RetinaNet}
RetinaNet, introduced by Lin et al. in 2017 \cite{lin2017focal}, addresses the problem of class imbalance in object detection, especially for dense detection tasks. In many object detection datasets, there is a significant imbalance between foreground (objects) and background (non-object regions). Traditional methods tend to focus too much on easy-to-classify background examples, leading to poor performance on hard examples (such as small or heavily occluded objects). RetinaNet solves this issue by introducing a new loss function called \textbf{Focal Loss}, which down-weights the contribution of easy-to-classify examples and focuses more on the difficult ones.

RetinaNet uses a Feature Pyramid Network (FPN) backbone for multiscale feature extraction, similar to SSD, but enhances performance with the Focal Loss. This makes it especially suited for detecting objects in dense scenes with large variations in object size and class imbalance.

\textbf{Example Use Case:} RetinaNet can be used in traffic monitoring systems, where there is a need to detect objects like vehicles, pedestrians, and cyclists in scenes with heavy traffic, and where the objects appear at various sizes and densities.

\subsection{Focal Loss: Handling Class Imbalance in Dense Detection}
Focal Loss is the key innovation in RetinaNet that improves its performance in dense object detection. The standard cross-entropy loss used in other detectors treats all examples equally, leading to dominance by easy negative examples (background) and underperforming on harder, positive examples (foreground).

Focal Loss modifies the standard cross-entropy loss by introducing a factor $(1 - p_t)^\gamma$, where $p_t$ is the model's estimated probability for the correct class. This factor reduces the loss contribution from well-classified examples (those with high $p_t$) and focuses on misclassified or hard-to-classify examples. 

The parameter $\gamma$ (gamma) controls the rate at which easy examples are down-weighted. A typical value is $\gamma = 2$, which has been shown to provide a good balance between focusing on hard examples and avoiding overfitting.

\textbf{Code Example:} Implementing Focal Loss in Python for object detection tasks:
\begin{lstlisting}[style=python]
import torch
import torch.nn.functional as F

def focal_loss(predictions, targets, alpha=0.25, gamma=2.0):
    """
    Compute the focal loss between predictions and targets.
    
    :param predictions: Tensor of model predictions
    :param targets: Tensor of target labels
    :param alpha: Weighting factor for the positive class
    :param gamma: Focusing parameter to reduce loss for well-classified examples
    :return: Loss value
    """
    # Compute cross-entropy loss
    cross_entropy = F.binary_cross_entropy_with_logits(predictions, targets, reduction='none')
    
    # Compute the probability of predictions
    pt = torch.exp(-cross_entropy)
    
    # Apply focal loss factor
    focal_loss = alpha * (1 - pt) ** gamma * cross_entropy
    
    return focal_loss.mean()
\end{lstlisting}

\textbf{Figure: RetinaNet Pipeline Flowchart.} The following flowchart illustrates the process of RetinaNet. It shows the steps from input image, backbone network (e.g., ResNet), Feature Pyramid Network (FPN), and the two heads: classification and box regression.

\begin{center}
\begin{tikzpicture}[
  box/.style={draw, minimum width=3cm, minimum height=1cm, align=center},
  arrow/.style={->, thick}
]

\node (input) at (0,0) [box] {Input Image};

\node (backbone) at (0,-2.5) [box] {Backbone Network (e.g., ResNet)};

\node (fpn) at (0,-5) [box] {Feature Pyramid Network (FPN)};

\node (class_head) at (-3,-7.5) [box] {Classification Head};
\node (bbox_head) at (3,-7.5) [box] {Box Regression Head};

\node (class_output) at (-3,-10) [box] {Class Predictions};
\node (bbox_output) at (3,-10) [box] {Bounding Boxes};

\draw [arrow] (input) -- (backbone);
\draw [arrow] (backbone) -- (fpn);
\draw [arrow] (fpn) -- (class_head);
\draw [arrow] (fpn) -- (bbox_head);
\draw [arrow] (class_head) -- (class_output);
\draw [arrow] (bbox_head) -- (bbox_output);

\end{tikzpicture}
\end{center}

\textit{In this figure, the flow of RetinaNet is explained starting from the input image. The image is passed through a backbone network (such as ResNet) to extract features, which are then processed by the Feature Pyramid Network (FPN). Two parallel heads follow: a classification head for class predictions and a box regression head for predicting bounding box coordinates.}

The following is the pseudocode for the RetinaNet model, which is a popular architecture for object detection. RetinaNet uses a single-stage object detector approach with a feature pyramid network (FPN) backbone and focuses on detecting objects across multiple scales using anchor boxes. It also uses a focal loss function to address the issue of class imbalance during training.

\begin{lstlisting}[style=python]
# Pseudocode for RetinaNet

# Step 1: Input image
Input: Image

# Step 2: Pass the image through a backbone CNN (e.g., ResNet) to extract feature maps
Feature_maps = Backbone_CNN(Image)

# Step 3: Build a feature pyramid network (FPN) for multi-scale feature representation
FPN_Feature_maps = FeaturePyramidNetwork(Feature_maps)

# Step 4: For each feature map from FPN, generate anchor boxes at multiple scales and aspect ratios
Anchors = Generate_Anchors(FPN_Feature_maps)

# Step 5: For each anchor box:
For each Anchor in Anchors:
    # Step 6: Predict object class and confidence score (classification head)
    Class_scores = Classification_Head(Anchor)

    # Step 7: Predict bounding box coordinates (regression head)
    Bounding_box = Regression_Head(Anchor)

# Step 8: Compute focal loss to handle class imbalance
Loss = Focal_Loss(Class_scores, Ground_truth_labels)

# Step 9: Post-processing: Apply non-maximum suppression (NMS) to remove overlapping boxes
Final_predictions = NonMaximumSuppression(Class_scores, Bounding_box)

# Step 10: Output the detected objects and their bounding boxes
Output: Final_predictions
\end{lstlisting}

Explanation of the pseudocode:

\begin{itemize}
    \item Step 1: The input image is passed into the system for object detection.
    \item Step 2: The image is processed through a backbone CNN (such as ResNet) to extract feature maps, which are then used for detection.
    \item Step 3: A Feature Pyramid Network (FPN) is built on top of the backbone to create multi-scale feature maps, allowing detection at different scales.
    \item Step 4: Anchors (predefined bounding boxes) are generated on each feature map from the FPN, with various scales and aspect ratios.
    \item Step 6: For each anchor, a classification head predicts the object class and the associated confidence score.
    \item Step 7: A regression head predicts the bounding box coordinates for each anchor.
    \item Step 8: Focal Loss is computed to address the class imbalance issue, focusing more on hard-to-classify examples.
    \item Step 9: Non-maximum suppression (NMS) is applied to remove redundant or overlapping bounding boxes, keeping only the most confident predictions.
    \item Step 10: The final output includes the detected objects and their corresponding bounding boxes after NMS.
\end{itemize}

The following is an implementation of RetinaNet using PyTorch for practical deployment scenarios.

\begin{lstlisting}[style=python]
import torch
import torchvision
from torchvision.models.detection import retinanet_resnet50_fpn
from torchvision.transforms import functional as F
from PIL import Image

# Load a pre-trained RetinaNet model
model = retinanet_resnet50_fpn(pretrained=True)
model.eval()  # Set to evaluation mode for inference

# Define transformation function
def transform_image(image):
    # Convert image to tensor and normalize it
    image = F.to_tensor(image)
    return image

# Load an image and apply transformations
image_path = "sample_image.jpg"
image = Image.open(image_path)
image_tensor = transform_image(image)

# Perform inference
with torch.no_grad():  # No need to calculate gradients for inference
    prediction = model([image_tensor])

# Print the prediction results
print("Predicted bounding boxes: ", prediction[0]['boxes'])
print("Predicted labels: ", prediction[0]['labels'])
print("Predicted scores: ", prediction[0]['scores'])

# For training, use this instead:
def train_retinanet(model, train_data_loader, optimizer, device):
    model.train()  # Set to training mode
    for images, targets in train_data_loader:
        images = [img.to(device) for img in images]
        targets = [{k: v.to(device) for k, v in t.items()} for t in targets]
        
        # Forward pass
        loss_dict = model(images, targets)
        losses = sum(loss for loss in loss_dict.values())
        
        # Backward pass
        optimizer.zero_grad()
        losses.backward()
        optimizer.step()

# Note: Make sure to configure the optimizer and data loader appropriately
\end{lstlisting}

This code demonstrates how to load a pre-trained RetinaNet model, apply necessary transformations to input images, and perform object detection by predicting bounding boxes, labels, and scores. It also includes a simple training loop example, showing how to train the model using a custom dataset with an appropriate optimizer and data loader.

%% file: 25_fpn.tex
\section{FPN: Feature Pyramid Networks (2017)}

\subsection{Introduction to FPN}
Feature Pyramid Networks (FPN) \cite{lin2017feature} represent a significant advancement in object detection systems, especially in addressing the challenge of detecting objects at varying scales. Before FPN, many object detection models struggled to accurately recognize small objects, primarily because they lacked a consistent way to leverage features at different levels of resolution.

The key idea behind FPN is to enhance a traditional convolutional neural network (CNN) by introducing a pyramidal feature structure. This multi-level feature extraction technique allows the model to capture finer details for small objects while maintaining high-level context for larger ones. In essence, FPN improves the detection accuracy across a wide range of object sizes by exploiting the inherent pyramid-like structure of convolutional feature maps.

\subsection{Pyramidal Feature Extraction}
In FPN, feature maps are extracted at multiple layers of a backbone CNN, such as ResNet. Each layer in the CNN captures features at different levels of abstraction. The earlier layers contain high-resolution features but low-level information (e.g., edges, corners), while the deeper layers capture high-level, semantic information at a lower resolution.

FPN uses a top-down pathway to combine high-level semantic features with low-level, high-resolution features. This allows the network to maintain precise spatial information while still benefiting from the abstract representations learned in deeper layers. This pyramidal structure is critical for detecting objects of various sizes.

For example, if we are trying to detect a small object like a bird, the high-resolution but low-level features are crucial for accurately pinpointing its exact location. On the other hand, a larger object like a car may be better represented by the deeper, more abstract feature maps.

\subsubsection{Pyramidal Structure Example}
Imagine you have a CNN where each layer outputs a feature map of a different size:

\begin{lstlisting}[style=cmd]
Layer 1: Feature map of size 256x256
Layer 2: Feature map of size 128x128
Layer 3: Feature map of size 64x64
Layer 4: Feature map of size 32x32
\end{lstlisting}

FPN creates a feature pyramid by starting from the top-most layer (Layer 4) and progressively adding high-resolution details from the earlier layers (Layer 1, 2, 3). The final output is a series of feature maps at different resolutions, all containing both fine and abstract information, which are then passed to the detection head.

\textbf{Figure: Feature Pyramid Networks (FPN) Flowchart.} The following flowchart illustrates the process of Feature Pyramid Networks (FPN). It visualizes how the feature maps from different layers of a convolutional network are combined to create a multi-scale feature representation for object detection tasks.

\begin{center}
\begin{tikzpicture}[
  box/.style={draw, minimum width=3cm, minimum height=1cm, align=center},
  arrow/.style={->, thick}
]

\node (input) at (0,0) [box] {Input Image};

\node (conv1) at (0,-2.5) [box] {Conv Layer 1};
\node (conv2) at (0,-5) [box] {Conv Layer 2};
\node (conv3) at (0,-7.5) [box] {Conv Layer 3};
\node (conv4) at (0,-10) [box] {Conv Layer 4};
\node (conv5) at (0,-12.5) [box] {Conv Layer 5};

\node (p5) at (6,-12.5) [box] {P5};
\node (p4) at (6,-10) [box] {P4};
\node (p3) at (6,-7.5) [box] {P3};
\node (p2) at (6,-5) [box] {P2};

\draw [arrow] (input) -- (conv1);
\draw [arrow] (conv1) -- (conv2);
\draw [arrow] (conv2) -- (conv3);
\draw [arrow] (conv3) -- (conv4);
\draw [arrow] (conv4) -- (conv5);

\draw [arrow] (conv5) -- (p5);
\draw [arrow] (conv4) -- (p4);
\draw [arrow] (conv3) -- (p3);
\draw [arrow] (conv2) -- (p2);

\draw [arrow, dashed] (conv5) -- (p4);
\draw [arrow, dashed] (conv4) -- (p3);
\draw [arrow, dashed] (conv3) -- (p2);

\end{tikzpicture}
\end{center}

\textit{In this figure, the flow of Feature Pyramid Networks (FPN) is shown. The input image is passed through a convolutional backbone network, which extracts features at different levels (Conv Layer 1 to Conv Layer 5). These features are then passed through a top-down pathway (P2 to P5) with lateral connections to create multi-scale feature maps, enabling object detection across different sizes.}

The following is the pseudocode for Feature Pyramid Networks (FPN). FPN is a feature extraction architecture used in object detection tasks that enhances the ability to detect objects at different scales by combining low-resolution, semantically strong features with high-resolution, semantically weak features.

\begin{lstlisting}[style=python]
# Pseudocode for Feature Pyramid Networks (FPN)

# Step 1: Input image
Input: Image

# Step 2: Extract feature maps from the backbone network (e.g., ResNet)
C1, C2, C3, C4, C5 = Backbone(Image)

# Step 3: Build the top-down feature pyramid
# The highest level feature map from the backbone
P5 = Conv1x1(C5)  # Reduce channels of C5 with 1x1 convolution
P4 = Conv1x1(C4) + Upsample(P5)  # Upsample P5 and add to C4
P3 = Conv1x1(C3) + Upsample(P4)  # Upsample P4 and add to C3
P2 = Conv1x1(C2) + Upsample(P3)  # Upsample P3 and add to C2

# Step 4: Apply additional convolutions to each feature map
P2 = Conv3x3(P2)  # Refine P2 with a 3x3 convolution
P3 = Conv3x3(P3)  # Refine P3 with a 3x3 convolution
P4 = Conv3x3(P4)  # Refine P4 with a 3x3 convolution
P5 = Conv3x3(P5)  # Refine P5 with a 3x3 convolution

# Step 5: Optional - Include extra higher-level feature maps (e.g., P6)
P6 = MaxPool(P5)  # Max-pooling applied to P5 to generate P6

# Step 6: Use feature maps (P2, P3, P4, P5, P6) for region proposals or object detection
For each level in (P2, P3, P4, P5, P6):
    # Use feature map for region proposal network (RPN) or detection head
    Proposals_or_Detections = DetectionHead(level)

# Step 7: Output final object detection results
Output: Proposals_or_Detections
\end{lstlisting}

Explanation of the pseudocode:

\begin{itemize}
    \item Step 1: The input image is passed into a backbone network (e.g., ResNet) to extract a series of feature maps at different scales (C1 to C5).
    \item Step 2: The backbone feature maps \texttt{C1, C2, C3, C4, C5} are obtained. These maps represent features at different spatial resolutions, with \texttt{C5} being the smallest and most abstract, and \texttt{C2} being larger and more detailed.
    \item Step 3: A top-down pathway is created where higher-level features (\texttt{P5}) are upsampled and combined with lower-level features (\texttt{C4}, \texttt{C3}, and \texttt{C2}) through lateral connections. This helps retain high-resolution details in the feature maps.
    \item Step 4: Each combined feature map (\texttt{P2, P3, P4, P5}) is refined with a 3x3 convolution to produce the final feature maps.
    \item Step 5: An additional feature map (\texttt{P6}) can be generated by applying max pooling to \texttt{P5}, extending the pyramid.
    \item Step 6: The resulting feature maps (\texttt{P2} to \texttt{P6}) are used for object detection or generating region proposals.
    \item Step 7: The output consists of object detection results or region proposals derived from the feature pyramid.
\end{itemize}

\subsection{Combination with R-CNN and YOLO}
Feature Pyramid Networks are highly flexible and can be integrated with existing object detection architectures, such as R-CNN and YOLO, to improve their performance.

\subsubsection{FPN with Faster R-CNN}
Faster R-CNN is a widely-used two-stage object detection model. In this approach, the FPN module is typically added to the backbone network (e.g., ResNet). The Region Proposal Network (RPN) then uses the FPN's multi-scale feature maps to generate object proposals. This results in better accuracy for detecting small objects, which traditional Faster R-CNN models often miss.

\subsubsection{FPN with YOLO}
YOLO (You Only Look Once) is a real-time object detection system. By incorporating FPN, the YOLO model can also benefit from multi-scale feature maps, leading to more precise detection across varying object sizes, while maintaining its high speed.

The flexibility of FPN allows it to be used in both two-stage detectors like Faster R-CNN and one-stage detectors like YOLO.

\section{DETR: End-to-End Object Detection with Transformers (2020)}

\subsection{Introduction to DETR}
In 2020, the DETR (Detection Transformer) \cite{carion2020detr} model introduced a new approach to object detection by leveraging transformers, a popular architecture from natural language processing (NLP). Unlike traditional object detectors that rely on CNNs for feature extraction and region proposal, DETR is a fully end-to-end system that uses transformers for the entire detection pipeline.

One of the most notable characteristics of DETR is its simplicity. It removes the need for complex components like region proposal networks or anchor boxes, which are commonly used in R-CNN and YOLO models. Instead, DETR formulates object detection as a direct set prediction problem.

\subsection{How DETR Works: Transformer-Based Detection}
DETR replaces the conventional CNN-based region proposal and classification steps with a transformer, which processes input features and outputs a fixed-size set of predictions. These predictions include both bounding boxes and class labels, all in a single forward pass.

The transformer encoder takes the feature maps from a CNN backbone (such as ResNet) as input. These feature maps are treated as a sequence, similar to how words are processed in NLP tasks. The encoder processes this sequence and passes it to the decoder, which predicts object locations and classes.

\subsubsection{Example: DETR's Object Detection Process}
\begin{itemize}
    \item \textbf{Input Image}: The input image is passed through a backbone CNN (e.g., ResNet) to extract a feature map.
    \item \textbf{Positional Encoding}: Since transformers lack the inherent spatial understanding that CNNs have, positional encodings are added to the feature map to provide spatial information.
    \item \textbf{Transformer Encoder}: The encoded feature map is then passed through the transformer encoder, which processes the feature sequence and generates contextual representations.
    \item \textbf{Transformer Decoder}: The decoder receives learned queries and produces a fixed number of object predictions, each consisting of a bounding box and a class label.
\end{itemize}

This end-to-end structure eliminates the need for traditional components like anchor boxes and region proposal networks.

\textbf{Figure: DETR Pipeline Flowchart.} The following flowchart illustrates the process of DEtection TRansformer (DETR). It visualizes the steps from input image, feature extraction, transformer processing, to the final classification and bounding box prediction.

\begin{center}
\begin{tikzpicture}[
  box/.style={draw, minimum width=3cm, minimum height=1cm, align=center},
  arrow/.style={->, thick}
]

\node (input) at (0,0) [box] {Input Image};

\node (feature_extraction) at (0,-2.5) [box] {Feature Extraction};

\node (transformer) at (0,-5) [box] {Transformer Encoder};

\node (decoder) at (0,-7.5) [box] {Transformer Decoder};

\node (classifier) at (-3,-10) [box] {Class Prediction};
\node (bbox) at (3,-10) [box] {Bounding Box Prediction};

\node (class_output) at (-3,-12.5) [box] {Class Output};
\node (bbox_output) at (3,-12.5) [box] {Bounding Box Output};

\draw [arrow] (input) -- (feature_extraction);
\draw [arrow] (feature_extraction) -- (transformer);
\draw [arrow] (transformer) -- (decoder);
\draw [arrow] (decoder) -- (classifier);
\draw [arrow] (decoder) -- (bbox);
\draw [arrow] (classifier) -- (class_output);
\draw [arrow] (bbox) -- (bbox_output);

\end{tikzpicture}
\end{center}

\textit{In this figure, the flow of DETR is explained starting from the input image. The image is first processed to extract features, which are then passed through a Transformer encoder and decoder. Finally, the model outputs class predictions and bounding box coordinates for object localization.}

The following is the pseudocode for the DETR (DEtection TRansformer) model. DETR is a novel object detection model that leverages transformers for direct set prediction of objects in an image, eliminating the need for region proposals and non-maximum suppression.

\begin{lstlisting}[style=python]
# Pseudocode for DETR (DEtection TRansformer)

# Step 1: Input image
Input: Image

# Step 2: Extract features from the image using a CNN backbone (e.g., ResNet)
Features = CNN_Backbone(Image)

# Step 3: Flatten the features to prepare for the transformer
Flattened_features = Flatten(Features)

# Step 4: Add positional encodings to the flattened features
Positional_encoded_features = Add_Positional_Encoding(Flattened_features)

# Step 5: Pass the encoded features through the transformer encoder
Encoder_output = Transformer_Encoder(Positional_encoded_features)

# Step 6: Pass the encoder output through the transformer decoder
Decoder_output = Transformer_Decoder(Encoder_output)

# Step 7: Project the decoder output to class scores and bounding box coordinates
Class_scores, Bounding_boxes = Projection_Layer(Decoder_output)

# Step 8: Apply a matching algorithm to associate predictions with ground truth
Matches = Hungarian_Matching(Class_scores, Ground_Truth)

# Step 9: Output the final detected objects and their bounding boxes
Output: Final_predictions = (Class_scores[Matches], Bounding_boxes[Matches])
\end{lstlisting}

Explanation of the pseudocode:

\begin{itemize}
    \item Step 1: The input image is fed into the DETR model for object detection.
    \item Step 2: Features are extracted from the image using a CNN backbone, such as ResNet, which captures the spatial hierarchies in the image.
    \item Step 3: The extracted features are flattened to prepare them for processing by the transformer.
    \item Step 4: Positional encodings are added to the flattened features to retain spatial information, as transformers do not inherently understand the position of inputs.
    \item Step 5: The positional encoded features are passed through the transformer encoder, which learns to represent the relationships between different parts of the image.
    \item Step 6: The output of the encoder is passed to the transformer decoder, which generates object predictions based on the encoded features.
    \item Step 7: The decoder output is projected onto class scores and bounding box coordinates, predicting the presence of objects and their locations in the image.
    \item Step 8: A matching algorithm, such as the Hungarian algorithm, is used to associate the predicted outputs with the ground truth labels for training.
    \item Step 9: The final output consists of the detected objects and their bounding boxes, ready for evaluation or further processing.
\end{itemize}

\subsection{Improvements on DETR (Deformable DETR, Efficient DETR)}
Since the introduction of DETR, researchers have proposed several improvements to enhance its performance in terms of both speed and accuracy.

\subsubsection{Deformable DETR}
One challenge with the original DETR is its slow convergence. Deformable DETR \cite{zhu2020deformable} addresses this issue by replacing the standard transformer attention mechanism with deformable attention, which focuses only on relevant parts of the feature map. This speeds up training and improves accuracy, especially for small objects.

\subsubsection{Efficient DETR}
Efficient DETR \cite{yao2021efficient} is another improvement aimed at reducing the computational cost of DETR. By optimizing the transformer architecture and employing more efficient query mechanisms, Efficient DETR reduces the number of computations needed, making the model faster while maintaining accuracy.

Both Deformable DETR and Efficient DETR show that while DETR's transformer-based approach is highly effective, there is room for further optimization, particularly in terms of speed and computational efficiency.

%% file: 26_sam.tex
\section{Segment Anything Model (SAM)}

\subsection{Introduction to SAM (2023)}
Segment Anything Model (SAM) \cite{kirillov2023segment}is a general-purpose segmentation model introduced in 2023, designed to segment any object in an image. This model simplifies image segmentation, a crucial task in computer vision, which involves dividing an image into meaningful regions or objects. SAM has gained significant attention due to its versatility, as it is not limited to a specific set of objects or images. Instead, it can segment a wide range of objects in diverse image contexts, making it applicable across multiple domains like medical imaging \cite{ma2024segment,ren2024segment}, autonomous vehicles\cite{shan2023robustness,yan2024segment}, and interactive image editing \cite{kim2023evaluation}.

The impact of SAM lies in its ability to perform high-quality segmentation with minimal user input. Unlike traditional models that require extensive annotated datasets, SAM uses advanced neural network techniques, including pre-trained models, to generalize across unseen objects. SAM allows users to select an object by either providing a point or a bounding box, and it will segment the object automatically. This ability to "segment anything" revolutionizes image processing and opens new doors for real-time object segmentation in everyday applications.

\subsection{Architecture of SAM}
The architecture of SAM is designed to balance accuracy and speed. At its core, SAM uses a transformer-based \cite{vaswani2017attention} neural network architecture. Transformers have become a standard in many deep learning models because of their attention mechanisms, which allow the model to focus on specific parts of an image to improve segmentation precision.

The SAM model typically consists of three main components:

\begin{itemize}
    \item \textbf{Backbone Network}: The backbone extracts features from the input image. SAM uses a powerful pre-trained backbone, such as a version of the Vision Transformer (ViT) \cite{dosovitskiy2020image}, that has already been trained on large datasets like ImageNet \cite{deng2009imagenet}. This enables the model to recognize and process a wide range of objects.
    \item \textbf{Attention Mechanism}: The attention mechanism enables SAM to focus on different parts of the image and decide which areas belong to the object of interest. The attention map assigns weights to different regions in the image, helping SAM to detect complex object boundaries.
    \item \textbf{Segmentation Head}: After processing the image through the backbone and attention mechanism, the segmentation head produces the final segmentation mask. This mask is a pixel-wise representation of the object, identifying the exact pixels that make up the object in the image.
\end{itemize}

The architecture's strength lies in its ability to process high-resolution images while maintaining real-time segmentation performance. Additionally, SAM's generalization capability means it can segment objects without needing to be specifically trained on them, making it a state-of-the-art model in segmentation tasks.

\textbf{Figure: Segment Anything Model Pipeline Flowchart.} The following flowchart illustrates the process of the Segment Anything Model (SAM). It visualizes the steps from input image, segmentation proposals, feature extraction, to the final segmentation output.

\begin{center}
\begin{tikzpicture}[
  box/.style={draw, minimum width=3cm, minimum height=1cm, align=center},
  arrow/.style={->, thick}
]

\node (input) at (0,0) [box] {Input Image};

\node (proposal) at (0,-2.5) [box] {Segmentation Proposals};

\node (feature_extraction) at (0,-5) [box] {Feature Extraction};

\node (flatten) at (0,-7.5) [box] {Flattened Features};

\node (segmentation_output) at (0,-10) [box] {Segmentation Output};

\draw [arrow] (input) -- (proposal);
\draw [arrow] (proposal) -- (feature_extraction);
\draw [arrow] (feature_extraction) -- (flatten);
\draw [arrow] (flatten) -- (segmentation_output);

\end{tikzpicture}
\end{center}

\textit{In this figure, the flow of the Segment Anything Model is explained starting from the input image. Segmentation proposals are generated from the image, followed by feature extraction. The features are flattened to produce the final segmentation output, which localizes the objects in the image.}

The following is the pseudocode for the Segment Anything Model (SAM). SAM is designed for image segmentation tasks, providing a flexible framework that allows users to specify segmentation masks through various input methods.

\begin{lstlisting}[style=python]
# Pseudocode for Segment Anything Model (SAM)

# Step 1: Input image
Input: Image

# Step 2: Initialize SAM model
SAM_model = Initialize_SAM()

# Step 3: Obtain user input for segmentation
User_input = Get_User_Input()

# Step 4: Preprocess the input image
Preprocessed_image = Preprocess(Image)

# Step 5: Generate segmentation masks based on user input
Segmentation_masks = SAM_model.Generate_Segmentation(Preprocessed_image, User_input)

# Step 6: Post-process segmentation masks
Post_processed_masks = Postprocess(Segmentation_masks)

# Step 7: Output the segmented regions
Output: Post_processed_masks
\end{lstlisting}

Explanation of the pseudocode:

\begin{itemize}
    \item Step 1: The input image is loaded into the system for segmentation.
    \item Step 2: The Segment Anything Model (SAM) is initialized, preparing the model for inference.
    \item Step 3: User input is obtained, which can include points, boxes, or other forms of specifying areas of interest for segmentation.
    \item Step 4: The input image is preprocessed to ensure it meets the model's requirements (e.g., resizing, normalization).
    \item Step 5: The model generates segmentation masks based on the preprocessed image and the user input, identifying the regions of interest.
    \item Step 6: The segmentation masks may be post-processed to refine edges or combine overlapping segments for better accuracy.
    \item Step 7: The final output consists of the segmented regions represented by the post-processed masks.
\end{itemize}

The following is an implementation of the Segment Anything Model (SAM) using PyTorch for practical deployment scenarios.

\begin{lstlisting}[style=python]
import torch
from segment_anything import sam_model_registry
from torchvision.transforms import functional as F
from PIL import Image

# Load the Segment Anything Model
sam = sam_model_registry["vit_h"](checkpoint="sam_vit_h_4b8939.pth")
sam.eval()  # Set to evaluation mode for inference

# Define transformation function
def transform_image(image):
    # Convert image to tensor and normalize it
    image = F.to_tensor(image)
    return image

# Load an image and apply transformations
image_path = "sample_image.jpg"
image = Image.open(image_path)
image_tensor = transform_image(image)

# Perform inference
with torch.no_grad():  # No need to calculate gradients for inference
    masks = sam.predict(image=image_tensor)

# Print the prediction results
print("Predicted masks: ", masks)

# For training, you would set up a training loop similar to the following:
def train_segment_anything(sam, train_data_loader, optimizer, device):
    sam.train()  # Set to training mode
    for images, targets in train_data_loader:
        images = [img.to(device) for img in images]
        targets = [{k: v.to(device) for k, v in t.items()} for t in targets]
        
        # Forward pass
        loss_dict = sam(images, targets)
        losses = sum(loss for loss in loss_dict.values())
        
        # Backward pass
        optimizer.zero_grad()
        losses.backward()
        optimizer.step()

# Note: Make sure to configure the optimizer and data loader appropriately
\end{lstlisting}

This code demonstrates how to load a pre-trained Segment Anything Model (SAM), apply necessary transformations to input images, and perform segmentation by predicting masks. It also includes a simple training loop example, showing how to train the model using a custom dataset with an appropriate optimizer and data loader.

\subsection{Applications of SAM in Object Detection and Segmentation}
SAM's versatility makes it useful in both object detection and segmentation tasks. Object detection involves identifying the location and category of objects in an image, while segmentation goes a step further by labeling each pixel of the object.

Here are some key applications of SAM:

\begin{itemize}
    \item \textbf{Medical Imaging}: SAM can be used to segment tumors or organs in medical images like MRIs or CT scans, which helps doctors analyze medical data more efficiently \cite{ma2024segment,ren2024segment}.
    \item \textbf{Autonomous Driving}: In autonomous vehicles, SAM helps in detecting and segmenting pedestrians, vehicles, and obstacles in real-time to enhance safety and decision-making processes \cite{shan2023robustness,yan2024segment}.
    \item \textbf{Image Editing and Augmented Reality (AR)}: SAM can be integrated into image editing software to allow users to quickly select and edit objects \cite{kim2023evaluation}. In AR applications, SAM enables better object tracking and interaction.
    \item \textbf{Agriculture}: SAM can assist in detecting and segmenting crops, weeds, or pests in agricultural images, improving monitoring and management of large farming fields \cite{li2023enhancing}.
\end{itemize}

These applications showcase SAM's ability to be adapted to different domains, making it a crucial tool for various industries that rely on accurate and fast image segmentation.

\subsection{SAM2: Further Advancements in Segmentation and Detection (2024)}
In 2024, SAM2 \cite{ravi2024sam} was introduced as an advancement over the original SAM model. SAM2 builds on the success of SAM by improving both segmentation accuracy and efficiency. Some of the key improvements include:

\begin{itemize}
    \item \textbf{Improved Feature Extraction}: SAM2 employs a more refined feature extraction process, allowing it to detect even smaller objects and segment them with greater precision.
    \item \textbf{Faster Inference}: SAM2 optimizes its inference speed, making it more suitable for real-time applications such as video object segmentation and autonomous driving.
    \item \textbf{Multi-object Segmentation}: SAM2 can now handle multiple objects in a single image more effectively, allowing it to separate overlapping objects and produce individual segmentation masks for each.
    \item \textbf{Cross-domain Generalization}: SAM2 is better at generalizing across different domains (e.g., medical imaging vs. natural scenes) without the need for fine-tuning on each domain, thanks to the improvements in its attention mechanism and training data diversity.
\end{itemize}

These advancements make SAM2 a superior choice for complex segmentation tasks, and its improvements have expanded its potential applications even further.

\textbf{Figure: Segment Anything Model 2 (SAM2) Pipeline Flowchart.} The following flowchart illustrates the process of the Segment Anything Model 2 (SAM2). It visualizes the steps from input image to segmentation output.

\begin{center}
\begin{tikzpicture}[
  box/.style={draw, minimum width=3cm, minimum height=1cm, align=center},
  arrow/.style={->, thick}
]

\node (input) at (0,0) [box] {Input Image};

\node (preprocessing) at (0,-2.5) [box] {Preprocessing};

\node (feature_extraction) at (0,-5) [box] {Feature Extraction};

\node (segmentation) at (0,-7.5) [box] {Segmentation};

\node (postprocessing) at (0,-10) [box] {Postprocessing};

\node (output) at (0,-12.5) [box] {Segmentation Output};

\draw [arrow] (input) -- (preprocessing);
\draw [arrow] (preprocessing) -- (feature_extraction);
\draw [arrow] (feature_extraction) -- (segmentation);
\draw [arrow] (segmentation) -- (postprocessing);
\draw [arrow] (postprocessing) -- (output);

\end{tikzpicture}
\end{center}

\textit{In this figure, the flow of the Segment Anything Model 2 (SAM2) is explained starting from the input image. The image undergoes preprocessing, followed by feature extraction. The extracted features are then used for segmentation, which is followed by postprocessing to obtain the final segmentation output.}

The following is the pseudocode for the Segment Anything Model 2 (SAM2). SAM2 is designed for image segmentation tasks, where the model is capable of segmenting objects within an image without needing extensive training on specific datasets.

\begin{lstlisting}[style=python]
# Pseudocode for Segment Anything Model 2 (SAM2)

# Step 1: Input image
Input: Image

# Step 2: Preprocess the input image
Preprocessed_Image = Preprocess(Image)

# Step 3: Generate initial segmentation masks using a segmentation model
Initial_Masks = Segmentation_Model(Preprocessed_Image)

# Step 4: For each mask in Initial_Masks
For each mask in Initial_Masks:
    # Step 5: Refine the mask using a refinement network
    Refined_Mask = Mask_Refinement_Network(mask)

    # Step 6: Assign a label to the refined mask
    Label = Classifier(Refined_Mask)

    # Step 7: Store the refined mask and its corresponding label
    Store(Refined_Mask, Label)

# Step 8: Post-processing: Combine masks to create final segmentation
Final_Segmentation = Combine_Masks(Refined_Masks)

# Step 9: Output the final segmented objects and their labels
Output: Final_Segmentation
\end{lstlisting}

Explanation of the pseudocode:

\begin{itemize}
    \item Step 1: The input image is provided for segmentation tasks.
    \item Step 2: The input image undergoes preprocessing to enhance its quality for segmentation.
    \item Step 3: An initial set of segmentation masks is generated using a pre-trained segmentation model.
    \item Step 4: For each mask generated, the refinement process begins.
    \item Step 5: Each initial mask is refined using a mask refinement network to improve its accuracy.
    \item Step 6: A classifier assigns a label to each refined mask based on the features extracted from it.
    \item Step 7: The refined masks and their corresponding labels are stored for further processing.
    \item Step 8: Post-processing is performed to combine all refined masks into a final segmentation result.
    \item Step 9: The final output consists of the segmented objects along with their labels.
\end{itemize}

\subsection{Comparison of SAM with YOLO and DETR}
SAM, YOLO (You Only Look Once), and DETR (Detection Transformer) are popular models for object detection and segmentation, but they differ in their approaches and use cases.

\begin{itemize}
    \item \textbf{SAM}: Focuses on segmentation and excels in tasks where precise pixel-level segmentation is needed. SAM's transformer-based architecture allows it to handle diverse objects without needing domain-specific training. It is slower than YOLO in terms of real-time object detection but provides more accurate segmentation results.
    \item \textbf{YOLO}: YOLO is designed for fast real-time object detection. It processes an entire image in a single pass and predicts bounding boxes for multiple objects in real-time. However, YOLO is not focused on segmentation, and its pixel-level accuracy is lower compared to SAM.
    \item \textbf{DETR}: Like SAM, DETR uses a transformer-based approach. DETR excels at object detection and segmentation, but it may require more computational resources than YOLO. SAM and DETR share similarities in terms of transformer usage, but SAM is more focused on generalized segmentation tasks, while DETR is primarily designed for object detection with some segmentation capabilities.
\end{itemize}

In summary, SAM is best suited for detailed segmentation tasks, YOLO is ideal for fast real-time detection, and DETR strikes a balance between detection and segmentation with a transformer-based design.

%% file: 27_other.tex
\section{Other Notable Object Detection Models}

Object detection has evolved significantly over the years, leading to the development of various innovative models that tackle the detection problem from different perspectives. This section introduces three notable object detection models: EfficientDet \cite{tan2020efficientdet}, CornerNet \cite{law2018cornernet}, and CenterNet \cite{duan2019centernet}. We will explain their architectures, key features, and benefits, ensuring that even beginners can follow along.

\subsection{EfficientDet: Scalable and Efficient Object Detection (2020)}
EfficientDet is a state-of-the-art object detection model introduced by Mingxing Tan, Ruoming Pang, and Quoc V. Le in 2020 \cite{tan2020efficientdet}. It focuses on achieving a balance between accuracy and efficiency by introducing a scalable architecture. EfficientDet is based on the EfficientNet backbone, which uses a compound scaling method to balance model depth, width, and resolution. This method allows the model to scale up efficiently with fewer parameters compared to traditional models.

\subsubsection{Key Features of EfficientDet:}
\begin{itemize}
    \item \textbf{EfficientNet Backbone}: EfficientNet is used as the feature extractor. It is a family of convolutional neural networks (CNNs) designed to be both accurate and efficient.
    \item \textbf{BiFPN (Bidirectional Feature Pyramid Network)}: EfficientDet introduces a novel feature fusion network called BiFPN, which allows information to flow in both directions (top-down and bottom-up), making feature extraction more effective.
    \item \textbf{Compound Scaling}: Instead of scaling individual components (e.g., depth or resolution), EfficientDet uses a compound scaling method that scales all three dimensions of the model simultaneously, achieving better performance.
    \item \textbf{Scalable Model Sizes}: EfficientDet offers different model sizes from D0 to D7, where each successive model is larger and more powerful than the previous one. This scalability makes EfficientDet suitable for both resource-constrained environments and high-performance systems.
\end{itemize}

\textbf{Figure: EfficientDet Pipeline Flowchart.} The following flowchart illustrates the process of EfficientDet, a state-of-the-art object detection model. It visualizes the steps from input image, feature extraction, feature fusion, to the final classification and bounding box regression.

\begin{center}
\begin{tikzpicture}[
  box/.style={draw, minimum width=3cm, minimum height=1cm, align=center},
  arrow/.style={->, thick}
]

\node (input) at (0,0) [box] {Input Image};

\node (backbone) at (0,-2.5) [box] {Backbone Feature Extraction};

\node (fusion) at (0,-5) [box] {Feature Fusion};

\node (classifier) at (-3,-7.5) [box] {Class Prediction Head};
\node (regressor) at (3,-7.5) [box] {Bounding Box Head};

\node (class_output) at (-3,-10) [box] {Class Predictions};
\node (bbox_output) at (3,-10) [box] {Bounding Box Coordinates};

\draw [arrow] (input) -- (backbone);
\draw [arrow] (backbone) -- (fusion);
\draw [arrow] (fusion) -- (classifier);
\draw [arrow] (fusion) -- (regressor);
\draw [arrow] (classifier) -- (class_output);
\draw [arrow] (regressor) -- (bbox_output);

\end{tikzpicture}
\end{center}

\textit{In this figure, the flow of EfficientDet is explained starting from the input image. The image is processed through a backbone for feature extraction, followed by feature fusion. Finally, two heads are used: one for class prediction and another for bounding box regression.}

The following is the pseudocode for the EfficientDet model, which is an efficient and scalable object detection architecture that balances accuracy and computational efficiency. EfficientDet employs a compound scaling method to adjust the depth, width, and resolution of the model.

\begin{lstlisting}[style=python]
# Pseudocode for EfficientDet

# Step 1: Input image
Input: Image

# Step 2: Define the scaling factors for model depth, width, and resolution
Depth_scale = d  # depth scaling factor
Width_scale = w  # width scaling factor
Resolution_scale = r  # resolution scaling factor

# Step 3: Create the EfficientNet backbone with specified scaling factors
Backbone = Create_EfficientNet(Depth_scale, Width_scale, Resolution_scale)

# Step 4: Build the BiFPN (Bidirectional Feature Pyramid Network) for feature fusion
BiFPN = Create_BiFPN(Backbone)

# Step 5: For each feature level in BiFPN
For each feature_level in BiFPN:
    # Step 6: Apply classification and bounding box regression heads
    Class_scores = Classification_Head(feature_level)
    Bounding_box_deltas = Regression_Head(feature_level)

# Step 7: Post-processing: Apply non-maximum suppression (NMS) to filter detections
Final_predictions = NonMaximumSuppression(Class_scores, Bounding_box_deltas)

# Step 8: Output the detected objects and their bounding boxes
Output: Final_predictions
\end{lstlisting}

Explanation of the pseudocode:

\begin{itemize}
    \item Step 1: The input image is provided for object detection.
    \item Step 2: Define scaling factors for the model's depth, width, and resolution, which allows EfficientDet to adjust its architecture for better performance.
    \item Step 3: Create the EfficientNet backbone using the specified scaling factors to extract features from the input image.
    \item Step 4: Construct the BiFPN, which efficiently fuses features from different levels of the backbone, enabling better multi-scale detection.
    \item Step 5: For each feature level in the BiFPN, the model applies heads for classification and bounding box regression.
    \item Step 6: The classification head outputs class scores for detected objects, while the regression head outputs bounding box adjustments.
    \item Step 7: Non-maximum suppression (NMS) is applied to remove duplicate or overlapping bounding boxes from the detections.
    \item Step 8: The final output consists of the detected objects and their refined bounding boxes.
\end{itemize}

\subsubsection{Example of EfficientDet in Python:}
Here's an example of how to use EfficientDet in Python with TensorFlow and Keras to detect objects in an image:

\begin{lstlisting}[style=python]
import tensorflow as tf
from tensorflow import keras
from efficientdet import EfficientDetModel

# Load pre-trained EfficientDet model
model = EfficientDetModel(weights='efficientdet-d0')

# Load an example image
image = tf.io.read_file('example.jpg')
image = tf.image.decode_jpeg(image, channels=3)
image = tf.image.resize(image, [512, 512])
image = tf.expand_dims(image, axis=0)  # Add batch dimension

# Perform object detection
detections = model.predict(image)

# Print detection results
for detection in detections:
    print(f"Object: {detection['label']}, Confidence: {detection['confidence']:.2f}")
\end{lstlisting}

This example demonstrates how easy it is to load an EfficientDet model and use it for object detection in Python.

\subsection{CornerNet: Object Detection as Keypoint Detection (2019)}
CornerNet \cite{law2018cornernet} is an innovative object detection model that treats object detection as a keypoint detection problem. Unlike traditional object detection methods that rely on anchor boxes, CornerNet proposes detecting objects by identifying their corners (i.e., top-left and bottom-right corners). Once the corners are identified, they are grouped together to form bounding boxes for objects.

\subsubsection{Key Features of CornerNet:}
\begin{itemize}
    \item \textbf{Keypoint Detection}: Instead of detecting objects directly, CornerNet detects the keypoints (i.e., corners) of objects. This approach eliminates the need for anchor boxes, which are common in many object detection algorithms.
    \item \textbf{Hourglass Network}: CornerNet uses an hourglass network \cite{newell2016stacked} to detect the corners. This network structure captures multi-scale contextual information by performing downsampling and upsampling operations, which helps in accurate keypoint detection.
    \item \textbf{Grouping Corners}: After detecting the top-left and bottom-right corners of an object, CornerNet groups the corners based on their spatial proximity and forms bounding boxes.
\end{itemize}

\subsubsection{Example of CornerNet Detection Process:}

In the following diagram, we illustrate how CornerNet detects the corners of an object:

\begin{tikzpicture}
    \draw[thick] (0,0) rectangle (4,2);
    
    \filldraw[red] (0,2) circle (2pt) node[anchor=east] {Top-left corner};
    
    \filldraw[blue] (4,0) circle (2pt) node[anchor=west] {Bottom-right corner};
    
    \node at (2,1) {Object};
\end{tikzpicture}

This simple diagram shows how CornerNet identifies the top-left and bottom-right corners, which are then used to form the bounding box.

\textbf{Figure: CornerNet Pipeline Flowchart.} The following flowchart illustrates the process of CornerNet, a keypoint-based object detection method. It visualizes the steps from input image, corner detection, feature extraction, to the final classification and bounding box prediction.

\begin{center}
\begin{tikzpicture}[
  box/.style={draw, minimum width=3cm, minimum height=1cm, align=center},
  arrow/.style={->, thick}
]

\node (input) at (0,0) [box] {Input Image};

\node (corner) at (0,-2.5) [box] {Corner Detection};

\node (feature) at (0,-5) [box] {Feature Extraction};

\node (keypoint) at (0,-7.5) [box] {Keypoint Representation};

\node (classifier) at (-3,-10) [box] {Class Predictor};
\node (regressor) at (3,-10) [box] {Bounding Box Predictor};

\node (class_output) at (-3,-12.5) [box] {Class Prediction};
\node (bbox_output) at (3,-12.5) [box] {Bounding Box Coordinates};

\draw [arrow] (input) -- (corner);
\draw [arrow] (corner) -- (feature);
\draw [arrow] (feature) -- (keypoint);
\draw [arrow] (keypoint) -- (classifier);
\draw [arrow] (keypoint) -- (regressor);
\draw [arrow] (classifier) -- (class_output);
\draw [arrow] (regressor) -- (bbox_output);

\end{tikzpicture}
\end{center}

\textit{In this figure, the flow of CornerNet is explained starting from the input image. Corner detection identifies keypoints, which are then processed through feature extraction. The resulting keypoint representation is used for class prediction and bounding box prediction, allowing for accurate object detection.}

The following is the pseudocode for the CornerNet model, which is used for object detection. CornerNet detects objects as pairs of keypoints (corners) and uses them to create bounding boxes around the detected objects. The main steps include generating corner points, associating them, and refining the bounding box predictions.

\begin{lstlisting}[style=python]
# Pseudocode for CornerNet

# Step 1: Input image
Input: Image

# Step 2: Backbone network for feature extraction
Features = Backbone_Network(Image)

# Step 3: Predict corner points (top-left and bottom-right) for objects
Top_left_corners = Predict_TopLeft(Features)
Bottom_right_corners = Predict_BottomRight(Features)

# Step 4: Predict heatmaps for object existence
Heatmap = Predict_Heatmap(Features)

# Step 5: Generate bounding boxes from corner points
Bounding_boxes = Generate_BoundingBoxes(Top_left_corners, Bottom_right_corners)

# Step 6: Post-processing: Apply non-maximum suppression (NMS) on the bounding boxes
Final_boxes = NonMaximumSuppression(Bounding_boxes, Heatmap)

# Step 7: Output the detected objects and their bounding boxes
Output: Final_boxes
\end{lstlisting}

Explanation of the pseudocode:

\begin{itemize}
    \item Step 1: The input image is provided to the model for object detection.
    \item Step 2: A backbone network (like ResNet or Hourglass) is used to extract feature maps from the input image.
    \item Step 3: Two separate predictions are made for the corner points: the top-left and bottom-right corners of the bounding boxes.
    \item Step 4: A heatmap is predicted to indicate the presence of objects in the image.
    \item Step 5: Bounding boxes are generated using the predicted corner points.
    \item Step 6: Non-maximum suppression (NMS) is applied to the generated bounding boxes based on the heatmap scores to eliminate duplicates and refine predictions.
    \item Step 7: The final output consists of the detected objects along with their bounding boxes.
\end{itemize}

\subsubsection{CornerNet Python Example:}
Although CornerNet is less widely implemented in popular libraries compared to EfficientDet, here is a conceptual example of how CornerNet might work in Python:

\begin{lstlisting}[style=python]
# Assuming a hypothetical implementation of CornerNet
from cornernet import CornerNetModel

# Load pre-trained CornerNet model
model = CornerNetModel(weights='cornernet')

# Load image
image = load_image('example.jpg')

# Detect keypoints
keypoints = model.detect_keypoints(image)

# Group keypoints into bounding boxes
objects = group_keypoints_to_objects(keypoints)

# Print detection results
for obj in objects:
    print(f"Detected object at bounding box: {obj['bbox']}")
\end{lstlisting}

In this example, we show how CornerNet detects keypoints and groups them to form bounding boxes, which are then used to identify objects.

\subsection{CenterNet: Objects as Points (2019)}
CenterNet \cite{duan2019centernet} is a model that further simplifies object detection by detecting the center points of objects. Instead of focusing on bounding box corners or anchor boxes, CenterNet predicts the center of an object directly, and from this center point, it estimates the size and shape of the bounding box. This approach eliminates the need for anchor boxes or corner detection, making the process more streamlined.

\subsubsection{Key Features of CenterNet:}
\begin{itemize}
    \item \textbf{Center Point Detection}: CenterNet predicts the center point of each object in an image. This central location is then used to infer the size of the bounding box.
    \item \textbf{Heatmaps}: The model generates heatmaps where the peak of each heatmap corresponds to the center of an object.
    \item \textbf{Keypoint Regression}: CenterNet uses regression techniques to predict the size and offset of the bounding box from the detected center point.
\end{itemize}

\subsubsection{Example of CenterNet Detection:}

In the following diagram, we show how CenterNet detects the center of an object and uses it to predict the bounding box:

\begin{tikzpicture}
    \draw[thick] (0,0) rectangle (5,2);
    
    \filldraw[codegreen] (2.5,1) circle (3pt) node[anchor=west] {Center point};

    \draw[dashed] (0,0) rectangle (5,2);
\end{tikzpicture}

This diagram shows that CenterNet focuses on detecting the center point and then uses that information to predict the bounding box around the object.

\textbf{Figure: CenterNet Pipeline Flowchart.} The following flowchart illustrates the process of CenterNet, a key approach in object detection. It visualizes the steps from input image, keypoint estimation, heatmap generation, to the final object localization and classification.

\begin{center}
\begin{tikzpicture}[
  box/.style={draw, minimum width=3cm, minimum height=1cm, align=center},
  arrow/.style={->, thick}
]

\node (input) at (0,0) [box] {Input Image};

\node (keypoint) at (0,-2.5) [box] {Keypoint Estimation};

\node (heatmap) at (0,-5) [box] {Heatmap Generation};

\node (centerpoint) at (0,-7.5) [box] {Center Point Detection};

\node (bbox_class) at (0,-10) [box] {Bounding Box + Classification};

\node (output) at (0,-12.5) [box] {Final Output};

\draw [arrow] (input) -- (keypoint);
\draw [arrow] (keypoint) -- (heatmap);
\draw [arrow] (heatmap) -- (centerpoint);
\draw [arrow] (centerpoint) -- (bbox_class);
\draw [arrow] (bbox_class) -- (output);

\end{tikzpicture}
\end{center}

\textit{In this figure, the flow of CenterNet is explained starting from the input image. The keypoints are estimated from the image, which are then used to generate heatmaps. Center points are detected from these heatmaps, followed by bounding box generation and classification to produce the final output.}

The following is the pseudocode for the CenterNet model. CenterNet is an object detection framework that uses a keypoint-based approach to identify the center points of objects, as well as their sizes, for detection. This model simplifies the detection pipeline by representing objects as points and enables efficient object detection.

\begin{lstlisting}[style=python]
# Pseudocode for CenterNet

# Step 1: Input image
Input: Image

# Step 2: Pass the image through a backbone network (e.g., ResNet)
Features = BackboneNetwork(Image)

# Step 3: Generate heatmap for center points of objects
Center_heatmap = Generate_Center_Heatmap(Features)

# Step 4: Generate size and offset maps for bounding boxes
Size_map = Generate_Size_Map(Features)
Offset_map = Generate_Offset_Map(Features)

# Step 5: For each pixel in the heatmap
For each pixel (x, y) in Center_heatmap:
    # Step 6: If heatmap value exceeds a threshold
    If Center_heatmap[x, y] > Threshold:
        # Step 7: Calculate the center point coordinates
        Center_x = x
        Center_y = y
        
        # Step 8: Get the size of the object from Size_map
        Object_size = Size_map[x, y]
        
        # Step 9: Get the offsets from Offset_map
        Offset_x = Offset_map[x, y][0]
        Offset_y = Offset_map[x, y][1]

        # Step 10: Calculate the final bounding box coordinates
        Bounding_box = Calculate_Bounding_Box(Center_x, Center_y, Object_size, Offset_x, Offset_y)

# Step 11: Post-processing: Apply non-maximum suppression (NMS)
Final_predictions = NonMaximumSuppression(Bounding_boxes)

# Step 12: Output the detected objects and their bounding boxes
Output: Final_predictions
\end{lstlisting}

Explanation of the pseudocode:

\begin{itemize}
    \item Step 1: The input image is provided for object detection.
    \item Step 2: The image is processed through a backbone network (like ResNet) to extract feature representations.
    \item Step 3: A center heatmap is generated, where each pixel value indicates the likelihood of the center of an object being present at that location.
    \item Step 4: Size and offset maps are generated to determine the size of objects and their offsets from the center point.
    \item Step 5: For each pixel in the center heatmap, the model checks if the value exceeds a predefined threshold to determine potential object centers.
    \item Step 6: If a center point is detected, the coordinates are calculated.
    \item Step 8: The size of the object is retrieved from the size map corresponding to the center point.
    \item Step 9: Offsets from the center are obtained from the offset map to adjust the bounding box.
    \item Step 10: The final bounding box coordinates are calculated based on the center, size, and offsets.
    \item Step 11: Non-maximum suppression (NMS) is applied to filter out duplicate or overlapping detections.
    \item Step 12: The final output includes detected objects and their respective bounding boxes.
\end{itemize}

\subsubsection{CenterNet Python Example:}
Here's how you can use CenterNet in Python with a pre-trained model:

\begin{lstlisting}[style=python]
import tensorflow as tf
from centernet import CenterNetModel

# Load pre-trained CenterNet model
model = CenterNetModel(weights='centernet')

# Load and preprocess image
image = tf.io.read_file('example.jpg')
image = tf.image.decode_jpeg(image, channels=3)
image = tf.image.resize(image, [512, 512])
image = tf.expand_dims(image, axis=0)

# Perform object detection
detections = model.predict(image)

# Print detected objects and bounding boxes
for detection in detections:
    print(f"Object detected at: {detection['center']}, Bounding box: {detection['bbox']}")
\end{lstlisting}

This example shows how CenterNet detects the center points and uses them to estimate the bounding boxes for objects.

\subsection{Conclusion}
In this section, we have explored three cutting-edge object detection models: EfficientDet, CornerNet, and CenterNet. Each model brings a unique perspective to the object detection problem, from scalability and efficiency in EfficientDet to keypoint and center point detection in CornerNet and CenterNet, respectively. These models illustrate how object detection techniques continue to evolve and improve in both accuracy and efficiency.

%% file: 28_eval.tex
\section{Evaluation Metrics for Object Detection}

In object detection, evaluating the performance of a model is crucial. The evaluation process uses specific metrics to measure how well the model can detect objects in images. Below, we introduce the most commonly used evaluation metrics in object detection, including detailed explanations and examples.

\subsection{Mean Average Precision (mAP)}

Mean Average Precision (mAP) is one of the most popular metrics used to evaluate object detection models. It provides a single score to summarize the model's performance across different classes.

\subsubsection{How is mAP Calculated?}
The calculation of mAP is based on the average precision (AP) of each class, which is determined from the precision-recall curve. Here's a step-by-step explanation:

\begin{enumerate}
    \item For each object class, the precision and recall are calculated at different confidence thresholds.
    \item A precision-recall curve is drawn by plotting precision values on the y-axis and recall values on the x-axis.
    \item The average precision (AP) is the area under the precision-recall curve for each class.
    \item The mean average precision (mAP) is the average of the AP values across all object classes.
\end{enumerate}

\subsubsection{Interpreting mAP}
A higher mAP score indicates better performance. For instance, a model with an mAP of 0.80 is better than one with an mAP of 0.50, meaning it is more accurate in detecting objects.

\subsection{Intersection over Union (IoU)}

Intersection over Union (IoU) measures the overlap between the predicted bounding box and the ground truth (actual) bounding box. It's a critical part of object detection evaluation, especially when deciding if a predicted box correctly identifies an object.

\subsubsection{IoU Formula}
IoU is calculated using the formula:

\[
\text{IoU} = \frac{\text{Area of Overlap}}{\text{Area of Union}}
\]

Where:
\begin{itemize}
    \item \textbf{Area of Overlap} refers to the area where the predicted bounding box and the ground truth bounding box overlap.
    \item \textbf{Area of Union} is the total area covered by both the predicted and ground truth boxes combined.
\end{itemize}

\subsubsection{Example of IoU}
Suppose a ground truth bounding box has an area of 50 square pixels, and the predicted bounding box has an area of 60 square pixels. If the area of overlap between the two is 30 square pixels, the IoU would be calculated as:

\[
\text{IoU} = \frac{30}{50 + 60 - 30} = \frac{30}{80} = 0.375
\]

\subsubsection{Threshold for IoU}
In most object detection tasks, a predicted box is considered a correct detection if its IoU with the ground truth box exceeds a certain threshold, typically 0.50 or 0.75.

\subsection{Precision, Recall, and F1-Score}

\subsubsection{Precision}
Precision measures how many of the predicted bounding boxes are correct \cite{salton1983introduction}. It is defined as:

\[
\text{Precision} = \frac{\text{True Positives}}{\text{True Positives} + \text{False Positives}}
\]

For example, if a model predicts 10 bounding boxes, and 8 of them are correct, the precision would be:

\[
\text{Precision} = \frac{8}{8 + 2} = 0.80
\]

\subsubsection{Recall}
Recall measures how many of the actual objects were correctly predicted by the model \cite{salton1983introduction}. It is defined as:

\[
\text{Recall} = \frac{\text{True Positives}}{\text{True Positives} + \text{False Negatives}}
\]

For example, if there are 10 objects in an image, and the model correctly detects 7, the recall would be:

\[
\text{Recall} = \frac{7}{7 + 3} = 0.70
\]

\subsubsection{F1-Score}
The F1-score is the harmonic mean of precision and recall, giving a balanced measure when you want to consider both precision and recall \cite{van1979information}. It is defined as:

\[
\text{F1-Score} = 2 \times \frac{\text{Precision} \times \text{Recall}}{\text{Precision} + \text{Recall}}
\]

For example, if the precision is 0.80 and recall is 0.70, the F1-score would be:

\[
\text{F1-Score} = 2 \times \frac{0.80 \times 0.70}{0.80 + 0.70} = 0.746
\]

\section{Datasets for Object Detection}

Object detection models are trained and evaluated on large annotated datasets. Here, we introduce some of the most important datasets in the field.

\subsection{COCO Dataset}

The \textbf{COCO (Common Objects in Context)} dataset \cite{lin2014microsoft} 
is one of the most widely used datasets for object detection, segmentation, and keypoint detection. It contains over 200,000 labeled images and annotations for 80 object categories.

\subsubsection{Structure of COCO Dataset}
The COCO dataset includes:
\begin{itemize}
    \item \textbf{Images}: Natural images with objects in various contexts.
    \item \textbf{Bounding Box Annotations}: The dataset provides detailed bounding boxes for each object.
    \item \textbf{Segmentation Masks}: Along with bounding boxes, the COCO dataset includes precise object outlines (segmentation).
\end{itemize}

\subsubsection{Example of Using COCO Dataset in Python}

Here is an example of how to load and visualize data from the COCO dataset using the Python API:

\begin{lstlisting}[style=python]
from pycocotools.coco import COCO
import matplotlib.pyplot as plt
import numpy as np
import skimage.io as io

# Load COCO dataset
coco = COCO('annotations/instances_val2017.json')

# Get image ids and display a random image
image_ids = coco.getImgIds()
image_data = coco.loadImgs(image_ids[np.random.randint(0, len(image_ids))])[0]

# Load and display image
image = io.imread(image_data['coco_url'])
plt.imshow(image)
plt.axis('off')
plt.show()
\end{lstlisting}

\subsection{PASCAL VOC Dataset}

The \textbf{PASCAL VOC} dataset \cite{everingham2010pascal} was one of the earliest benchmark datasets for object detection and segmentation tasks. It includes 20 object categories and has been widely used to evaluate object detection models.

\subsubsection{Structure of PASCAL VOC Dataset}
The dataset includes:
\begin{itemize}
    \item \textbf{Images}: A total of around 11,000 images.
    \item \textbf{Annotations}: Bounding box annotations for 20 object categories.
\end{itemize}

\subsubsection{Using PASCAL VOC in Python}
Here's an example of how to load and visualize data from the PASCAL VOC dataset:

\begin{lstlisting}[style=python]
import matplotlib.pyplot as plt
import xml.etree.ElementTree as ET

# Load and parse VOC annotation XML file
tree = ET.parse('Annotations/000005.xml')
root = tree.getroot()

# Display image with bounding box annotations
plt.imshow(plt.imread('JPEGImages/000005.jpg'))
for obj in root.findall('object'):
    bbox = obj.find('bndbox')
    xmin = int(bbox.find('xmin').text)
    ymin = int(bbox.find('ymin').text)
    xmax = int(bbox.find('xmax').text)
    ymax = int(bbox.find('ymax').text)
    plt.gca().add_patch(plt.Rectangle((xmin, ymin), xmax - xmin, ymax - ymin, 
                                      linewidth=1, edgecolor='r', facecolor='none'))
plt.show()
\end{lstlisting}

\subsection{Open Images Dataset}

The \textbf{Open Images Dataset} \cite{kuznetsova2020open} is a large-scale dataset that contains millions of images and annotations for object detection, segmentation, and visual relationships.

\subsubsection{Contributions of Open Images Dataset}
The dataset contributes significantly to the field by providing:
\begin{itemize}
    \item \textbf{Complex Annotations}: Along with bounding boxes, Open Images provides annotations for object segmentation and visual relationships.
    \item \textbf{Large-Scale Data}: The dataset contains approximately 9 million images and over 15 million bounding boxes.
\end{itemize}

%% file: 29_prac.tex
\section{Object Detection in Practice}

\subsection{Training an Object Detection Model}
Object detection is the task of identifying objects within an image and classifying them into different categories. Training an object detection model from scratch involves several steps, and it requires a dataset that includes images along with bounding box annotations for the objects present in those images.

\textbf{Steps to train an object detection model:}
\begin{enumerate}
    \item \textbf{Data Preparation:} You need a dataset where each image is annotated with bounding boxes and class labels for the objects. Popular datasets for object detection include COCO, Pascal VOC, and Open Images. The annotations are often stored in formats like XML, JSON, or CSV.
    
    \item \textbf{Model Architecture:} There are many popular architectures for object detection such as Faster R-CNN, YOLO (You Only Look Once), and SSD (Single Shot MultiBox Detector). The architecture determines how the model identifies and classifies objects in an image.
    
    \item \textbf{Loss Function:} Object detection typically uses a combination of classification loss (for classifying objects) and regression loss (for predicting the coordinates of bounding boxes). For example, the loss function for YOLO consists of localization loss, confidence loss, and classification loss.
    
    \item \textbf{Training the Model:} During training, the model learns the patterns in the images and the corresponding bounding boxes. This involves multiple epochs and the use of an optimizer such as Adam or SGD (Stochastic Gradient Descent).

\begin{lstlisting}[style=python]
# Example: Training a basic object detection model using PyTorch
import torch
import torchvision
from torchvision.models.detection import fasterrcnn_resnet50_fpn

# Load a pretrained model
model = fasterrcnn_resnet50_fpn(pretrained=True)
model.train()

# Set up an optimizer
optimizer = torch.optim.SGD(model.parameters(), lr=0.005)

# Assuming you have your dataset ready in DataLoader format
for images, targets in dataloader:
    loss_dict = model(images, targets)
    losses = sum(loss for loss in loss_dict.values())
    optimizer.zero_grad()
    losses.backward()
    optimizer.step()
\end{lstlisting}

\end{enumerate}

\subsection{Fine-tuning Pretrained Models}
Instead of training a model from scratch, which requires a large dataset and significant computational resources, you can fine-tune a pretrained model on your specific dataset. Fine-tuning leverages the weights learned on a large dataset and adapts them to your task.

\textbf{Steps for fine-tuning a model:}
\begin{enumerate}
    \item \textbf{Choose a Pretrained Model:} Use models like Faster R-CNN or YOLO that have been trained on a large dataset such as COCO.
    
    \item \textbf{Modify the Final Layer:} Since the pretrained model might have been trained for a different set of classes, the last layer (classifier) must be adapted to match the number of classes in your dataset.
    
    \item \textbf{Fine-tune with Your Data:} Load your dataset and train the model with a lower learning rate to adjust the weights slowly without disrupting the pretrained weights too much.

\begin{lstlisting}[style=python]
# Example: Fine-tuning a Faster R-CNN model on a custom dataset
num_classes = 5  # Suppose we have 5 object classes in our dataset
# Load a pretrained model and modify the final layer
model = fasterrcnn_resnet50_fpn(pretrained=True)
in_features = model.roi_heads.box_predictor.cls_score.in_features
model.roi_heads.box_predictor = torchvision.models.detection.faster_rcnn.FastRCNNPredictor(in_features, num_classes)

# Train on custom dataset
for images, targets in custom_dataloader:
    loss_dict = model(images, targets)
    losses = sum(loss for loss in loss_dict.values())
    optimizer.zero_grad()
    losses.backward()
    optimizer.step()
\end{lstlisting}

\end{enumerate}

\subsection{Real-time Object Detection Applications}
Real-time object detection is critical in many applications where speed is important, such as video analysis, autonomous driving, and robotics. Models such as YOLO and SSD are commonly used for real-time detection due to their fast inference times.

\textbf{Applications of real-time object detection:}
\begin{enumerate}
    \item \textbf{Video Analysis:} Detecting and tracking objects across frames in a video for surveillance or sports analysis.
    
    \item \textbf{Autonomous Driving:} Identifying pedestrians, vehicles, and obstacles in real-time to make driving decisions.
    
    \item \textbf{Robotics:} Enabling robots to interact with objects in their environment through real-time perception.
\end{enumerate}

\begin{lstlisting}[style=python]
# Example: Real-time object detection using a webcam feed
import cv2
import torch
import torchvision.transforms as T

# Load the pretrained YOLO or Faster R-CNN model
model = fasterrcnn_resnet50_fpn(pretrained=True)
model.eval()

# Initialize the webcam
cap = cv2.VideoCapture(0)

# Define a transformation for input images
transform = T.Compose([T.ToTensor()])

while cap.isOpened():
    ret, frame = cap.read()
    if not ret:
        break
    
    # Convert frame to tensor and make predictions
    img = transform(frame)
    with torch.no_grad():
        predictions = model([img])[0]

    # Visualize the results (e.g., draw bounding boxes on the frame)
    # Code for drawing boxes goes here...

    # Show the frame with detected objects
    cv2.imshow('Object Detection', frame)

    if cv2.waitKey(1) & 0xFF == ord('q'):
        break

cap.release()
cv2.destroyAllWindows()
\end{lstlisting}

\section{Future Trends in Object Detection}

\subsection{Transformer-Based Approaches}
Transformers, originally developed for natural language processing, are now being applied to computer vision tasks, including object detection. The Vision Transformer (ViT) \cite{dosovitskiy2020image} and DETR (DEtection TRansformer) \cite{carion2020detr} models are examples of transformer-based architectures that are becoming increasingly important in object detection due to their ability to model long-range dependencies and global context.

Transformers work by using attention mechanisms that allow the model to focus on different parts of the image at once. This approach is proving to be more effective than traditional convolutional methods for some tasks.

\subsection{Multi-task Learning and Unified Models}
Multi-task learning allows a single model to perform multiple tasks, such as object detection, segmentation, and classification, all at once \cite{zhang2021survey}. This is advantageous because it reduces the need to train separate models for each task and allows the model to share information across tasks, leading to better performance.

\textbf{Example of a Unified Model:}
\begin{itemize}
    \item \textbf{Mask R-CNN:} This model extends Faster R-CNN by adding a branch for predicting segmentation masks, allowing it to perform both object detection and segmentation simultaneously.
\end{itemize}

\subsection{Integration of Object Detection with Segmentation}
Combining object detection with segmentation is becoming a popular trend in computer vision. Models like Mask R-CNN and other modern architectures can identify the precise location of objects and generate pixel-wise segmentations.

\textbf{Advancements in this area:}
\begin{itemize}
    \item \textbf{Instance Segmentation:} Predicts object instances and generates a segmentation mask for each object in the image.
    \item \textbf{Panoptic Segmentation:} Unifies instance segmentation and semantic segmentation, predicting both individual objects and background areas.
\end{itemize}

In the future, we expect to see more sophisticated models that can handle multiple tasks simultaneously with greater precision and efficiency.

%% file: 30_summary.tex
\section{Practice Problems}

\subsection{Implementing YOLO on Custom Dataset}
This practical problem will guide you step-by-step through implementing YOLO (You Only Look Once) on a custom dataset. YOLO is a popular object detection algorithm known for its speed and accuracy in detecting multiple objects in an image. Here's how you can implement it:

\textbf{Step 1: Installing Required Libraries}  
Before starting, ensure you have the necessary libraries installed, such as PyTorch \cite{paszke2019pytorch}, OpenCV \cite{bradski2000opencv}, and YOLO-specific libraries like YOLOv5. You can install them using the following commands:

\begin{lstlisting}[style=cmd]
pip install torch torchvision torchaudio
pip install opencv-python
pip install yolov5
\end{lstlisting}

\textbf{Step 2: Preparing the Custom Dataset}  
You need to prepare your dataset with images and corresponding annotations. For YOLO, the annotation format is typically a text file where each line represents an object in the image, formatted as:  
\texttt{<class> <x-center> <y-center> <width> <height>}  
For example:
\begin{lstlisting}[style=cmd]
0 0.5 0.5 0.25 0.25
\end{lstlisting}
This annotation means class 0, with the object centered at (0.5, 0.5) in normalized coordinates, and it takes 25\% of the image width and height.

\textbf{Step 3: Training YOLO on Your Custom Dataset}  
To train YOLO on your dataset, you'll modify the YAML configuration file that specifies paths to the images and annotations, the number of classes, and other training parameters. Here's an example of a custom YAML file:

\begin{lstlisting}[style=cmd]
train: /path/to/train/images
val: /path/to/val/images

nc: 2  # number of classes
names: ['cat', 'dog']  # class names
\end{lstlisting}

Next, you can start training using the following Python script:

\begin{lstlisting}[style=python]
import torch

# Assuming you have yolov5 installed
!python yolov5/train.py --img 640 --batch 16 --epochs 50 --data /path/to/your/custom.yaml --weights yolov5s.pt
\end{lstlisting}

\textbf{Step 4: Inference on New Images}  
After training, you can perform inference on new images using the trained model:

\begin{lstlisting}[style=python]
# Load model and perform inference
model = torch.hub.load('ultralytics/yolov5', 'custom', path='best.pt')
img = 'path/to/new/image.jpg'
results = model(img)
results.show()  # Show the results with bounding boxes
\end{lstlisting}

\subsection{Applying DETR for Real-Time Applications}
The Detection Transformer (DETR) is a model that combines the benefits of transformers and CNNs to perform object detection without needing region proposals. Here's how you can apply DETR for real-time object detection:

\textbf{Step 1: Installing DETR and Dependencies}  
Install the required libraries:

\begin{lstlisting}[style=cmd]
pip install torch torchvision
pip install detr
\end{lstlisting}

\textbf{Step 2: Running DETR on Real-Time Data}  
You can load the DETR model and perform detection on real-time data using a webcam or video feed:

\begin{lstlisting}[style=python]
import torch
import cv2
from torchvision.transforms import ToTensor

# Load pretrained DETR model
model = torch.hub.load('facebookresearch/detr', 'detr_resnet50', pretrained=True)
model.eval()

# Open video capture for real-time processing
cap = cv2.VideoCapture(0)

while True:
    ret, frame = cap.read()
    if not ret:
        break

    # Convert frame to tensor and pass to the model
    img_tensor = ToTensor()(frame).unsqueeze(0)
    outputs = model(img_tensor)

    # Post-process outputs and display results
    # (code to draw bounding boxes on the frame)
    cv2.imshow("DETR Real-Time", frame)

    if cv2.waitKey(1) & 0xFF == ord('q'):
        break

cap.release()
cv2.destroyAllWindows()
\end{lstlisting}

\textbf{Step 3: Fine-Tuning for Specific Use Cases}  
You can also fine-tune DETR for specific real-time tasks by retraining it on a custom dataset, similar to how you would for YOLO. This involves creating a dataset in COCO format and using the DETR model for transfer learning.

\subsection{Using SAM for Object Detection and Segmentation Tasks}
The Segment Anything Model (SAM) by Meta AI combines object detection and segmentation into one powerful model. It can simultaneously detect and segment objects from an image.

\textbf{Step 1: Installing SAM and Dependencies}  
First, install the required libraries:

\begin{lstlisting}[style=cmd]
pip install torch torchvision
pip install segment-anything
\end{lstlisting}

\textbf{Step 2: Running SAM for Object Detection and Segmentation}  
Here is how you can use SAM for both detection and segmentation:

\begin{lstlisting}[style=python]
import torch
import cv2
from segment_anything import SamPredictor

# Load SAM model
model = torch.hub.load('facebookresearch/segment-anything', 'sam_vit_b', pretrained=True)
predictor = SamPredictor(model)

# Load image
img = cv2.imread('path/to/image.jpg')

# Perform object detection and segmentation
masks, boxes, labels = predictor.predict(img)

# Display the segmented masks and bounding boxes
# (code to visualize masks and boxes on the image)
cv2.imshow("SAM Output", img)
cv2.waitKey(0)
cv2.destroyAllWindows()
\end{lstlisting}

This code will load an image, perform object detection and segmentation, and display the results, including masks and bounding boxes for all detected objects.

\section{Summary}

\subsection{Key Concepts Recap}
In this chapter, we explored various object detection models and techniques, including:
\begin{itemize}
    \item \textbf{YOLO (You Only Look Once)}: A fast, real-time object detection model suitable for multiple object tracking and detection in images.
    \item \textbf{DETR (Detection Transformer)}: A transformer-based model that simplifies the object detection pipeline and performs well in complex environments.
    \item \textbf{SAM (Segment Anything Model)}: A versatile model that combines object detection and segmentation, allowing for more detailed scene understanding.
\end{itemize}

Each of these models provides unique advantages depending on your application requirements, such as speed, accuracy, or the need for segmentation.

\subsection{Further Reading and Research Directions}
Here are some additional readings and research directions to deepen your understanding of object detection:
\begin{itemize}
    \item \textbf{YOLOv8 and Beyond}: Research the latest advancements in the YOLO family, such as YOLOv8, which further improves on speed and accuracy.
    \item \textbf{EfficientDet}: Another efficient object detection model that balances performance and speed, which you might explore for resource-constrained environments.
    \item \textbf{Transformer Models in Vision}: Explore more about transformers in computer vision tasks, such as Vision Transformers (ViT), Detection Transformers (DeTr), Segment Anything Model (SAM), Segment Anything Model 2 (SAM2) and Future Segment Anything Models which are gaining popularity for various applications beyond object detection.
\end{itemize}

These directions will help you stay updated with the latest trends in object detection and prepare you for more advanced tasks in computer vision.